\documentclass[preprint,12pt,authoryear]{elsarticle}

\usepackage{amssymb}

\usepackage{url}            
\usepackage{amsmath}
\usepackage{graphicx}
\usepackage{multirow}
\usepackage{spverbatim}
\usepackage{subcaption} 
\usepackage{array} 

\journal{Neural Networks}

\begin{document}

\begin{frontmatter}

\title{Context Meta-Reinforcement Learning via Neuromodulation}

\author[1]{Eseoghene Ben-Iwhiwhu\corref{cor1}}
\ead{e.ben-iwhiwhu@lboro.ac.uk}

\author[1]{Jeffery Dick}
\ead{j.dick@lboro.ac.uk}

\author[2]{Nicholas A. Ketz}
\ead{naketz@hrl.com}

\author[2]{Praveen K. Pilly}
\ead{pkpilly@hrl.com}

\author[1]{Andrea Soltoggio}
\ead{a.soltoggio@lboro.ac.uk}

\cortext[cor1]{Corresponding author}

\affiliation[1]{organization={Department of Computer Science, Loughborough University},
    country={UK}}

\affiliation[2]{organization={HRL Laboratories},
    addressline={Malibu, California},
    country={US}}

\begin{abstract}
Meta-reinforcement learning (meta-RL) algorithms enable agents to adapt quickly to tasks from few samples in dynamic environments. Such a feat is achieved through dynamic representations in an agent's policy network (obtained via reasoning about task context, model parameter updates, or both). However, obtaining rich dynamic representations for fast adaptation beyond simple benchmark problems is challenging due to the burden placed on the policy network to accommodate different policies. This paper addresses the challenge by introducing neuromodulation as a modular component to augment a standard policy network that regulates neuronal activities in order to produce efficient dynamic representations for task adaptation. The proposed extension to the policy network is evaluated across multiple discrete and continuous control environments of increasing complexity. To prove the generality and benefits of the extension in meta-RL, the neuromodulated network was applied to two state-of-the-art meta-RL algorithms (CAVIA and PEARL). The result demonstrates that meta-RL augmented with neuromodulation produces significantly better result and richer dynamic representations in comparison to the baselines.
\end{abstract}


\end{frontmatter}

\section{Introduction}
\label{section:introduction}
Human intelligence, though specialized in some sense, is able to generally adapt to new tasks and solve problems from limited experience or few interactions. The field of meta-reinforcement learning (meta-RL) seeks to replicate such a flexible intelligence by designing agents that are capable of rapidly adapting to tasks from few interactions in an environment. The recent progress in the field such as \citep{rakelly2019efficient, finn2017model, duan2016rl, wang1611learning, zintgraf2019fast, gupta2018meta} have showcased start-of-the-art results. Studies with agents endowed with such adaptation capabilities are a promising venue for developing much desired and needed artificial intelligence systems and robots with lifelong learning dynamics.

When an agent's policy for a meta-RL problem is encoded by a neural network, neural representations are adjusted from a base pre-trained point to a configuration that is optimal to solve a specific task. Such dynamic representations are a key feature to enable an agent to rapidly adapt to different tasks. These representations can be derived from gradient-based approaches \citep{finn2017model}, context-based approaches such as memory \citep{mishra2018a, wang1611learning, duan2016rl} and probabilistic \citep{rakelly2019efficient}, or hybrid approaches (i.e., combination of gradient and context methods) \citep{zintgraf2019fast}. The hybrid approach obtains a task context via gradient updates and thus dynamically alters the representations of the network. Context approaches such as CAVIA \citep{zintgraf2019fast} and PEARL \citep{rakelly2019efficient} are more interpretable as they disentangle task context from the policy network, thus the task context is used to achieve optimal policies for different tasks. 

One limitation of such approaches is that they do not scale well as the problem complexity increases because of the demand to store many diverse policies to be reached within a single network. In particular, it is possible that, as tasks grow in complexity, the tasks similarities reduce and thus the network's representations required to solve each task optimally becomes dissimilar. We hypothesize that standard policy networks are not likely to produce diverse policies from a trained base representation because all neurons have a homogeneous role or function: thus, significant changes in the policy require widespread changes across the network. From this observation, we speculate that a network endowed with modulatory neurons (neuromodulators) has a significantly higher ability to modify its policy. 

Our approach to overcome this limiting design factor in current meta-RL neural approaches is to introduce a neuromodulated policy network to increase its ability to encode rich and flexible dynamic representations. The rich representations are measured based on the dissimilarity of the representations across various tasks, and are useful when the optimal policy of an agent (input-to-action mapping) is less similar across tasks. When combined with the CAVIA and PEARL meta-learning frameworks, the proposed approach produced better dynamic representations for fast adaptation as the neuromodulators in each layer serve as a means of directly altering the representations of the layer in addition to the task context.

Several designs exist for neuromodulation \citep{doya2002metalearning}, either to gate plasticity \citep{soltoggio2008evolutionary, miconi2020backpropamine}, gate neural activations \citep{beaulieu2020learning} or alter high level behaviour \citep{xing2020neuromodulated}. The proposed mechanism in this work focuses on just one simple principle: modulatory signals alter the representations in each layer by gating the weighted sum of input of the standard neural component.

The primary contribution of this work is a neuromodulated policy network for meta-reinforcement learning for solving increasingly difficult problems. The modular approach of the design allows for the proposed layer to be used with other existing layers (such as standard fully connected layer, convolutional layer and so on) when stacking them to form a deep network. The experimental evidence in this work demonstrates that neuromodulation is beneficial to adapt network representations with more flexibility in comparison to standard networks. Experimental evaluations were conducted across high dimensional discrete and continuous control environments of increasing complexity using CAVIA and PEARL meta-RL algorithms. The results indicate that the neuromodulated networks show an increasing advantage as the problem complexity increases, while they perform comparably on simpler problems. The increased diversity of the representations from the neuromodulated policy network are examined and discussed. The open source implementation of the code can be found at: \url{https://github.com/dlpbc/nm-metarl}

\section{Related Work}
\label{section:related-work}
\textbf{Meta-reinforcement learning.} This work builds on the existing meta learning frameworks \citep{bengio1992optimization, schmidhuber1996simple, thrun1998learning, schweighofer2003meta} in the domain of reinforcement learning. Recent studies in meta-reinforcement learning (meta-RL) can be largely classified into optimization and context-based methods. Optimization methods \citep{finn2017model, li2017meta, stadie2018some, rothfuss2018promp} seek to learn good initial parameters of a model that can be adapted with a few gradient steps to a specific task. In contrast, context-based methods seek to adapt a model to a specific task based on few-shot experiences aggregated into context variables. The context can be derived via probabilistic methods \citep{rakelly2019efficient, liu2021decoupling}, recurrent memory \citep{duan2016rl, wang1611learning}, recursive networks \citep{mishra2018a} or the combination of probabilistic and memory \citep{zintgraf2020varibad, humplik2019meta}. Hybrid methods \citep{zintgraf2019fast, gupta2018meta} combine optimization and context-based methods whereby task specific context parameters are obtained via gradient updates.

\textbf{Neuromodulation.}
Neuromodulation in biological brains is a process whereby a neuron alters or regulates the properties of other neurons in the brain \citep{marder2012neuromodulation}. The altered properties can either be in the cellular activities or synaptic weights of the neurons. Well known biological neuromodulators include dopamine (DA), serotonin (5-HT), acetycholine (ACh), and noradrenaline (NA) \citep{bear2020neuroscience,avery2017neuromodulatory}. Such neuromodulators were described in \cite{doya2002metalearning}  within the reinforcement learning computation framework, with dopamine loosely mapped to the reward signal error (like TD error), serotonin representing discount factor, acetycholine representing learning rate and noradrenaline representing randomness in a policy's action distribution. Several studies have drawn inspiration from neuromodulation and applied it to gradient-based RL \citep{xing2020neuromodulated, miconi2020backpropamine} and neuroevolutionary RL \citep{soltoggio2007evolving, soltoggio2008evolutionary, velez2017diffusion} for dynamic task settings. In broader machine learning, neuromodulation has been applied to goal-driven perception \citep{zou2020neuromodulated}, and also in continual learning setting \citep{beaulieu2020learning} where it was combined with meta-learning to sequentially learn a number of classification tasks without catastrophic forgetting. The neuromodulators used in these studies have different designs or functions: plasticity gating \citep{soltoggio2008evolutionary, miconi2020backpropamine}, activation gating \citep{beaulieu2020learning}, direct action modification in a policy \citep{xing2020neuromodulated}.

\section{Background}
\label{section:background}
\subsection{Problem Formulation}
\label{subsection:problem-formulation}
In a meta-RL setting, tasks are sampled from a task distribution $p(\mathcal{T})$. Each task $\mathcal{T}_i$ is a Markov Decision Process (MDP), which is a tuple $M_{i}$=\{$\mathcal{S}, \mathcal{A}, q, r, q_0$\} consisting of a state space $\mathcal{S}$, an action space $\mathcal{A}$, a state transition distribution $q(s_{t+1} | s_t, a_t)$, a reward function $r(s_t, a_t, s_{t+1})$, and an initial state distribution $q_0(s_0)$. When presented with a task $\mathcal{T}_i$, an agent (with a policy $\pi$) is required to quickly adapt to the task from few interactions. Therefore, the goal of the agent for each task is to maximize the expected reward in the shortest time possible:

\begin{equation} \label{eq:rl_objective}
	\mathcal{J}(\pi) = \mathbf{E}_{q_0, q, \pi} \left[ \sum_{t=0}^{H-1} \gamma^t r(s_t, a_t, s_{t+1}) \right],
\end{equation}
where $H$ is a finite horizon and $\gamma \in [0,1]$ is the discount factor.

\subsection{Context Adaptation via Meta-Learning (CAVIA)}
\label{subsection:cavia}
The CAVIA meta-learning framework \citep{zintgraf2019fast} is an extension of the \emph{model-agnostic meta-learning algorithm (MAML)} \citep{finn2017model} that is interpretable and less prone to meta-overfitting. The key idea in CAVIA is the introduction of context parameters in a policy network. Therefore, the policy $\pi_{\theta, \phi}$ contains the standard network parameters $\theta$ and the context parameters $\phi$. During the adaptation phase for each task (the gradient updates in the inner loop), only the context parameters are updated, while the network parameters are updated during the outer loop gradient updates. There are different ways to provide the policy network with the context parameters. In \cite{zintgraf2019fast}, the parameters were concatenated to the input.

In the meta-RL framework, an agent is trained for a number of iterations. For each iteration, $N$ tasks represented as $\mathbf{T}$ are sampled from the task distribution $\mathcal{T}$. For each task $i$, a batch of trajectories $\tau^{train}_{i}$ is obtained using the policy $\pi_{\theta, \phi}$ with the context parameters set to an initial condition $\phi_0$. The obtained trajectories for task $i$ are used to perform a one step inner loop gradient update of the context parameters to new values $\phi_i$, shown in the equation below:

\begin{equation} \label{eq:inner_update_metarl}
\phi_i = 
\phi_0 - 
\alpha
\nabla_\phi 
\mathcal{J}_{\mathcal{T}_i} (\tau^{train}_{i}, \pi_{\theta, \phi_0}),
\end{equation}

where $\mathcal{J}_{\mathcal{T}_i} (\tau_{i}, \pi_{\theta, \phi})$ is the objective function for task $i$. After the one step gradient update of the policy, another batch of trajectories $\tau^{test}_{i}$ is collected using the updated task specific policy $\pi_{\theta, \phi_i}$.

After completing the above procedure for all tasks sampled from $\mathcal{T}$, a meta gradient step (also referred to as the outer loop update) is performed, updating $\theta$ to maximize the average performance of the policy across the task batch.
\begin{equation}
\theta = 
\theta - 
\beta
\nabla_\theta 
\frac{1}{N}
\sum\limits_{\tau_i \in \mathbf{T}}
\mathcal{J}_{\mathcal{T}_i} (\tau_i^{test}, \pi_{\theta, \phi_i}).
\end{equation}

\subsection{Probabilistic Embeddings for Actor-Critic Meta-RL (PEARL)}
\label{subsection:pearl}
PEARL \citep{rakelly2019efficient} is an off-policy meta-RL algorithm that is based on the soft actor-critic architecture \citep{haarnoja2018soft}. The algorithm derives the context of the task to which an agent is exposed through probabilistic sampling. Given a task, the agent maintains a prior belief of the task, and as the agent interacts with the environment, it updates the posterior distribution with the goal of identifying the specific task context. The context variables $\mathbf{z}$ are concatenated to the input of the actor and critic neural components of the setup. To estimate this posterior $p(\mathbf{z}|\mathbf{c})$, an additional neural component called an inference network $q_{\phi}(\mathbf{z}|\mathbf{c})$ is trained using the trajectories $\mathbf{c}$ collected for tasks sampled from the task distribution $\mathcal{T}$. The objective function for the actor, critic and inference neural components are described below,

\begin{equation}
\mathcal{L}_{actor} \!=\! \mathbb{E}_{\substack{\mathbf{s} \sim \mathcal{B}, \mathbf{a} \sim \pi_{\theta} \\ \mathbf{z} \sim q_\phi(\mathbf{z} | \mathbf{c})}} \!\left[\! D_{\text{KL}}\!\left(\!\pi_{\theta}(\mathbf{a} | \mathbf{s}, \bar{\mathbf{z}}) \! \left\| \frac{\text{exp}(Q_\theta(\mathbf{s}, \mathbf{a}, \bar{\mathbf{z}}))}{\mathcal{Z}_\theta(\mathbf{s})} \! \right. \right) \! \right]
\end{equation}
\begin{equation}
\mathcal{L}_{critic} = \mathbb{E}_{\substack{(\mathbf{s}, \mathbf{a}, r, \mathbf{s}') \sim \mathcal{B} \\ \mathbf{z} \sim q_\phi(\mathbf{z} | \mathbf{c})}} [Q_\theta(\mathbf{s}, \mathbf{a}, \mathbf{z}) - (r + \bar{V}(\mathbf{s} ', \bar{\mathbf{z}}))]^2
\end{equation}
\begin{equation}
\mathcal{L}_{inference} = \mathbb{E}_{\mathcal{T}} [\mathbb{E}_{\mathbf{z} \sim q_\phi(\mathbf{z} | \mathbf{c}^\mathcal{T})}[\mathcal{L}_{critic}
+ \beta D_{\text{KL}}(q_\phi(\mathbf{z} | \mathbf{c}^\mathcal{T}) || p(\mathbf{z}))]]
\end{equation}

where $\bar{V}$ is a target network and $\bar{\mathbf{z}}$ means that gradients are not being computed through it, $p(\mathbf{z})$ is a unit Gaussian prior over $Z$, $\mathcal{B}$ is the replay buffer and $\beta$ is a weighting hyper-parameter.

\section{Neuromodulated Network}
\label{section:method}
\begin{figure}[t!]
	\centering
	\begin{subfigure}{0.45\textwidth}
		\centering
		\includegraphics[width=\textwidth]{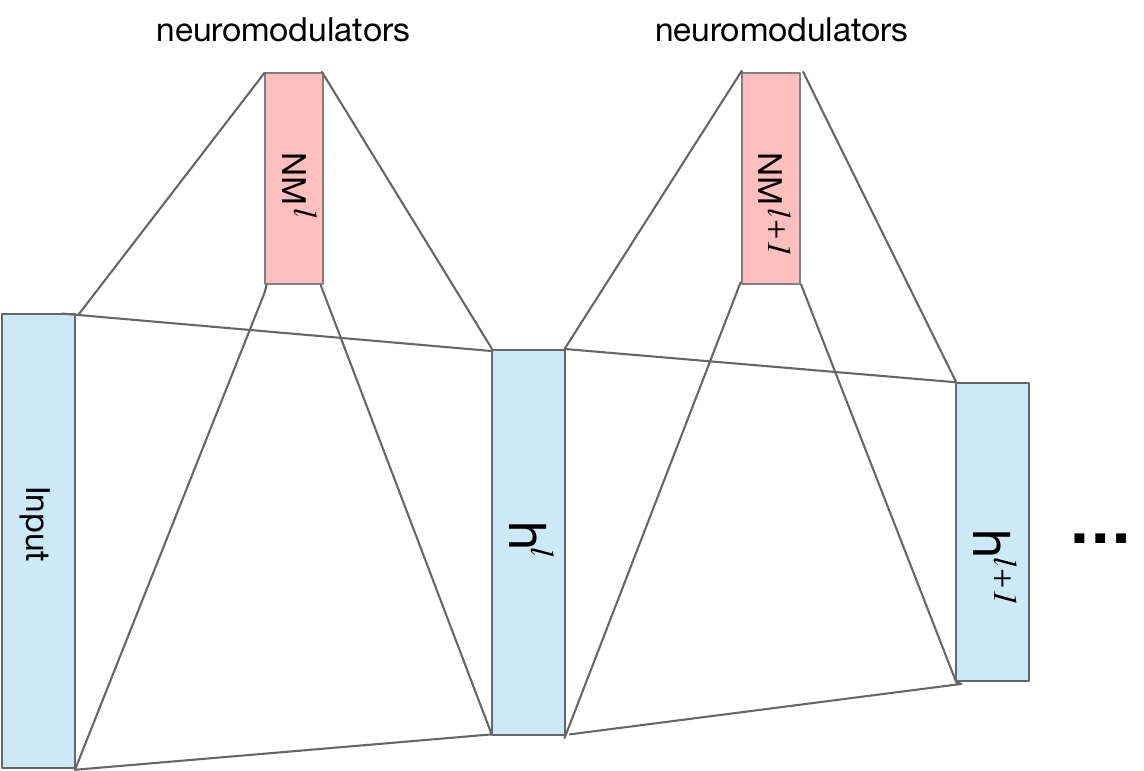}
		\caption{Modular format of the proposed architecture}
		\label{fig:nm-net}
	\end{subfigure}
	\hfill
	\begin{subfigure}{0.30\textwidth}
		\centering
		\includegraphics[width=\textwidth]{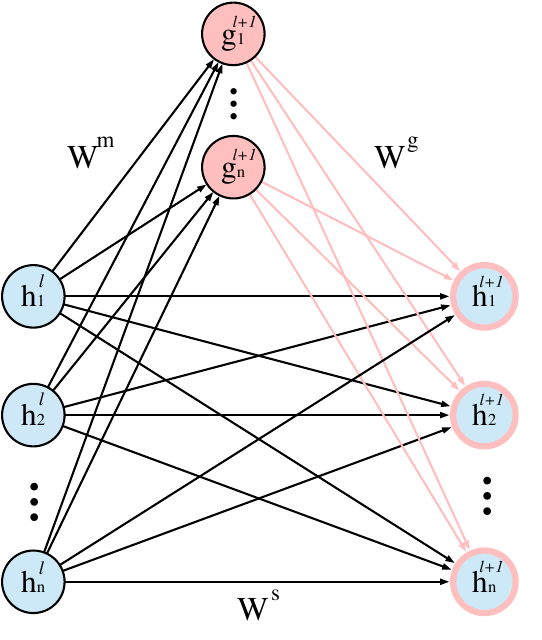}
		\caption{A single neuromodulated fully connected layer}
		\label{fig:nm-fc-layer}
	\end{subfigure}
	\caption{Overview of the proposed computational framework. (a) Light blue boxers indicate layers of standard neurons and pink boxes are layers of modulatory neurons. (b) Illustration of a single layer of the proposed architecture.}
	\label{fig:nm-policy-net}
\end{figure}
This section introduces the extension of the policy network with neuromodulation. A graphical representation of the network is shown in Figure \ref{fig:nm-net}. The neuromodulated policy network is a stack of neuromodulated fully connected layers.

\subsection{Computational Framework}
\label{subsection:computational-framework}
A neuromodulated fully connected layer contains two neural components: standard neurons and neuromodulators (see Figure \ref{fig:nm-fc-layer}). The standard neurons serve as the output of the layer (i.e., the layer's representations) and they are connected to the preceding layer via standard fully connected weights $W^s$. The neuromodulators serve as a means to alter the output of the standard neurons. They receive input via standard fully connected weights $W^g$ from the preceding layer in order to generate their neural activity, which is then projected to the standard neurons via another set of fully connected weights $W^m$. The function of the projected neuromodulatory activity defines the representation altering mechanism. For example, it could gate the plasticity of $W^s$, gate neural activation of $\mathbf{h}$ or do something else based on the designer's specification. While different types of neuromodulators can be used \citep{doya2002metalearning}, in this particular work, we employ an activity-gating neuromodulator. Such neuromodulator multiplies the activity of the target (standard) neurons before a non-linearity is applied to the layer. Formally, the structure can be described with three parameter matrices: $W^s$ defines weights connecting the input to the standard neurons, $W^g$ defines weights connecting the input to the neuromodulators and $W^m$ defines weights connecting the neuromodulators to the standard neurons. The step-wise computation of a forward pass through the neuromodulatory structure is given below:

\begin{align}
    &\mathbf{h}^s = W^s \cdot \mathbf{x} \\
    &\mathbf{g} = ReLU(W^g \cdot \mathbf{x}) \\
    &\mathbf{h}^m = tanh(W^m \cdot \mathbf{g}) \\
    &\mathbf{h} = ReLU(\mathbf{h}^s \otimes \mathbf{h}^m) \label{eq:nm-gating}
\end{align}

where $\mathbf{x}$ is the layer's input, $\mathbf{h}^s$ is the weighted sum of input of the standard neurons, $\mathbf{g}$ is activity of the neuromodulators derived from the weighted sum of input, $\mathbf{h}^m$ is the neuromodulatory activity projected onto the standard neurons, and $\mathbf{h}$ is the output of the layer. The key modulating process takes place in the element-wise multiplication of the $\mathbf{h}^s$ and $\mathbf{h}^m$.

The $tanh$ non-linearity is employed to enable positive and negative neuromodulatory signals, and thus gives the network the ability to affect both the magnitude and the sign of target activation values. When $ReLU$ is used as the non-linearity for the layer's output $h$, $\mathbf{h}^m$ has the intrinsic ability to dynamically turn on or off certain output in $\mathbf{h}$.

A simpler version of the proposed model can be achieved by only considering the sign, and not the magnitude, of the neuromodulatory signal, using the following variation of Equation \ref{eq:nm-gating}:

\begin{align}
    &\mathbf{h} = ReLU(\mathbf{h}^s \otimes \textbf{sign}(\mathbf{h}^m))
\end{align}
This variation is shown to be suited for discrete control problems. 

\section{Results and Analysis}
\label{section:results-and-analysis}
In this section, the results of the neuromodulated policy network evaluations across high dimensional discrete and continuous control environments with varying levels of complexity are presented. The continuous control environments are the simple 2D navigation, the half-cheetah direction \citep{finn2017model} and velocity \citep{finn2017model} Mujoco \citep{todorov2012mujoco} based environments and the meta-world ML1 and ML45 environments \citep{yu2020meta}. The discrete action environment is a graph navigation environment that supports configurable levels of complexity called the CT-graph \citep{soltoggio2021ctgraph, ladosz2021deep,ben2020evolving}. The experimental setup focused on investigating the beneficial effect of the proposed neuromodulatory mechanism when augmenting existing meta-RL frameworks (i.e., neuromodulation as complementary tool to meta-RL rather than competing). To this end, using CAVIA meta-RL method \citep{zintgraf2019fast}, a standard policy network (SPN) is compared against the neuromodulated policy network (NPN) across the aforementioned environments. Similarly, SPN is compared against NPN using PEARL \citep{rakelly2019efficient} method only in the continuous control environments because the soft actor-critic architecture employed by PEARL is designed for continuous control. We present the analysis of the learned dynamic representations from a standard and a neuromodulated network in Section \ref{subsection:analysis}. Finally, the policy networks were evaluated in a RGB autonomous vehicle navigation domain in the CARLA driving simulator using CAVIA and the results and discussions are presented in Appendix \ref{appendix:additional-experiments}.

\subsection{Performance}
\label{subsection:performance}
\begin{figure}[t!]
	\centering
	\begin{subfigure}{0.3\textwidth}
		\centering
		\includegraphics[width=\textwidth]{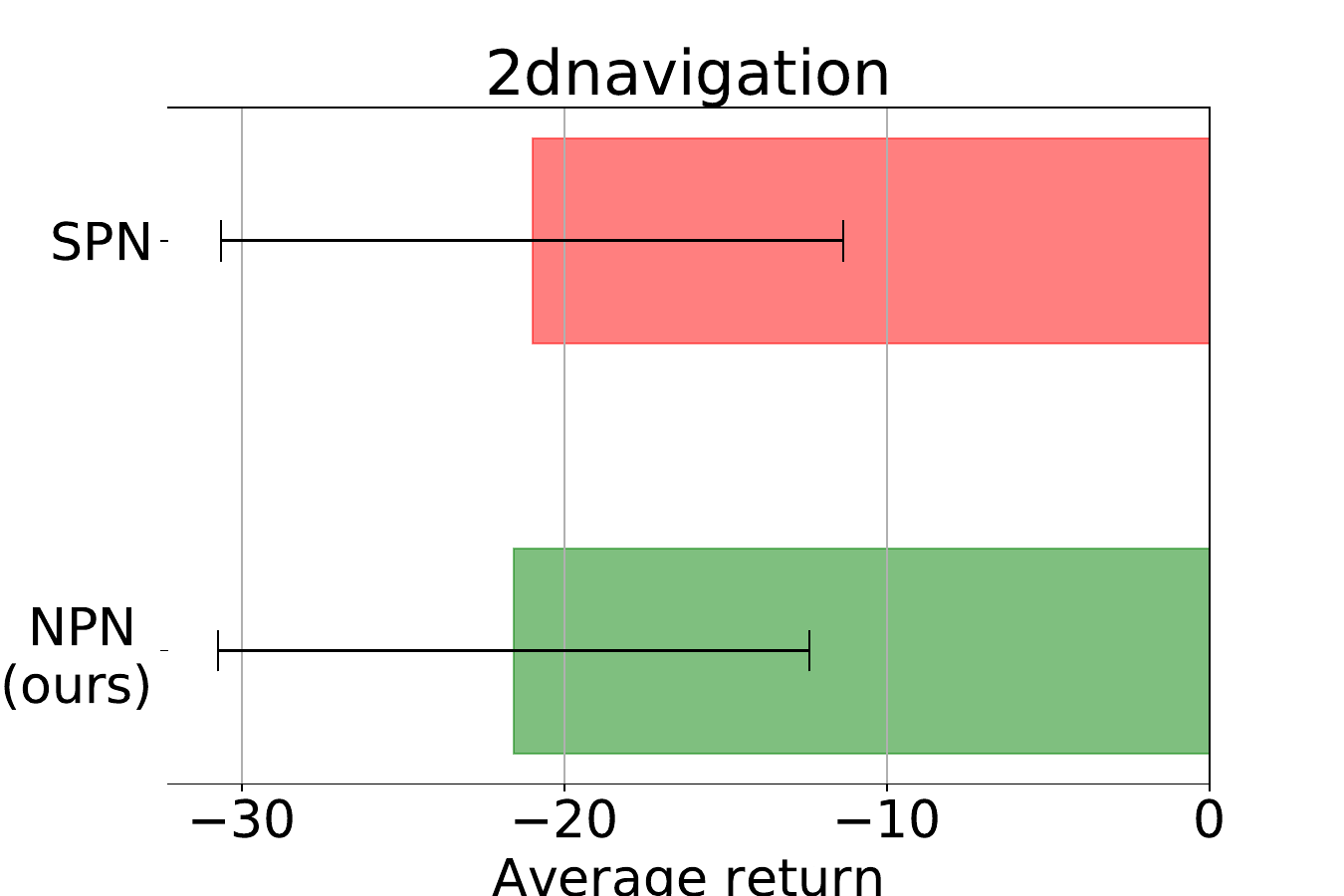}
		\caption{2D Navigation}
		\label{fig:res-cavia-2dnavigation}
	\end{subfigure}
	\begin{subfigure}{0.3\textwidth}
		\centering
		\includegraphics[width=\textwidth]{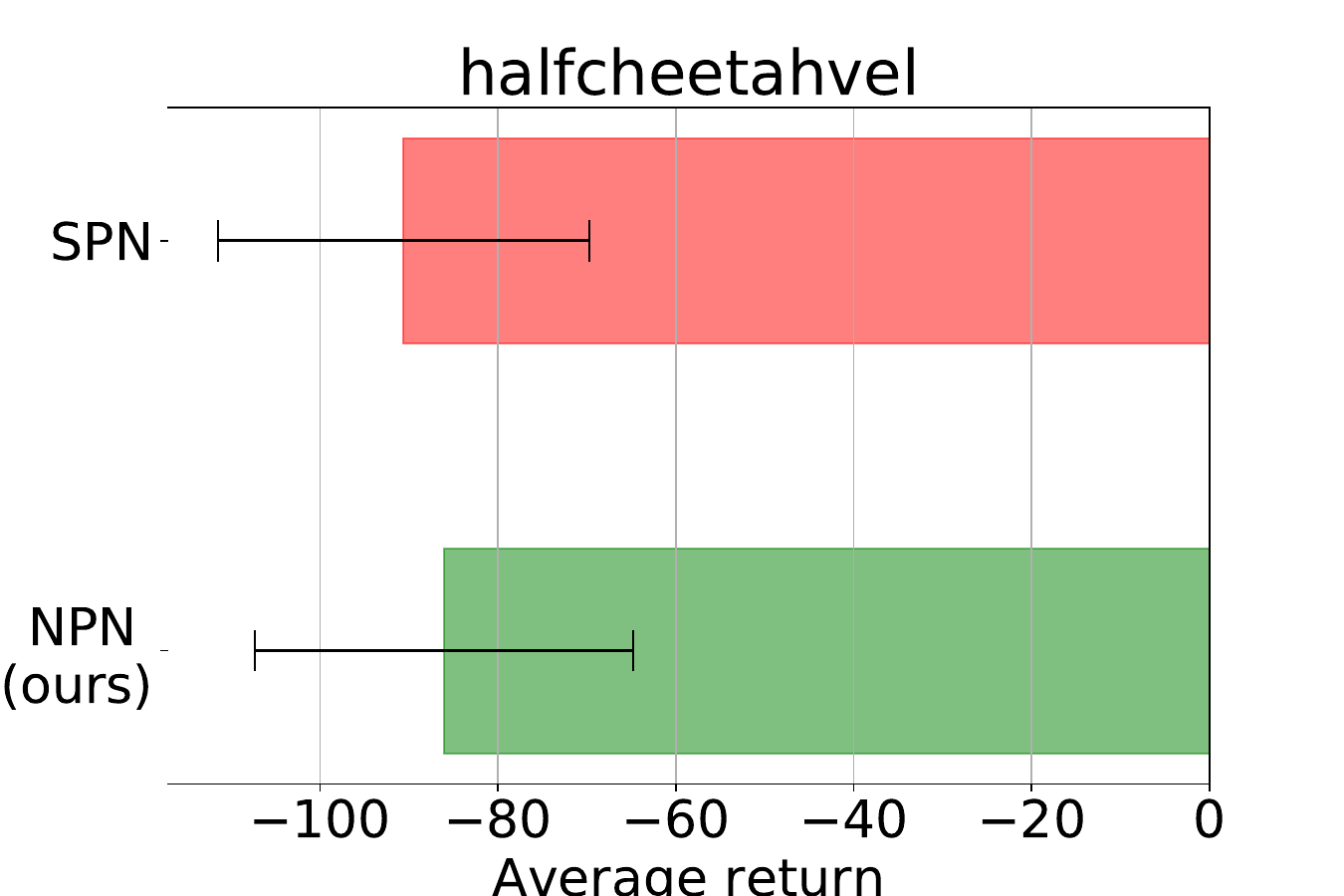}
		\caption{Half-Cheetah Velocity}
		\label{fig:res-cavia-halfcheetah-vel}
	\end{subfigure}
	\begin{subfigure}{0.3\textwidth}
		\centering
		\includegraphics[width=\textwidth]{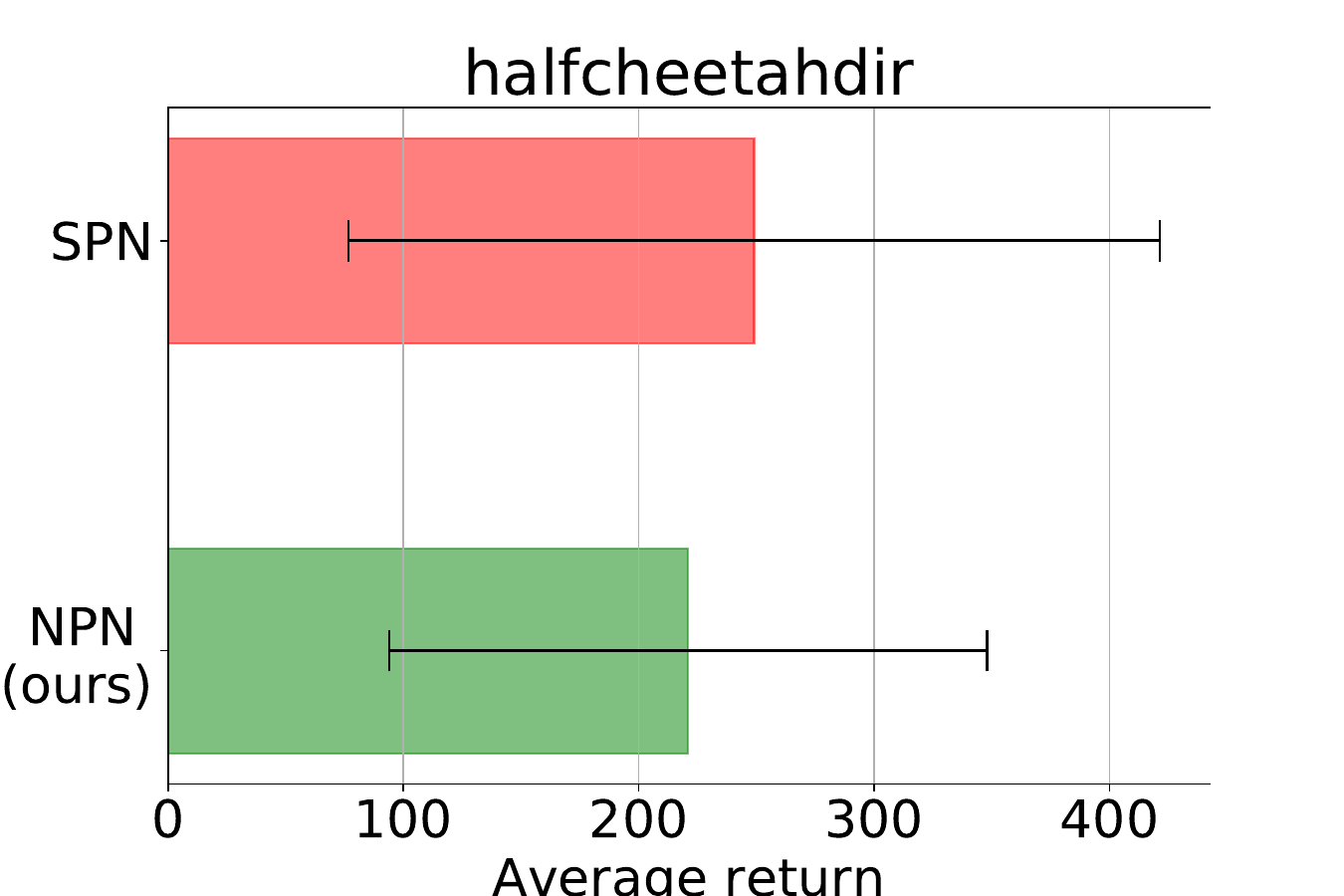}
		\caption{Half-Cheetah Direction}
		\label{fig:res-cavia-halfcheetah-dir}
	\end{subfigure}
	\begin{subfigure}{0.3\textwidth}
		\centering
		\includegraphics[width=\textwidth]{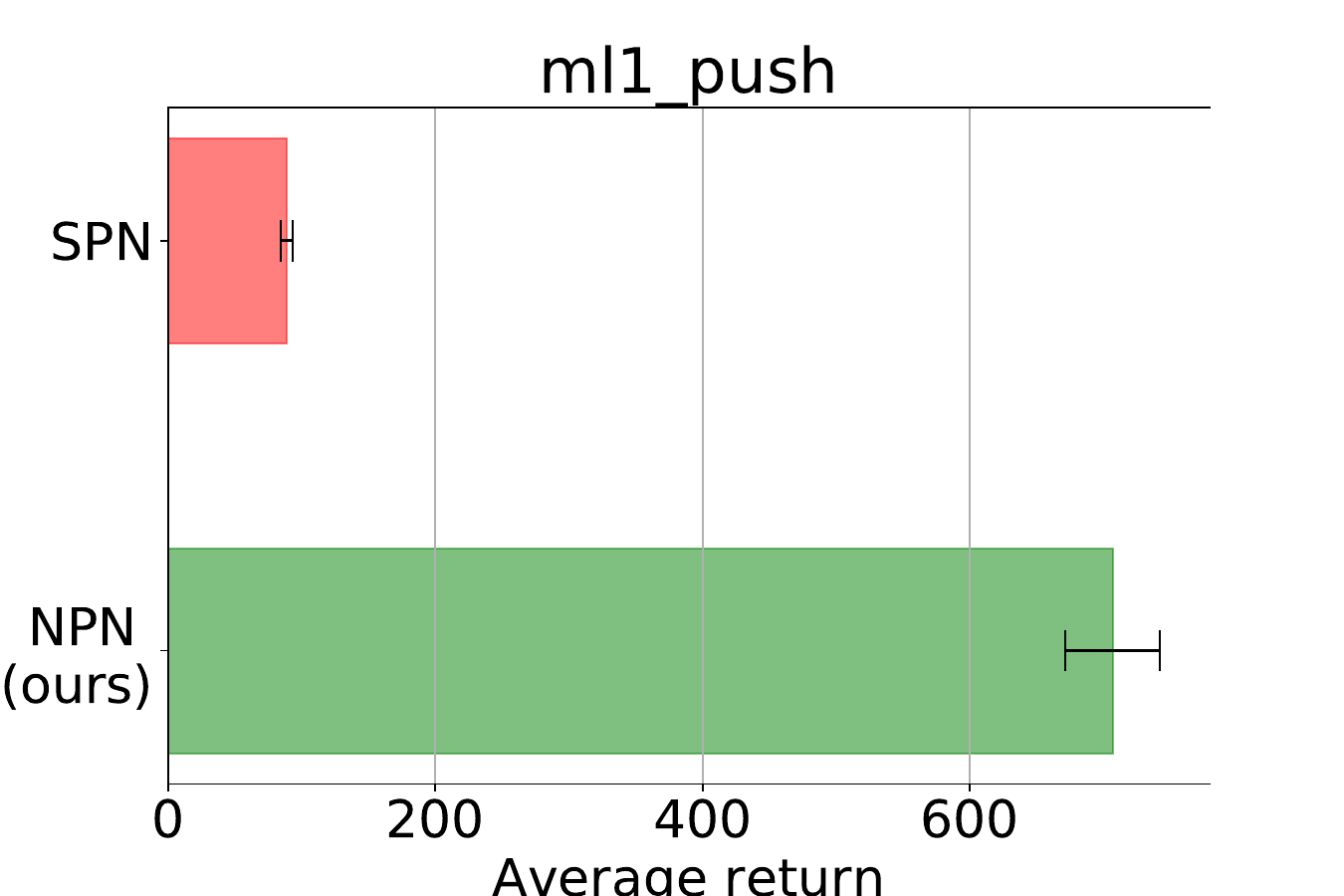}
		\caption{Meta-World ML1}
		\label{fig:res-cavia-ml1}
	\end{subfigure}
	\begin{subfigure}{0.3\textwidth}
		\centering
		\includegraphics[width=\textwidth]{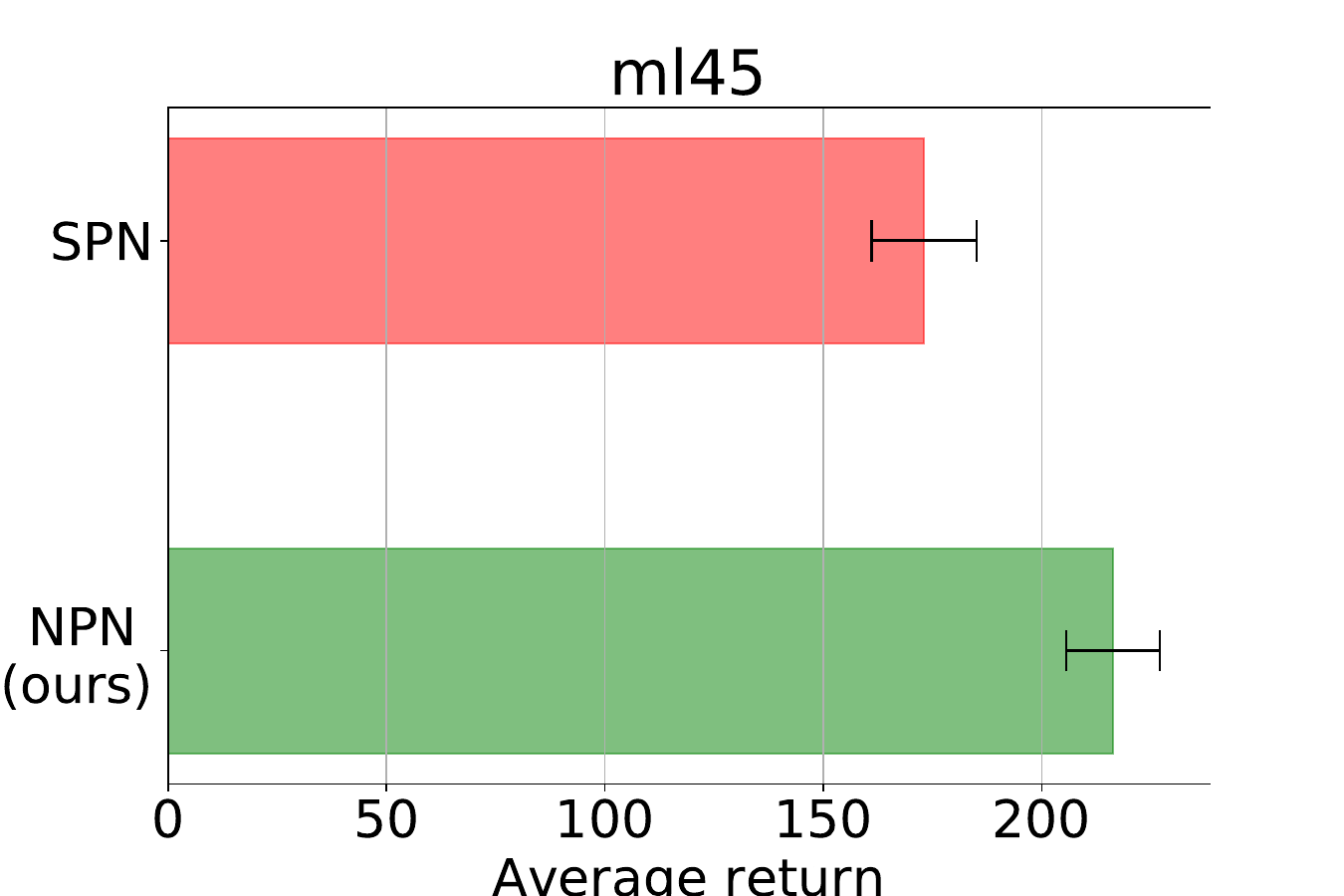}
		\caption{Meta-World ML45}
		\label{fig:res-cavia-ml45}
	\end{subfigure}
	\caption{Adaptation performance across tasks of the standard policy network (SPN) and the neuromodulated policy network (NPN) in continuous control environment using CAVIA meta-RL framework. Across three seed runs, the performance was measured based on average return from the rewards acquired during evaluation.}
	\label{fig:res-cavia-continuous-control}
\end{figure}

\begin{figure}[b!]
	\centering
	\begin{subfigure}{0.3\textwidth}
		\centering
		\includegraphics[width=\textwidth]{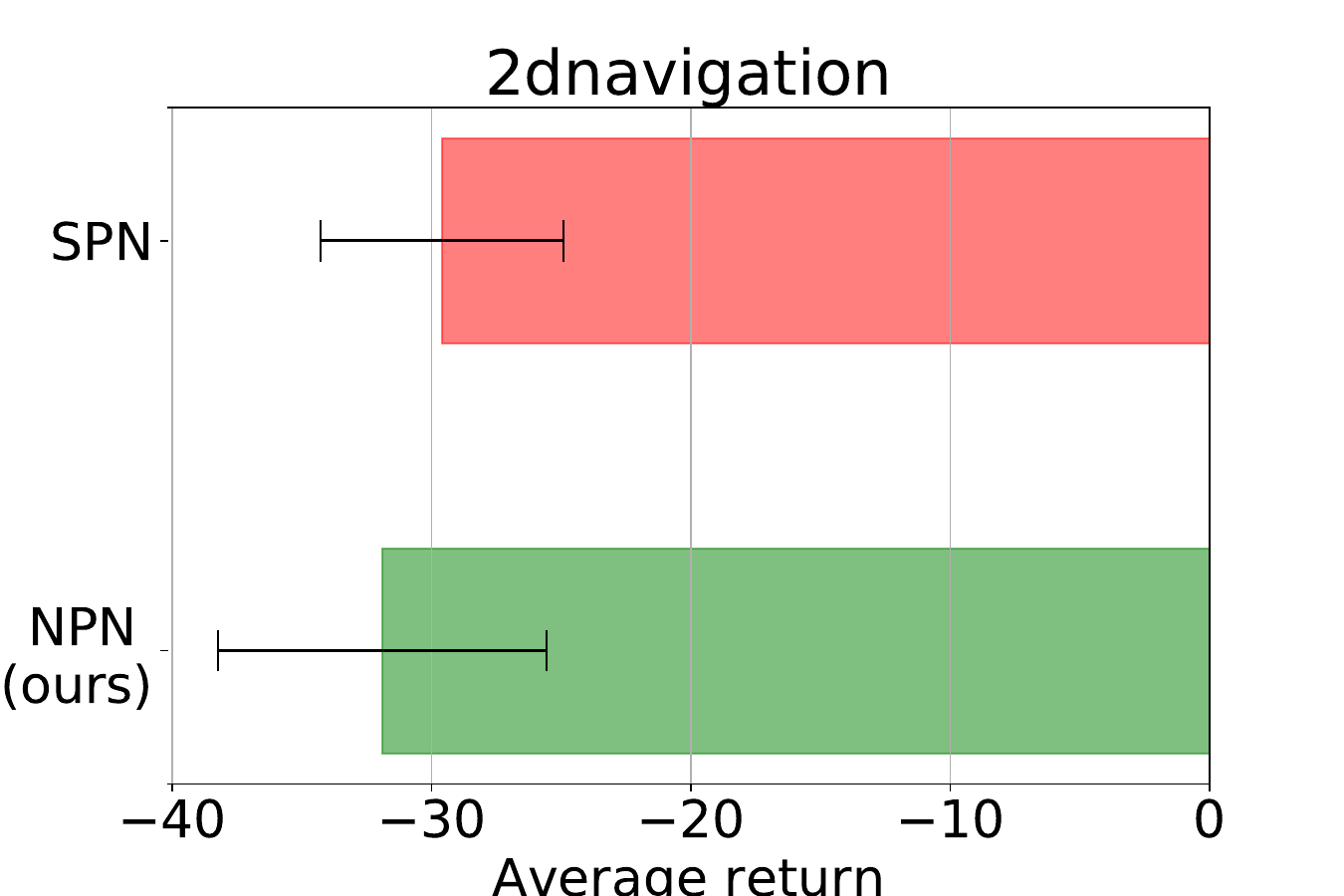}
		\caption{2D Navigation}
		\label{fig:res-pearl-2dnavigation}
	\end{subfigure}
	\begin{subfigure}{0.3\textwidth}
		\centering
		\includegraphics[width=\textwidth]{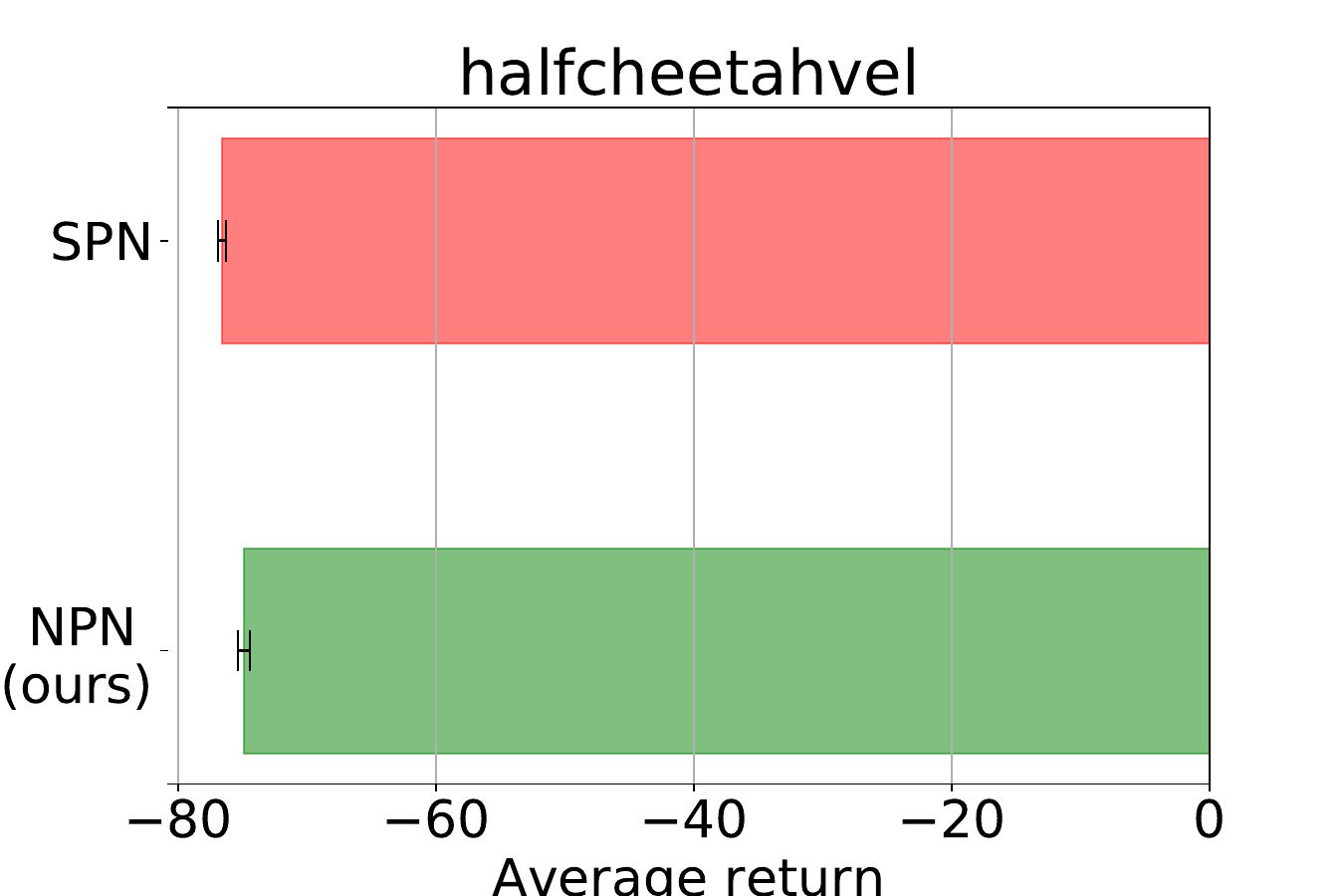}
		\caption{Half-Cheetah Velocity}
		\label{fig:res-pearl-halfcheetah-vel}
	\end{subfigure}
	\begin{subfigure}{0.3\textwidth}
		\centering
		\includegraphics[width=\textwidth]{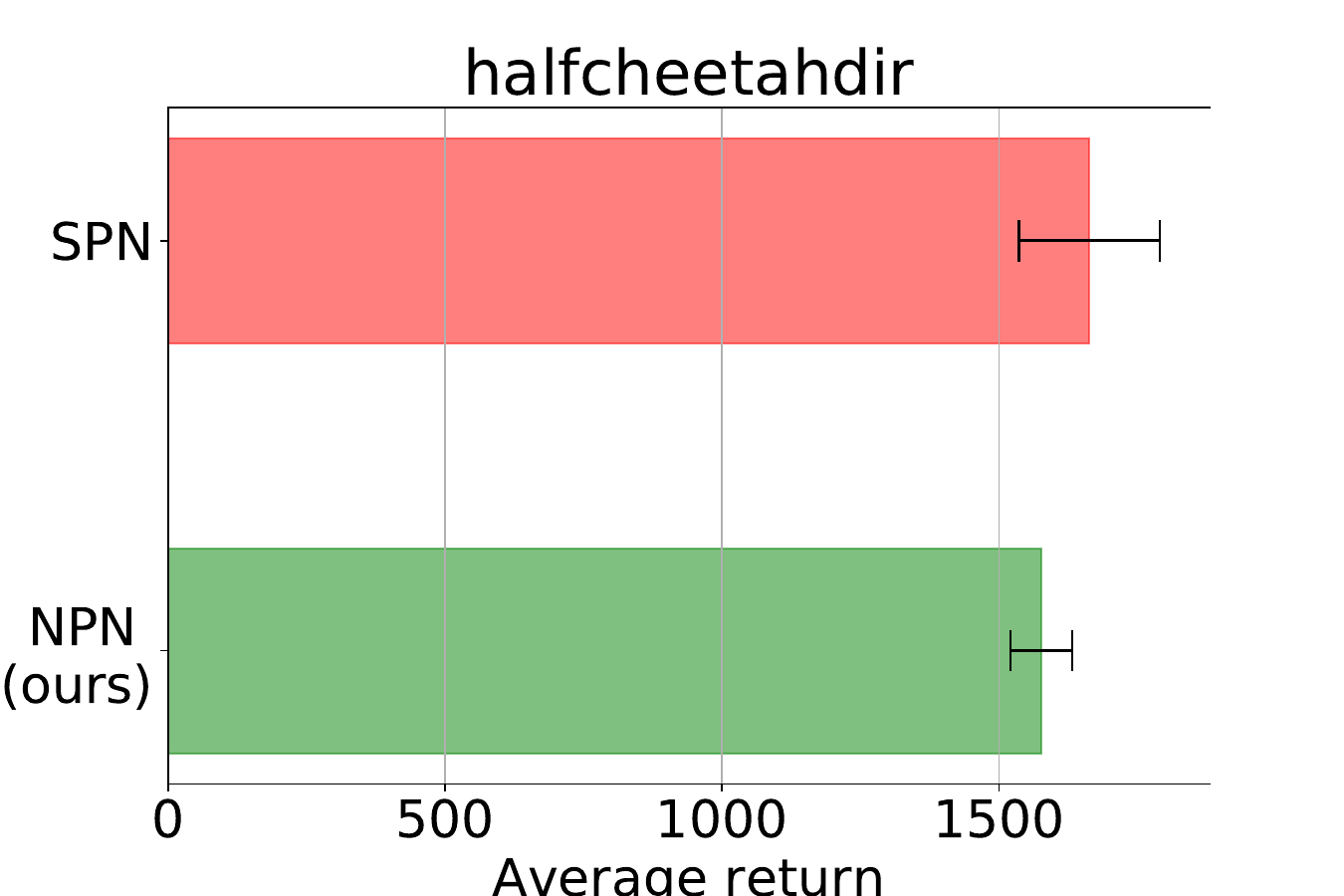}
		\caption{Half-Cheetah Direction}
		\label{fig:res-pearl-halfcheetah-dir}
	\end{subfigure}
	\begin{subfigure}{0.3\textwidth}
		\centering
		\includegraphics[width=\textwidth]{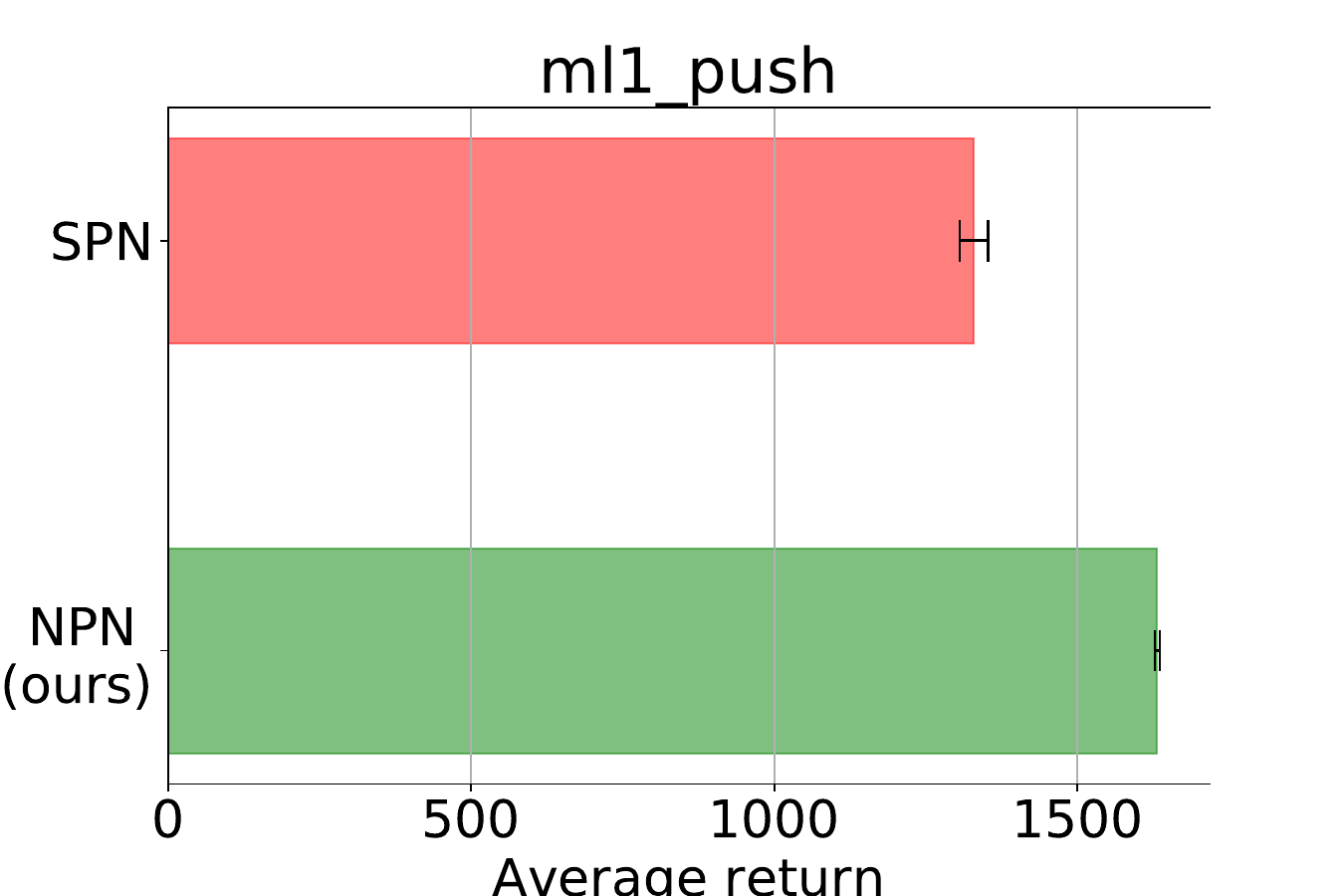}
		\caption{Meta-World ML1}
		\label{fig:res-pearl-ml1}
	\end{subfigure}
	\begin{subfigure}{0.3\textwidth}
		\centering
		\includegraphics[width=\textwidth]{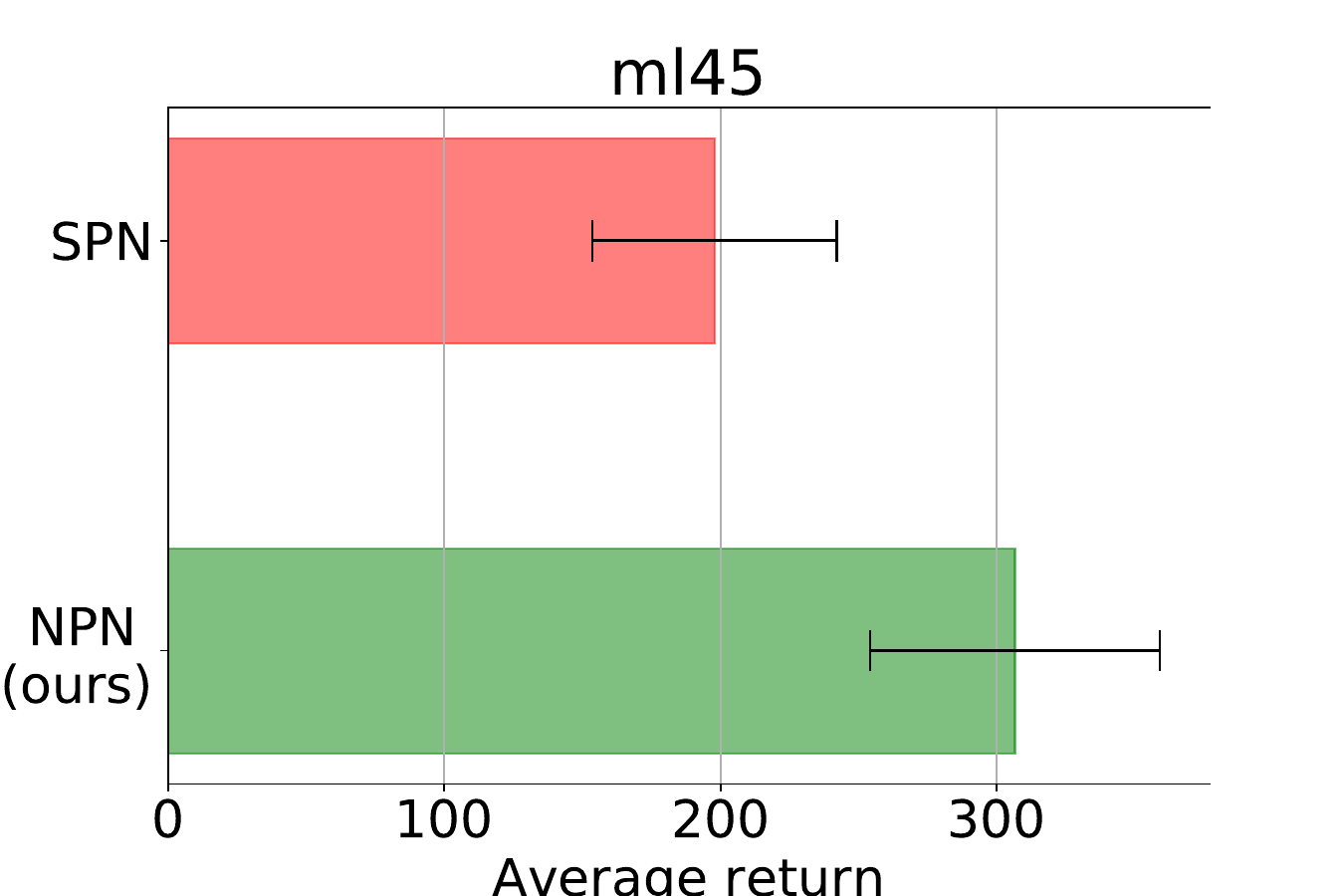}
		\caption{Meta-World ML45}
		\label{fig:res-pearl-ml45}
	\end{subfigure}
	\caption{Adaptation performance across tasks of the standard policy network (SPN) and the neuromodulated policy network (NPN) in continuous control environment using PEARL meta-RL framework. Across three seed runs, the performance was measured based on average return from the rewards acquired during evaluation.}
	\label{fig:res-pearl-continuous-control}
\end{figure}

\begin{figure}
	\centering
	\begin{subfigure}{0.22\textwidth}
		\centering
		\includegraphics[width=\textwidth]{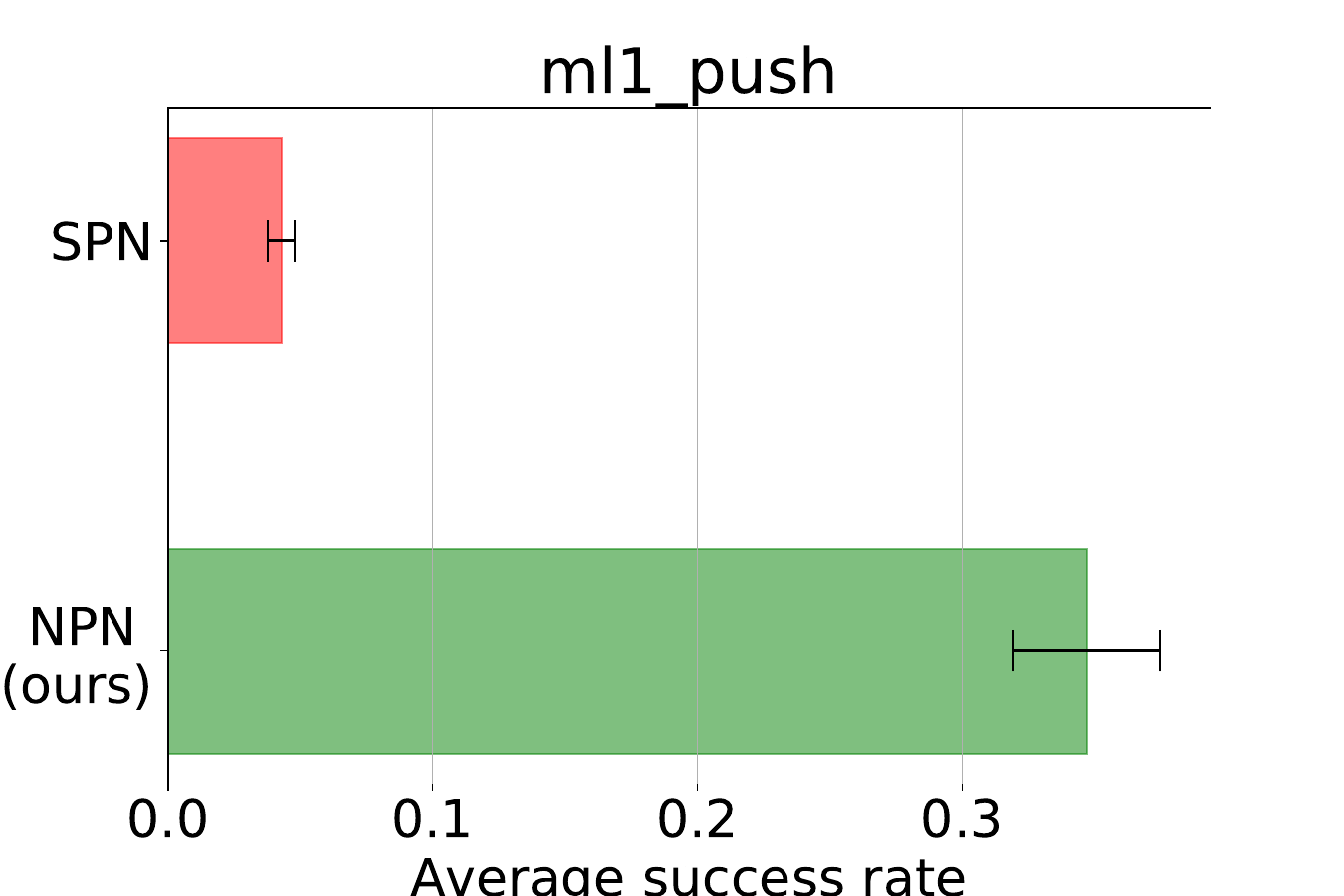}
		\caption{CAVIA, ML1}
		\label{fig:fig:res-cavia-mw-ml1-sr}
	\end{subfigure}
	\begin{subfigure}{0.22\textwidth}
		\centering
		\includegraphics[width=\textwidth]{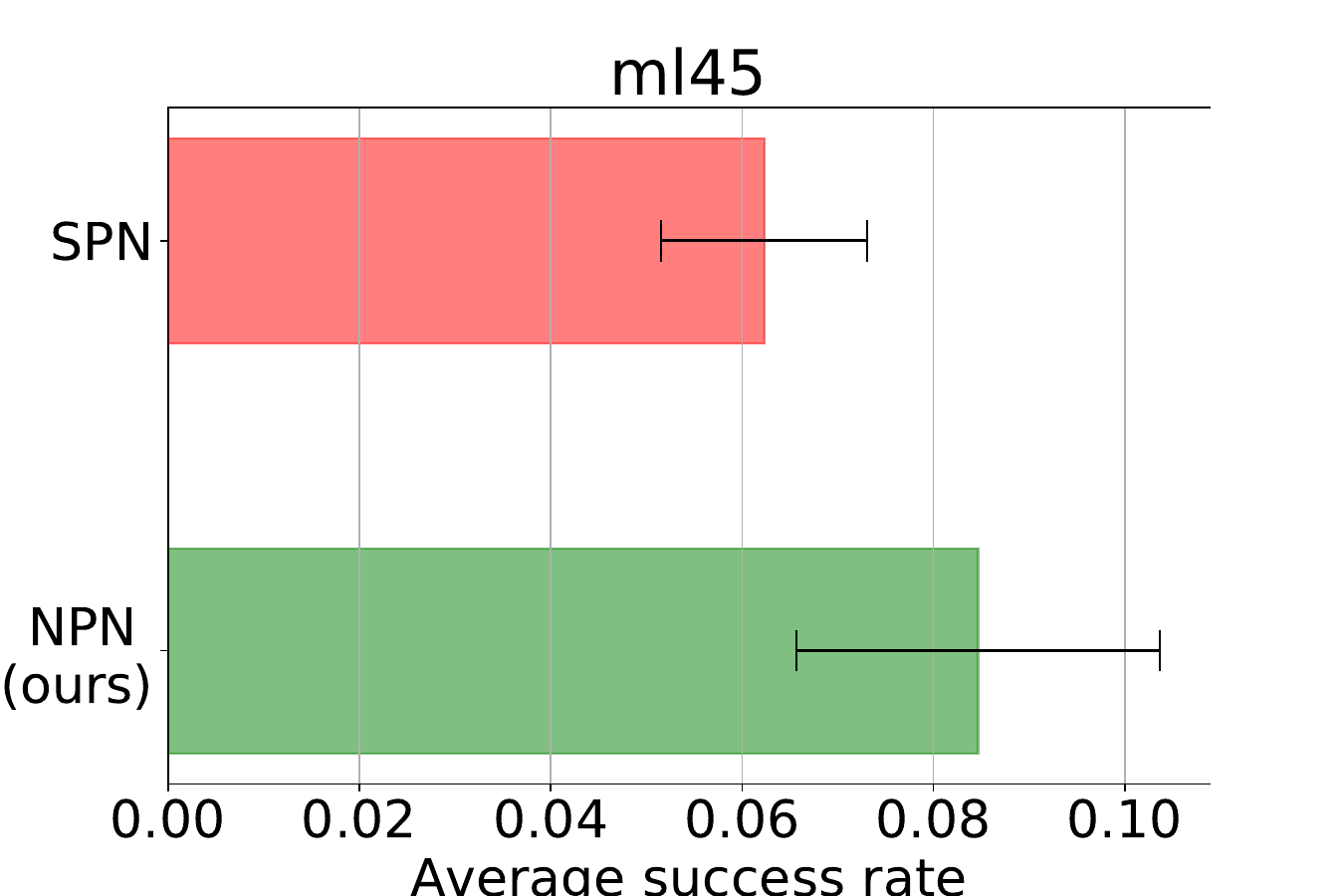}
		\caption{CAVIA, ML45}
		\label{fig:fig:res-cavia-mw-ml45-sr}
	\end{subfigure}
    \hfill
	\begin{subfigure}{0.22\textwidth}
		\centering
		\includegraphics[width=\textwidth]{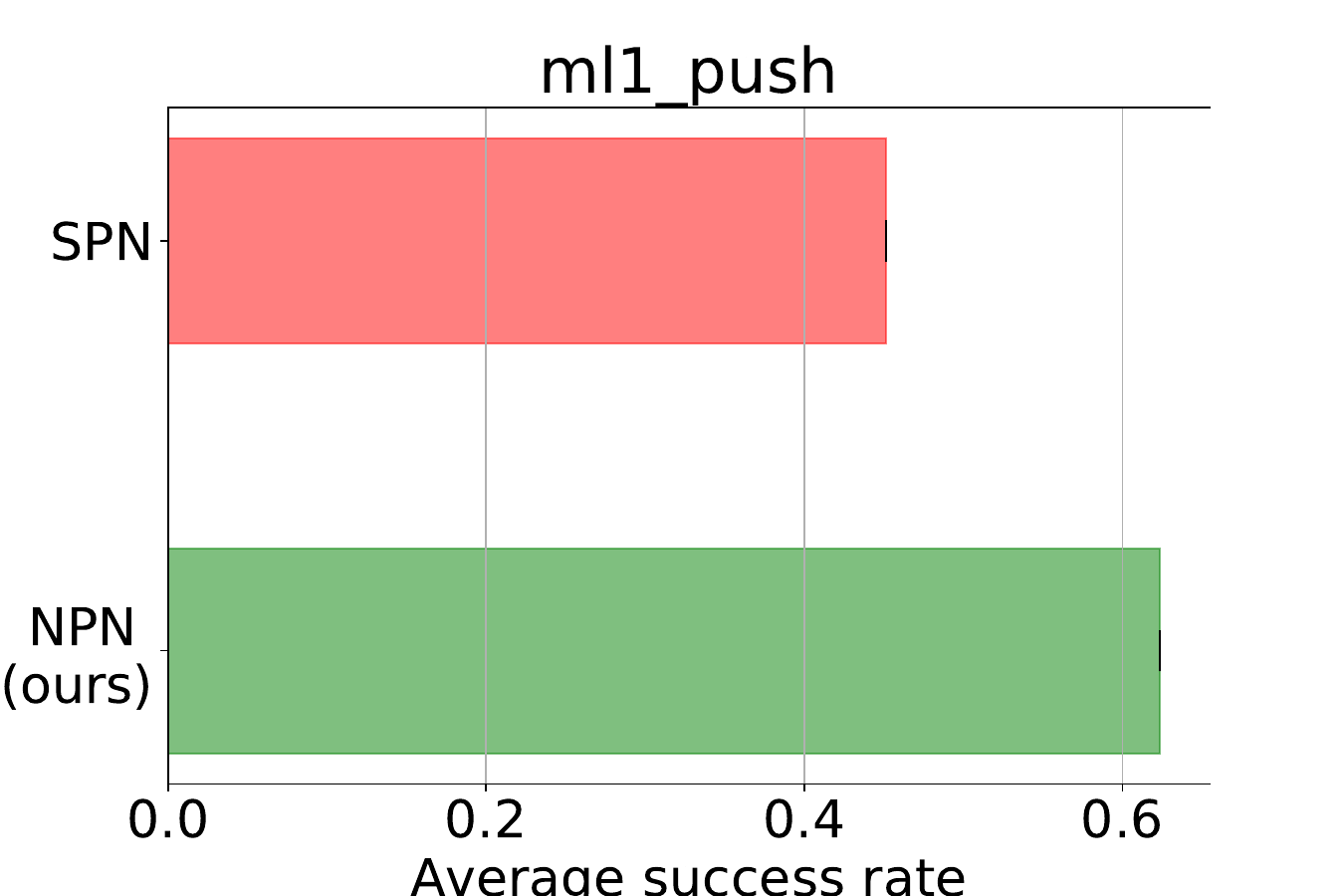}
		\caption{PEARL, ML1}
		\label{fig:fig:res-pearl-mw-ml1-sr}
	\end{subfigure}
	\hfill
	\begin{subfigure}{0.22\textwidth}
		\centering
		\includegraphics[width=\textwidth]{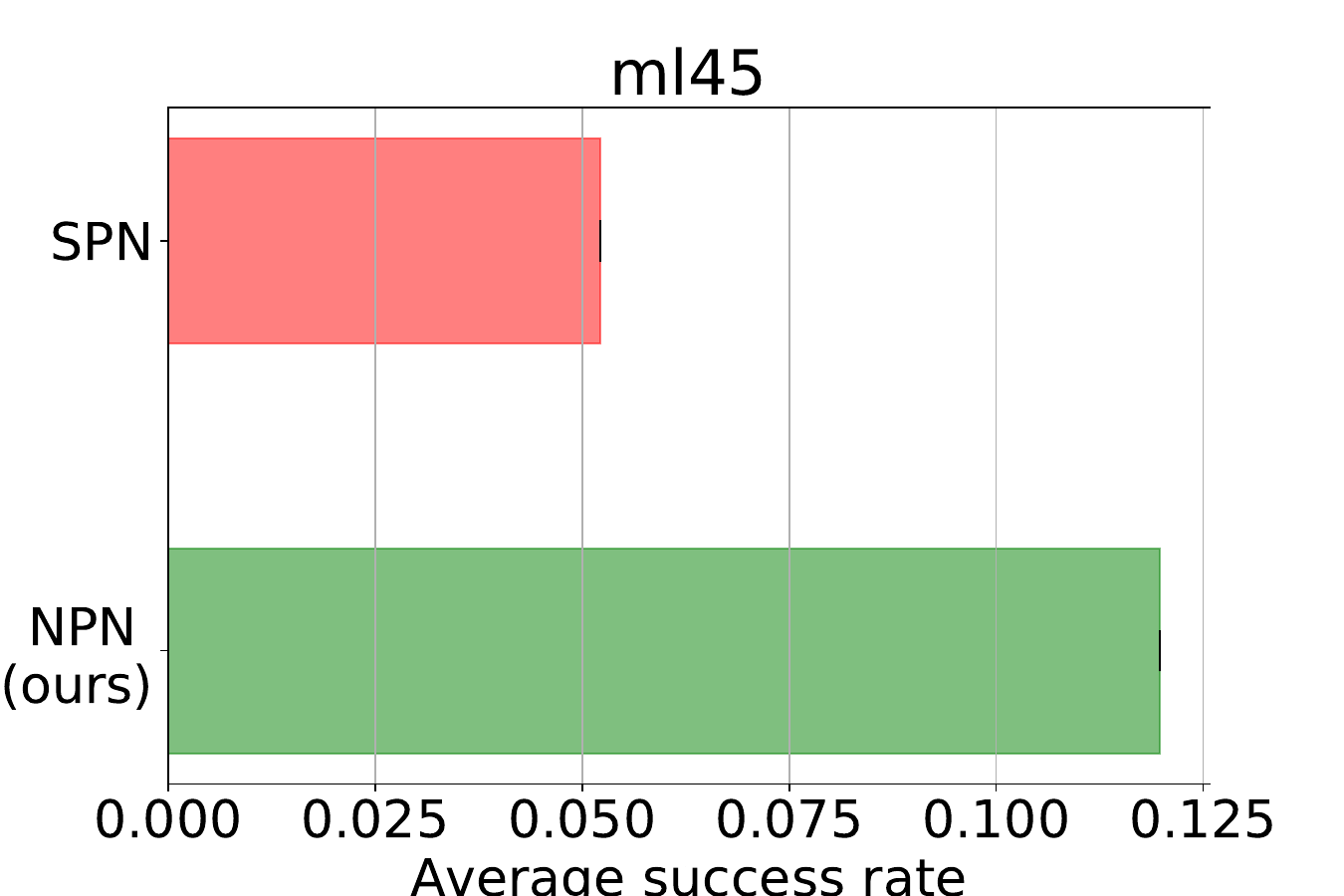}
		\caption{PEARL, ML45}
		\label{fig:res-pearl-mw-ml45-sr}
	\end{subfigure}
	\caption{Adaptation performance (across tasks, based on success rate metric) of the standard policy network (SPN) and the neuromodulated policy network (NPN) in CAVIA and PEARL. Across three seed runs, the performance was measured based on the success rate metric from the evaluation.}
	\label{fig:res-mw-success-rate}
\end{figure}

The experimental setup for CAVIA and PEARL as in \cite{zintgraf2019fast} and \cite{rakelly2019efficient} were followed. For PEARL, neuromodulation was applied only to the actor neural component. The details of the experimental setup and hyper-parameters are presented in \ref{appendix:experimental-configurations}. The performance reported are the meta-testing results of the agents in the evaluation environments after meta-training has been completed (Figures \ref{fig:res-cavia-continuous-control}, \ref{fig:res-pearl-continuous-control}, \ref{fig:res-mw-success-rate} and \ref{fig:res-cavia-discrete-control}). During meta-testing in CAVIA, the policy networks were fine-tuned for $4$ inner loop gradient steps. Lastly, depending on the evaluation environment, the metric used to judge evaluation performance was either return\footnote{return is a standard metric in RL that is computed as the sum of cumulative reward acquired by the agent.} or success rate\footnote{success rate is a metric introduced in Meta-World, having a value of $1$ if the agent has solved or is close to solving the task (i.e., if the distance between the current position of the task relevant object and goal position is smaller than some $\epsilon$ value, otherwise, it is set to $0$.}.

\subsubsection{2D Navigation Environment}
\label{subsubsection:2dnav}
The first simulations are in the 2D point navigation experiment introduced in \cite{finn2017model}. An agent is tasked with navigating to a randomly sampled goal position from a start position. A goal position is sampled from the interval [-0.5, 0.5]. The reward function is the negative squared distance between the current agent position and the goal. An observation is the agent's current 2D position while the actions are velocity commands clipped at [-0.1, 0.1]. The result of the meta-testing performance evaluation comparing both the standard policy network and neuromodulated policy network is presented in Figure \ref{fig:res-cavia-2dnavigation} for CAVIA and Figure \ref{fig:res-pearl-2dnavigation} for PEARL. The result shows that both policy networks had a relative good performance. Such optimal performance is expected from both policies as the environment is simple and the dynamic representations required for each task are not very distinct.

\subsubsection{Half-Cheetah}
\label{subsubsection:half-cheetah}
The half-cheetah is an environment based on the MuJoCo simulator \citep{todorov2012mujoco} that requires an agent to learn continuous control locomotion. We employ two standard meta-RL benchmarks using the environment as proposed in \cite{finn2017model}; (i) the direction task that requires the cheetah agent to run either forward or backward and (ii) the velocity task that requires the agent to run at a certain velocity sampled from a distribution of velocities. Although challenging (due to their high dimensional nature) in comparison to the 2D navigation task, these benchmark are still simplistic as the direction benchmark contains only two unique tasks and the velocity benchmark samples small range of velocities ($[0, 2.0)$ or $[0, 3.0)$). Therefore, the optimal policies across tasks in these benchmarks possess similar representations. The results of the experiments for both benchmarks are presented in Figures \ref{fig:res-cavia-halfcheetah-dir} and \ref{fig:res-cavia-halfcheetah-vel} for CAVIA, and Figures \ref{fig:res-pearl-halfcheetah-dir} and \ref{fig:res-pearl-halfcheetah-vel} for PEARL. Unsurprisingly, the results show comparable level of performance between the standard policy network and the neuromodulated policy network across CAVIA and PEARL. These benchmarks are of medium complexity and the optimal policy for each task is similar to others.

\subsubsection{Meta-World}
\label{subsubsection:meta-world}
The neuromodulated policy network was evaluated in a complex high-dimensional continuous control environment called meta-world \citep{yu2020meta}. In meta-world, an agent is required to manipulate a robotic arm to solve a wide range of tasks (e.g. pushing an object, pick and place objects, opening a door and more). Two instances of the benchmark ML1 and ML45 were employed. In ML1 instance, the robot is required to solve a single task that contains several parametric variations (e.g. push an object to different goal locations). The parametric variations of the selected task are used as the meta-train and meta-test tasks. ML45 is a more complex instance that contains a wide variety of tasks (each task with parametric variations). It consists of 45 distinct meta-train tasks and 5 distinct meta-test tasks. The standard policy network and neuromodulated policy network were evaluated in ML1 and ML45 instances using CAVIA and PEARL. The results\footnote{The experiments were conducted using the updated Meta-World (i.e., v2) environment containing the updated reward function.} are presented in Figures \ref{fig:res-cavia-ml1} and \ref{fig:res-cavia-ml45} for CAVIA, and Figures \ref{fig:res-pearl-ml1} and \ref{fig:res-pearl-ml45} for PEARL. In these complex benchmarks, the results show that the neuromodulated policy network outperforms the standard policy network in both CAVIA and PEARL, highlighting the advantage neuromodulation offers in complex problem setting. In addition to judging the performance based on reward, results are also presented using the success rate metric (introduced in \cite{yu2020meta} as a metric judge whether or not an agent is able to solve a task) in Figure \ref{fig:res-mw-success-rate}. The results again show that the neuromodulated policy network achieved significantly higher average success rate both in CAVIA and PEARL in comparison to the standard policy network.

\subsubsection{Configurable Tree graph (CT-graph)
Environment}
\label{subsubsection:CT-graph}
\begin{figure}[t!]
	\centering
	\begin{subfigure}{0.3\textwidth}
		\centering
		\includegraphics[width=\textwidth]{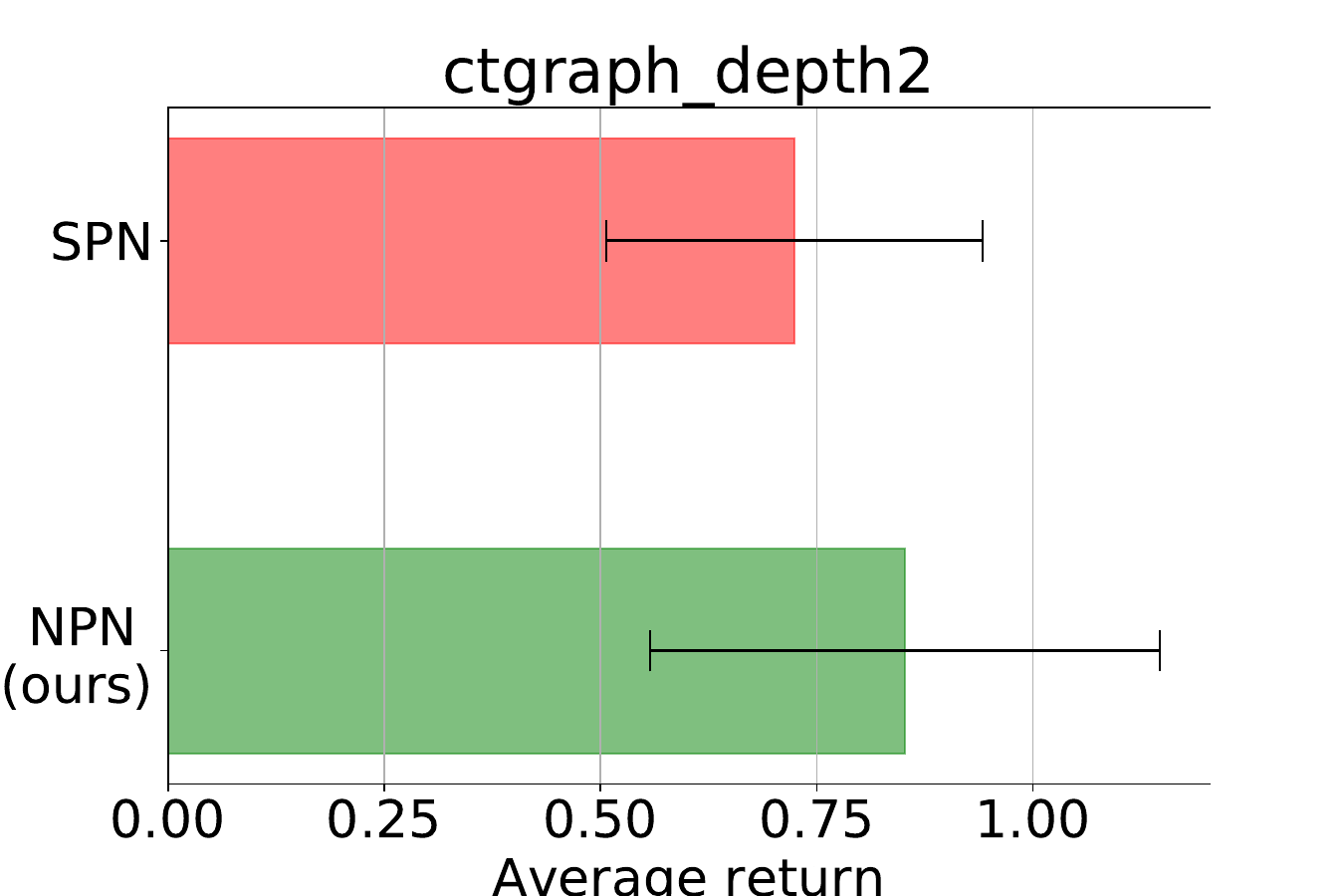}
		\caption{CT-graph ($b=2, d=2$)}
		\label{fig:res-ctgraph-depth2}
	\end{subfigure}
    \hfill
	\begin{subfigure}{0.3\textwidth}
		\centering
		\includegraphics[width=\textwidth]{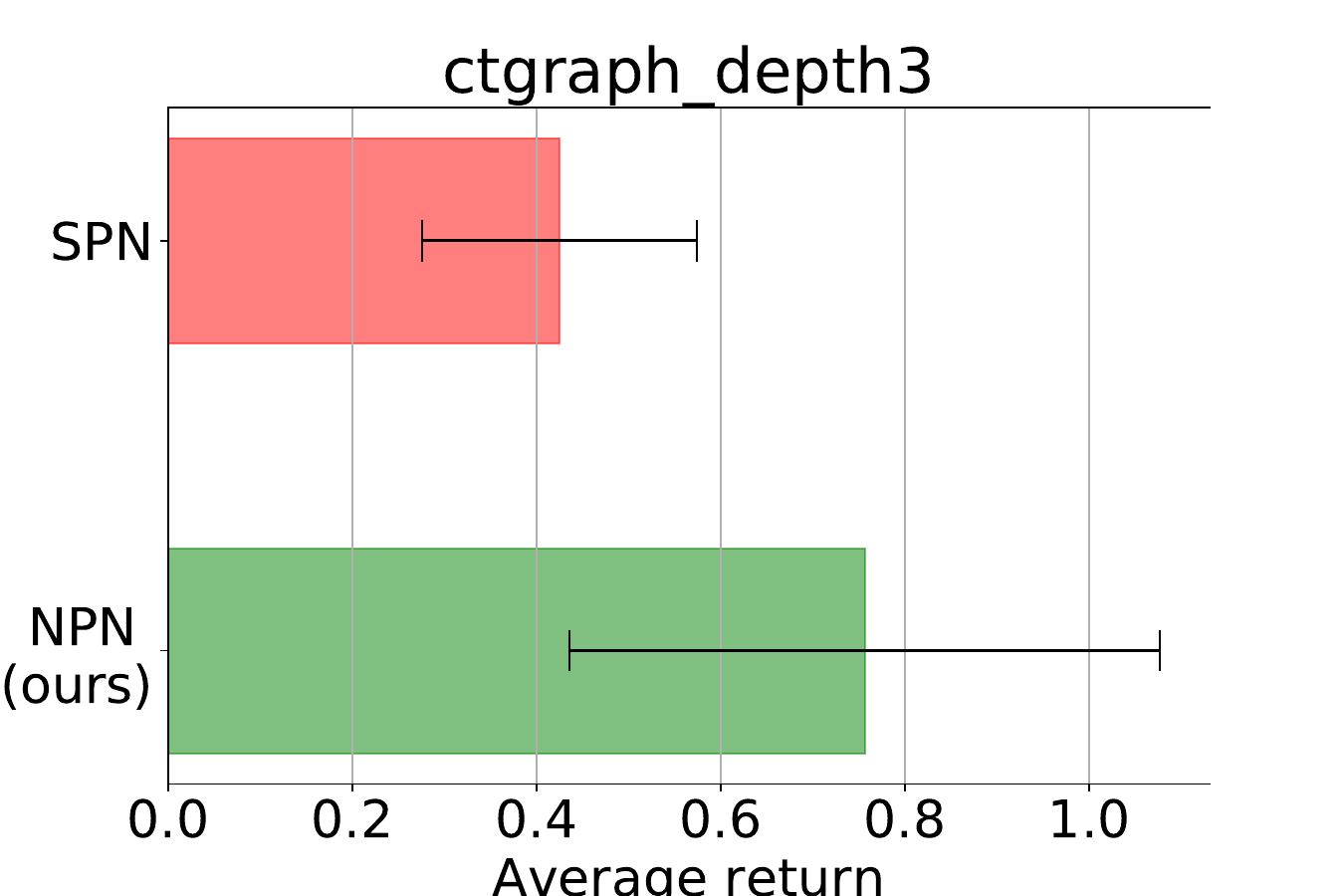}
		\caption{CT-graph ($b=2, d=3$)}
		\label{fig:res-ctgraph-depth3}
	\end{subfigure}
	\hfill
	\begin{subfigure}{0.3\textwidth}
		\centering
		\includegraphics[width=\textwidth]{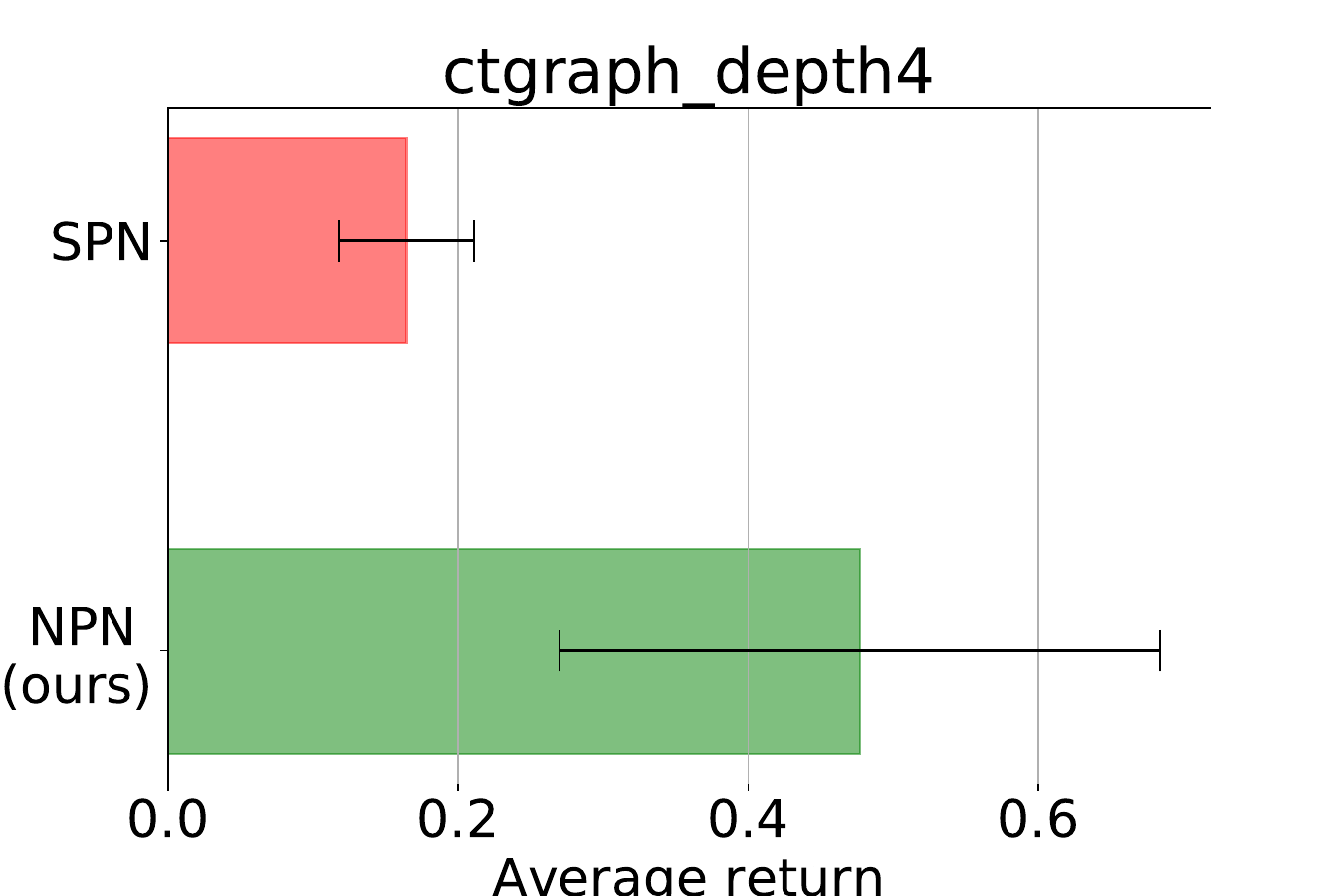}
		\caption{CT-graph ($b=2, d=4$)}
		\label{fig:res-ctgraph-depth4}
	\end{subfigure}
	\caption{Adaptation performance across tasks of the standard policy network (SPN) and the neuromodulated policy network (NPN) in three discrete control environments using CAVIA meta-RL framework. Across three seed runs, the performance was measured based on the success rate metric from the evaluation.}
	\label{fig:res-cavia-discrete-control}
\end{figure}

The CT-graph is a sparse reward discrete control graph environment with increasing complexity that is specified via parameters such as branch $b$ and depth $d$. An environment instance consists of a set of states including a start state and a number of end states. An agent is tasked with navigating to a randomly sampled end state from the start state. See Appendix \ref{appendix:CT-graph} for more details about the CT-graph. The three CT-graph instances used in this work were setup with varying depth parameter: with increasing depth, the sequence of actions grows linearly, but the search space for the policy network grows exponentially. The simplest instance has $d$ set to 2 (CT-graph depth2), and the next has $d$ set to 3 (CT-graph depth3) and the most complex instance has $d$ set to 4 (CT-graph depth4). The meta-testing results are presented in Figure \ref{fig:res-cavia-discrete-control}. The results show a significant difference in performance between standard and neuromodulated policy network. The optimal adaption performance from the neuromodulated policy network stems from the rich dynamic representations needed for adaptation as discussed in Section \ref{subsection:analysis}.

\subsection{Analysis}
\label{subsection:analysis}
In this section, we conduct analysis on the learnt representations of the standard and neuromodulated policy networks for tasks in the 2D Navigation and CT-graph environments. The policy networks trained using CAVIA was chosen for the analysis as the single neural component in CAVIA (i.e. the policy network) makes it easier to analyse in comparison to PEARL which contain multiple neural components. Furthermore, PEARL experiments were conducted only in continuous control environments (similar to the original paper), whereas CAVIA experiments covered both discrete and continuous control environments. Hence, analysis in CAVIA allowed for more coverage across benchmarks.

To measure representation similarity across task, we employ the use of the centered kernel alignment (CKA) \citep{kornblith2019similarity} similarity index, comparing per layer representations of both standard and neuromodulated policy networks across different tasks. There exist several similarity index measures such as canonical correlation analysis (CCA) \citep{morcos2018insights}, representation similarity analysis (RSA) \citep{kriegeskorte2008representational}, Hilbert-Schmidt Independence Criterion (HSIC) \citep{gretton2005measuring} and more. 

The principle behind CKA is the generation of a similarity measure between two representations by comparing the similarity structure of both representations. Each similarity structure is produced from the measure of similarity between pairwise examples or data points in a representation. Furthermore, CKA is a generalised extension of HSIC, with the inclusion of normalisation which introduces the property of isotropic scaling invariance. Describing the formulation of the CKA is outside the scope of this work and we refer readers to the original paper for detailed theoretical formulations. While the RSA similarity index measure employed in \cite{goerttler2021exploring} is a valid alternative, we chose the CKA due to its robustness to random initialization and enabling comparison within layers of the same network and comparison across networks. Furthermore, CKA has been employed previously in meta-RL setting, e.g., demonstrated in \cite{Raghu2020Rapid}.

\subsubsection{Analysis: Representation similarities for standard and neuromodulated policy networks across tasks}
\label{subsubsection:analysis-layer-output}
\begin{figure}[t!]
	\centering
	\begin{subfigure}{\textwidth}
		\centering
		\includegraphics[width=\textwidth]{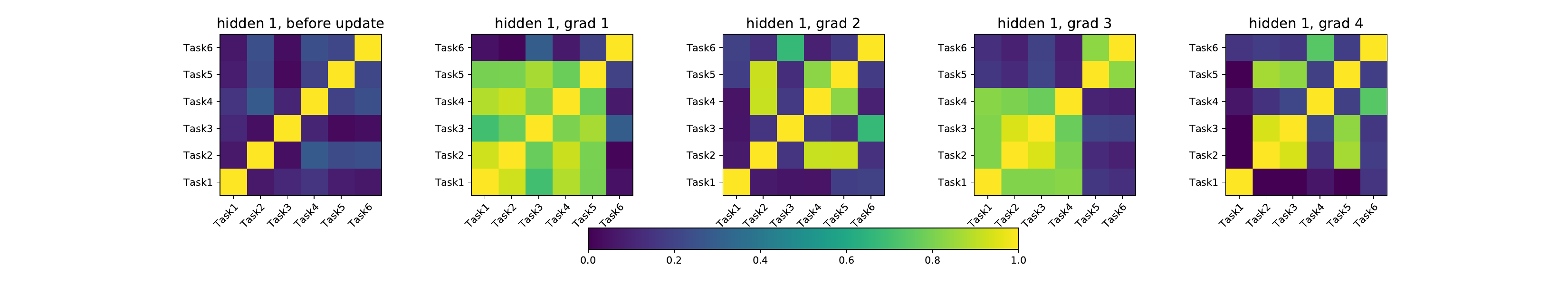}
		\caption{standard policy network, first hidden layer.}
		\label{fig:analysis-2dnav-spn-l1}
	\end{subfigure}
	\begin{subfigure}{\textwidth}
		\centering
		\includegraphics[width=\textwidth]{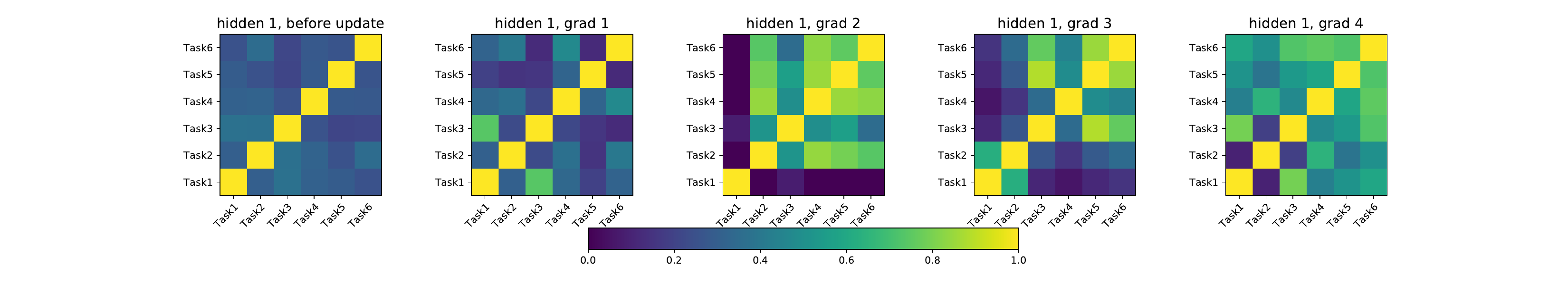}
		\caption{neuromodulated policy network, first hidden layer.}
		\label{fig:analysis-2dnav-npn-l1}
	\end{subfigure}
	\caption{Representation similarities between tasks in the 2D Navigation environment.}
	\label{fig:analysis-2dnav}
\end{figure}
\begin{figure}
	\centering
	\begin{subfigure}{\textwidth}
		\centering
		\includegraphics[width=\textwidth]{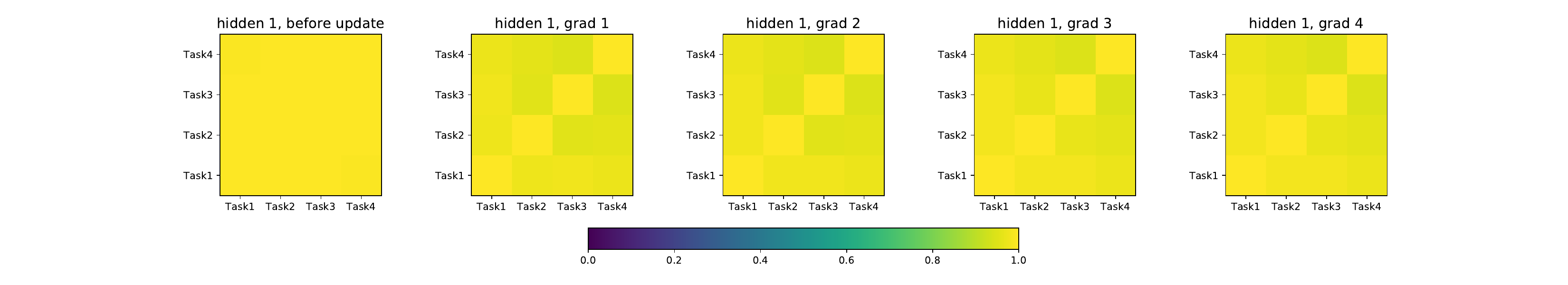}
		\caption{standard policy network, first hidden layer.}
		\label{fig:analysis-ctgraphd2-spn-l1}
	\end{subfigure}
	\begin{subfigure}{\textwidth}
		\centering
		\includegraphics[width=\textwidth]{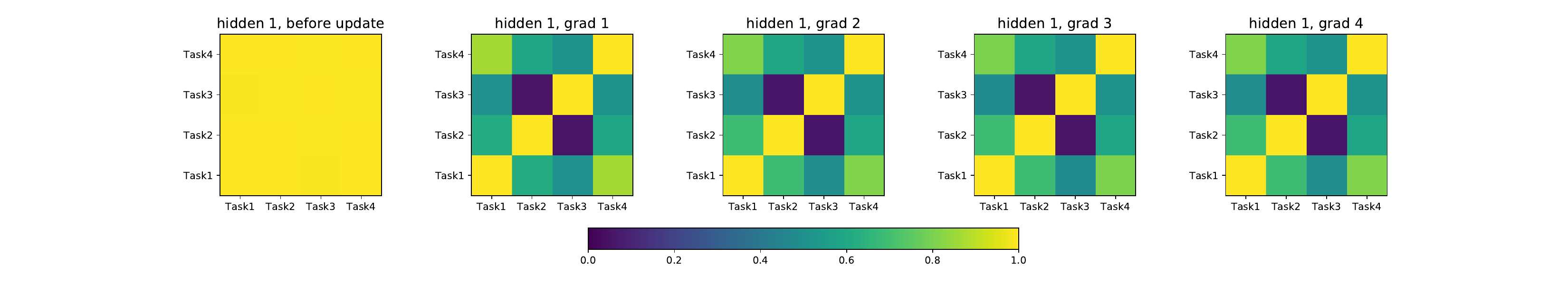}
		\caption{neuromodulated policy network, first hidden layer.}
		\label{fig:analysis-ctgraphd2-npn-l1}
	\end{subfigure}
	\caption{Representation similarities between tasks in the CT-graph depth2 environment.}
	\label{fig:analysis-ctgraphd2}
\end{figure}
The per layer output representation similarities between tasks were plotted as heat maps in Figure \ref{fig:analysis-2dnav} and \ref{fig:analysis-ctgraphd2}. Each heatmap in a row (for example \ref{fig:analysis-2dnav-spn-l1}) depict the similarity before or after few steps of gradient updates to the layer. Before any gradient updates, the representations are similar between tasks in the figure. After gradient updates, some dissimilarities between tasks begin to emerge. Additional analysis plots are presented in Appendix \ref{appendix:analysis}.

\textbf{2D Navigation}. For the simple 2D Navigation environment, the plots for the first hidden layer of the standard policy network shown in Figure \ref{fig:analysis-2dnav-spn-l1} depicts good dissimilarity between tasks, thus highlighting the fact that the learnt representations are sufficient to produce distinct task behaviours. The same is true as well for the first hidden layer of the neuromodulated policy network (see Figure \ref{fig:analysis-2dnav-npn-l1}). This further justifies why both policies obtained roughly comparable performance in this environment. The simplicity of the problem enables task distinct representations to be obtained easily. Appendix \ref{appendix:2dnav-ctgraphd2} contains the plots of the representation similarity for the second hidden layer of both policy networks.

\textbf{CT-graph}. In Figure \ref{fig:analysis-ctgraphd2-spn-l1} and \ref{fig:analysis-ctgraphd2-npn-l1}, we compare the representation similarity of the first hidden layer of the standard and neuromodulated policy networks in the CT-graph depth2 environment. We see that representations of the neuromodulated policy are more dissimilar between the tasks than those of the standard policy. Due to the complexity of the environment, the task specific representations required to solve each task are distinct from one another. Therefore, adaptation by fine-tuning the representations of a base network via few gradient steps of parameters update would require a significant jump in the solution space. Standard policy network struggles to enable such jump in the solution space. However, by incorporating neuromodulators that dynamically alters the representations, such jump becomes possible. Appendix \ref{appendix:2dnav-ctgraphd2} contains the plots of the representation similarity for the second hidden layer of both policy networks.

\subsubsection{Analysis: Representation similarities of the neuromodulatory units across tasks}
\label{subsubsection:analysis-neuromod-units}
\begin{figure}[b!]
	\centering
	\begin{subfigure}{\textwidth}
		\centering
		\includegraphics[width=\textwidth]{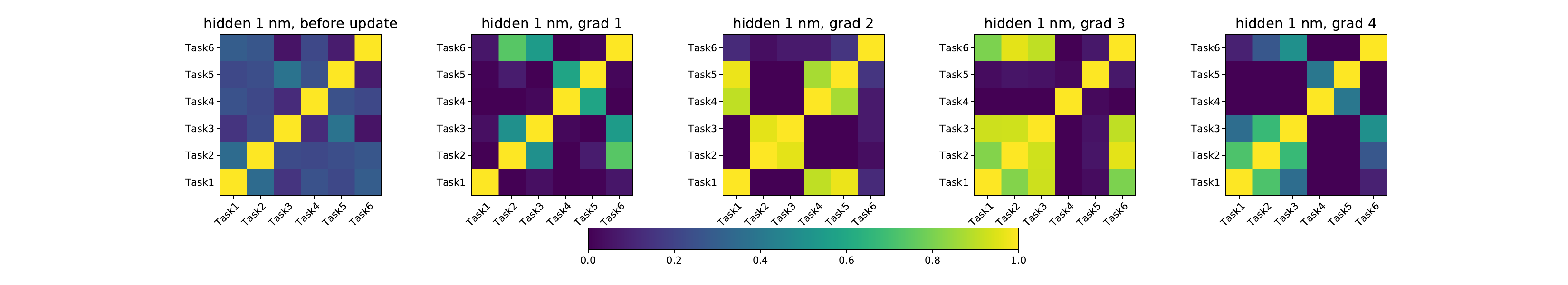}
		\caption{first hidden layer.}
		\label{fig:analysis-2dnav-nm-npn-l1}
	\end{subfigure}
	\begin{subfigure}{\textwidth}
		\centering
		\includegraphics[width=\textwidth]{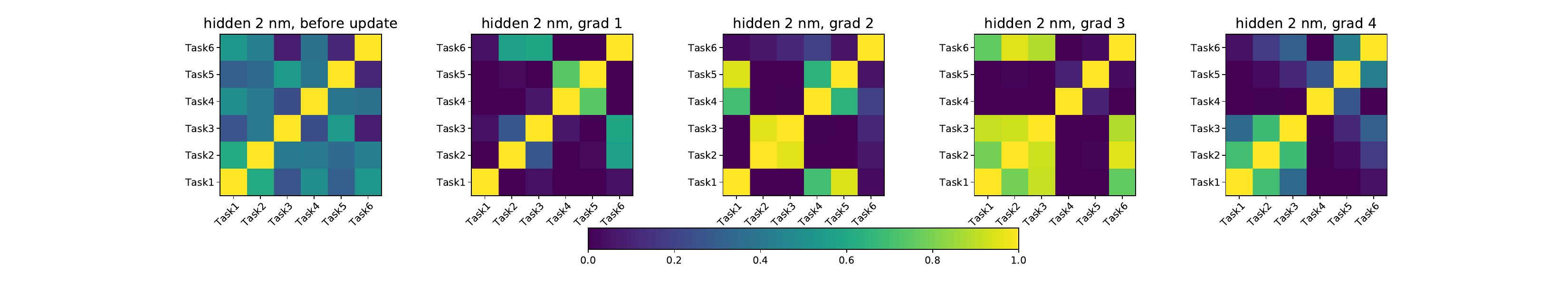}
		\caption{second hidden layer.}
		\label{fig:analysis-2dnav-nm-npn-l2}
	\end{subfigure}
	\caption{Representation similarities of neuromodulatory activities $h^m$ between tasks in the 2D Navigation environment across the first hidden layer (a), and the second hidden layer (b) of the network. Non-uniformity across heatmap plots show dissimilar representations emerge from neuromodulatory activity which helped the neuromodulated policy network to solve tasks in complex problem benchmarks.}
	\label{fig:analysis-2dnav-nm}
\end{figure}

\begin{figure}[b!]
	\centering
	\begin{subfigure}{\textwidth}
		\centering
		\includegraphics[width=\textwidth]{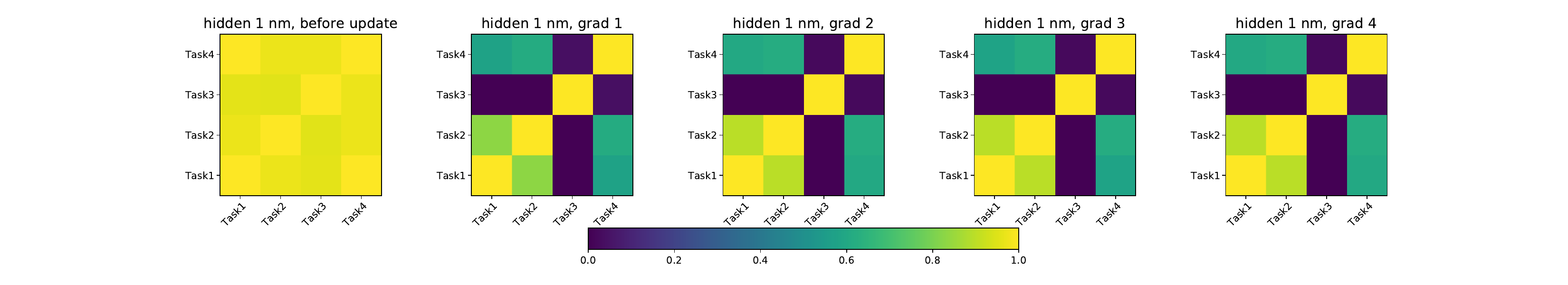}
		\caption{first hidden layer.}
		\label{fig:analysis-ctgraphd2-nm-npn-l1}
	\end{subfigure}
	\begin{subfigure}{\textwidth}
		\centering
		\includegraphics[width=\textwidth]{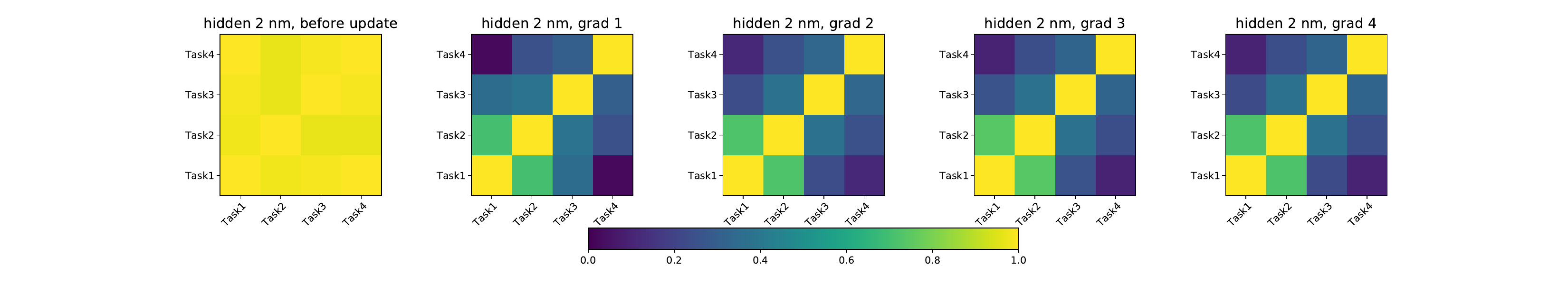}
		\caption{second hidden layer.}
		\label{fig:analysis-ctgraphd2-nm-npn-l2}
	\end{subfigure}
	\caption{Representation similarities of neuromodulatory activities $h^m$ between tasks in the CT-graph depth2 environment across the first hidden layer (a), and the second hidden layer (b) of the network. Non-uniformity across heatmap plots show dissimilar representations emerge from neuromodulatory activity which helped the neuromodulated policy network to solve tasks in complex problem benchmarks.}
	\label{fig:analysis-ctgraphd2-nm}
\end{figure}

\begin{figure}[t!]
	\centering
	\begin{subfigure}{\textwidth}
		\centering
		\includegraphics[width=\textwidth]{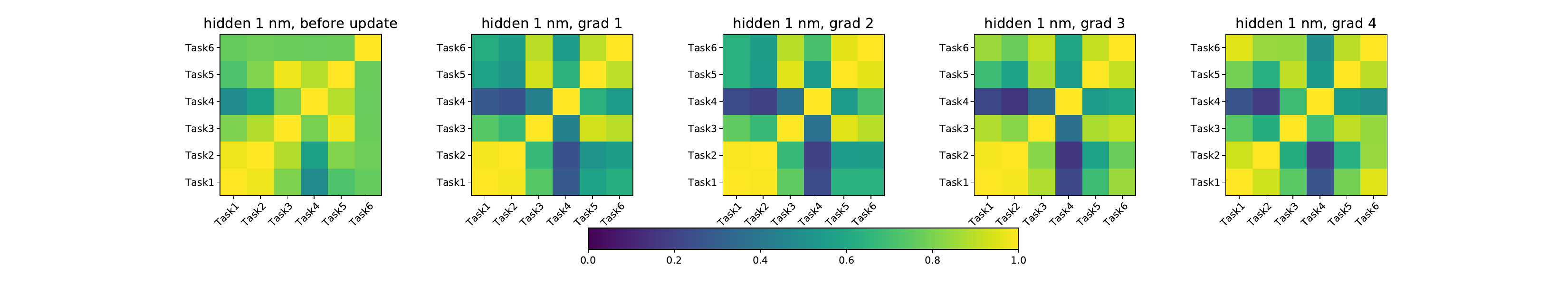}
		\caption{first hidden layer.}
		\label{fig:analysis-ml45-nm-npn-l1}
	\end{subfigure}
	\begin{subfigure}{\textwidth}
		\centering
		\includegraphics[width=\textwidth]{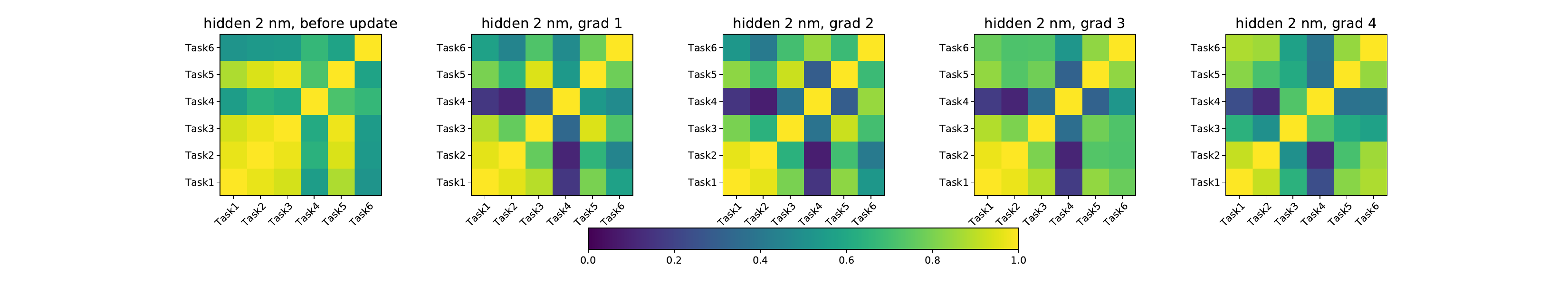}
		\caption{second hidden layer.}
		\label{fig:analysis-ml45-nm-npn-l2}
	\end{subfigure}
	\caption{Representation similarities of neuromodulatory activities $h^m$ between tasks in the Meta-World ML45 environment across the first hidden layer (a), and the second hidden layer (b) of the network. Non-uniformity across heatmap plots show dissimilar representations emerge from neuromodulatory activity which helped the neuromodulated policy network to solve tasks in complex problem benchmarks.}
	\label{fig:analysis-ml45-nm}
\end{figure}

Now we ask ourselves where the representational diversity (dissimilarity in representations across tasks) comes from. Is the neuromodulatory layer effectively contributing to rich representations as Figure \ref{fig:analysis-ctgraphd2-npn-l1} appeared to suggest? The analysis we present here shows task representation similarities measured more specifically across the neuromodulatory layers of the proposed architecture. From
Figure \ref{fig:analysis-ctgraphd2-npn-l1}, it appears that such dissimilarity is enhanced by the neuromodulatory activities in the NPN. Again the centered kernel alignment (CKA) was employed and we compare the neuromodulatory activities per layer across different tasks. Figures \ref{fig:analysis-2dnav-nm}, \ref{fig:analysis-ctgraphd2-nm} and \ref{fig:analysis-ml45-nm} present the heat map plots for the 2D navigation, CT-graph depth 2 and ML45 environments (additional plots for other environment are presented in \ref{analysis:other-repr-similarity-nm}). The non-uniformity in the heatmap plots, in contrast to those of Figure \ref{fig:analysis-ctgraphd2-spn-l1}, indicates that those layers encode diverse or dissimilar representations across task. We can therefore conclude that the neuromodulatory activities, when projected onto a layer's standard neurons, produce the desired dissimilar representations across tasks.

\subsection{Control Experiments: larger SPN, equaling the number of parameters of the NPN}
\label{subsection:control-experiments}
\begin{figure}[hb!]
	\centering
	\begin{subfigure}{0.45\textwidth}
		\centering
		\includegraphics[width=\textwidth]{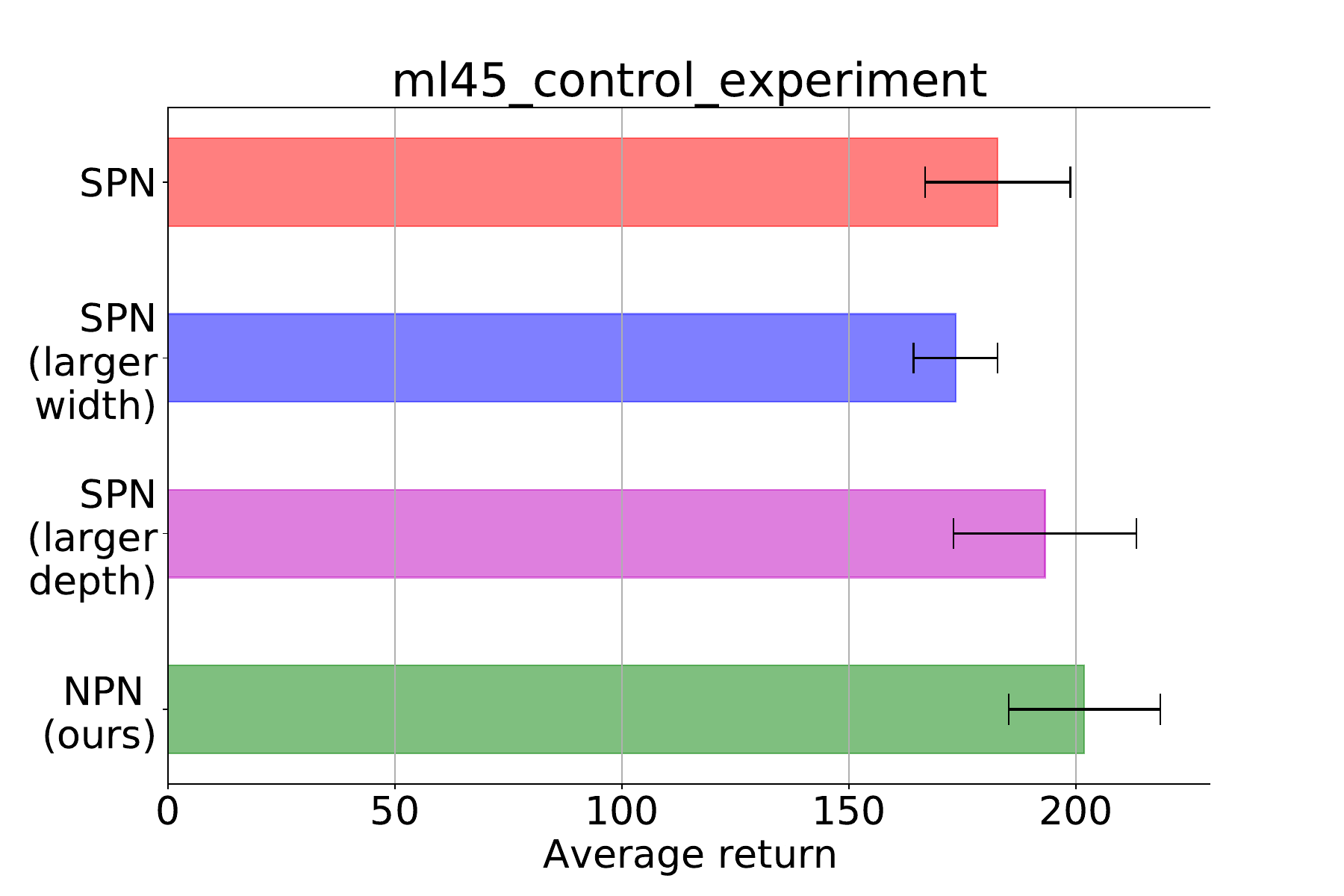}
		\caption{ML45}
		\label{fig:res-control-exp-ml45}
	\end{subfigure}
	\hfill
	\begin{subfigure}{0.45\textwidth}
		\centering
		\includegraphics[width=\textwidth]{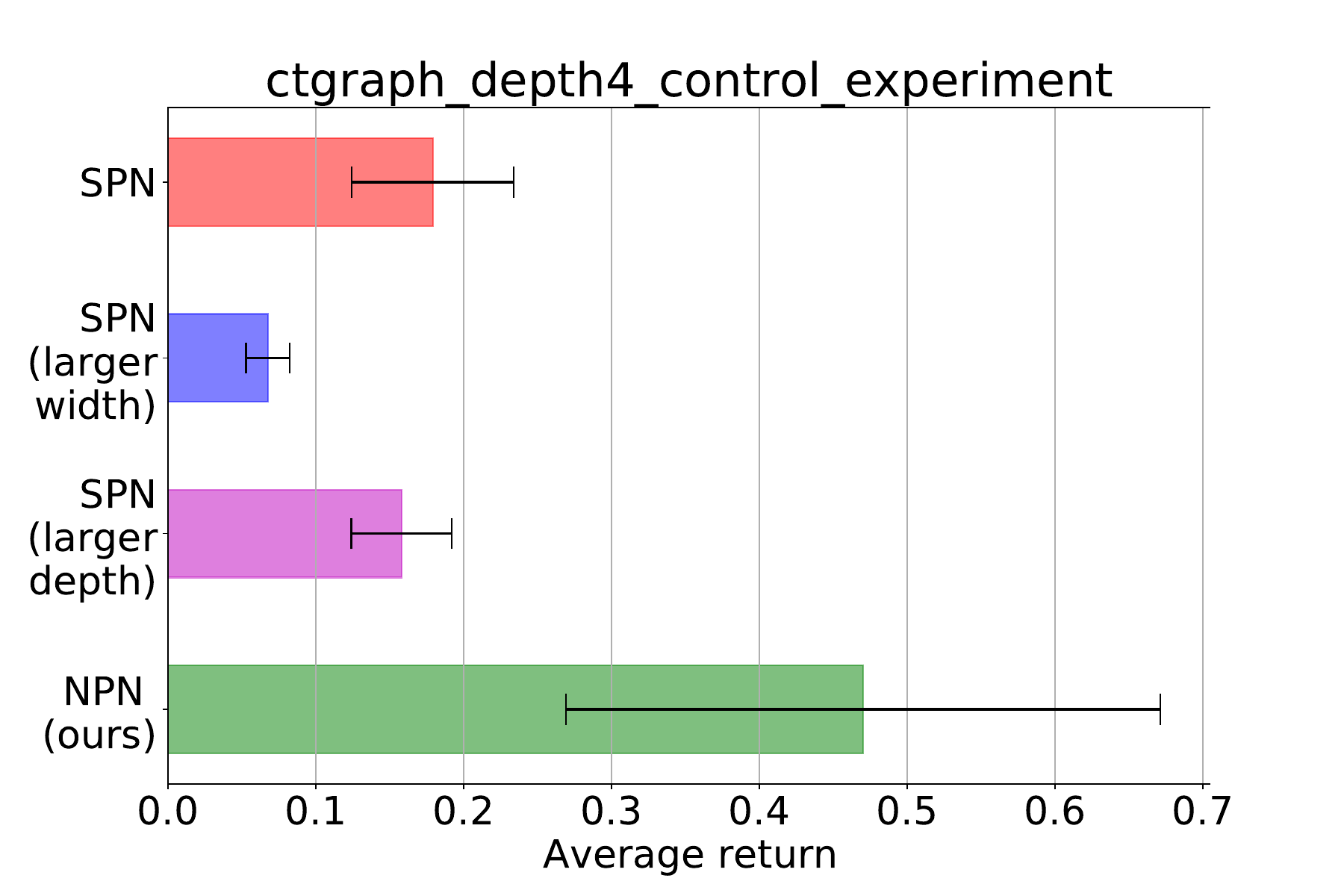}
		\caption{CT-graph depth4}
		\label{fig:res-control-exp-ctgraphd4}
	\end{subfigure}
	\caption{Control experiments. Adaptation performance of standard policy network (SPN), a larger SPN variant and neuromodulated policy network (NPN) using CAVIA. Note, the number of parameters in \emph{SPN (larger)} approximately matches that of the NPN in each environment.}
	\label{fig:res-control-exp-ctgraphd4-ml45}
\end{figure}

Since the inclusion of neuromodulators increases the number of parameters in a neuromodulated policy network, a set of control experiments were conducted in which the number of parameters in a standard policy network was configured to approximately match that of a neuromodulated policy network. This was achieved by increasing the size of each hidden layer of the standard policy network (called SPN larger width) in one experiment, and increasing the depth or number of hidden layers by 1, i.e., an additional layer (SPN larger depth) in another experiment. Using CAVIA, experiments were conducted in the CT-graph depth4 and the ML45 meta-world environments, comparing the standard policy network (i.e., the original size), its larger variants and a neuromodulated policy network. The results are presented in Figure \ref{fig:res-control-exp-ctgraphd4-ml45}. We observe from the results that the increase in the size of the policy network does not lead to match of the performance of the neuromodulated policy network.

\section{Discussions}
\label{section:discussions}
\textbf{Neuromodulation and gated recurrent networks:}
The neuromodulatory gating mechanism introduced in this work is reminiscent of the gating in recurrent/memory networks (LSTMs \citep{hochreiter1997long} and GRUs \citep{cho2014properties}). In this respect (with the observation of improved performance as a consequence of neuromodulatory gating in this work), the noteworthy performance demonstrated by meta-RL memory approaches \citep{duan2016rl, wang1611learning} could also be a consequence of such gating mechanisms\footnote{Although not the focus of this work, we ran an experiment using $RL^2$ (a memory based meta-RL method) in the ML45 environment and achieved an average meta-test success rate performance of $10\%$, which is comparable to the results obtained using neuromodulatory gating mechanism. See \ref{appendix:meta-world-perf-disccusions} for discussions about performance in meta-world, in relation to the performance reported in the original meta-world paper.}. Nonetheless, the present study aims to highlight the advantage of a simpler form of gating (i.e., neuromodulatory gating) on a MLP feedforward network, and thus could help to pinpoint the advantage of such dynamics in isolation. Furthermore, the advantage of our approach over gated recurrent variants is somewhat similar to the advantages derived from decoupling attention mechanism from recurrent models (where it was originally introduced) and applying it to MLP networks (i.e., Transformer models) \citep{vaswani2017attention}. By decoupling neuromodulatory (gating) mechanism from recurrent models and applying it to MLP models (as in our work), the advantages of faster training and better parallelization were achieved while maintaining the benefit of neuromodulatory gating. Therefore, our proposed approach is faster to train and more parallelizable in comparison to memory variants, while maintaining the advantages that neuromodulatory gating offers. Memory based approaches will still be required for problems where memory is advantageous such as sequential data processing and POMDPs.

\textbf{Task similarity measure and robust benchmarks:}
Increasing task complexity was presented in this work by moving from simple 2D point navigation environment to half-cheetah locomotion and then to the complex robotic arm setup of the Meta-World environment. Furthermore, exploiting the benefits of configurable parameters in the CT-graph environment, we were able to control the complexity in the environment. Overall, task complexity was viewed through the perspective of task similarity (i.e., environments with dissimilar task were viewed as more complex and vice versa). Despite these efforts, a precise measure of task complexity and similarity was not clearly outlined in this work and this is widely the case in meta-RL literatures. There is a need for the development of precise metrics for measuring task similarity and complexity in the field. The CT-graph with its configurable parameters allow for tasks to be mathematically defined, which is a first step towards alleviating this issue. However, a separate future research investigation would be necessary to develop explicit metrics that can be incorporated into meta-RL benchmarks.

We hypothesize that such a task similarity metric should be able to capture the precise change points in a task relative to other tasks. For example, a useful metric could be one that capture task change either as a function of change in reward, or state space, or transition function, or a combination of these factors. Most benchmarks in meta-RL have been focused on task change as reward function change. However, a more robust benchmark could include the aforementioned change points in order to further control the complexity. The CT-graph, Meta-world, and the recently developed Alchemy \citep{wang2021alchemy} environment are examples of benchmarks with early stage work in this direction, albeit implicitly. Therefore, the development of a precise measure of task similarity and complexity, as well as robust benchmarks with configurable change points (i.e., reward, state/input, and transition) would be highly beneficial to the meta-RL field.

\section{Conclusion and Future Work}
\label{section:conclusion}
This paper introduced an architectural extension of the standard meta-RL policy networks to include a neuromodulatory mechanism, investigating the beneficial effect of neuromodulation when augmenting existing meta-RL frameworks (i.e., neuromodulation as complementary tool to meta-RL rather than competing). The aim is to implement richer dynamic representations and facilitate rapid task adaptation in increasingly complex problems. The effectiveness of the proposed approach was evaluated in meta-RL setting using CAVIA and PEARL algorithms. In the experimental setup across environments of increasing complexity, the neuromodulated policy network significantly outperformed the standard policy network in complex problems while showcasing comparable performance in simpler problems. The results highlight the usefulness of neuromodulators to enable fast adaptation via rich dynamic representations in meta-RL problems. The architectural extension, although simple, presents a general framework for extending meta-RL policy networks with neuromodulators that expand their ability to encode different policies. The projected neuromodulatory activity can be designed to perform other functions apart from the one introduced in this work e.g., gating plasticity of weights, or including different neuromodulators in the same layer. The neuromodulatory extension could also be tested with a recurrent meta-RL policy, with the goal of enhancing the memory dynamics of the policy. Our analysis indicates that this framework is most suited to problems that require rapid change in optimal representations across tasks, while its advantage is reduced when tasks can be solved using similar representations.

\section*{Acknowledgment}
\label{section:acknowledgment}
This material is based upon work supported by the United States Air Force Research Laboratory (AFRL) and Defense Advanced Research Projects Agency (DARPA) under Contract No. FA8750-18-C0103. Any opinions, findings and conclusions or recommendations expressed in this material are those of the author(s) and do not necessarily reflect the views of the United States Air Force Research Laboratory (AFRL) and Defense Advanced Research Projects Agency (DARPA). The authors would like to thank Jeffrey Krichmar for useful discussion. Lastly, the authors thank the anonymous reviewers for their valuable feedback which helped to improve the paper.

\appendix
\newpage
{\Large Supplementary Material} 

\section{Experimental Configurations}
\label{appendix:experimental-configurations}

All experiments were conducted using machines containing Tesla K80 and GeForce RTX 2080 GPUs. Also note that across all experiments, the output layer in the neuromodulated policy network (in CAVIA and in PEARL) employed a regular fully connected linear layer while the preceding layers were neuromodulated fully connected layers.

\subsection{CAVIA}
\label{appendix:experimental-configurations-cavia}
Following the experimental setup of the original CAVIA paper \citep{zintgraf2019fast}, the context variables were concatenated to the input of the policy network and were reset to zero at the beginning of each task across all experiments. Also, during each training iteration, the policy was adapted using one gradient update in the inner loop as employed in \cite{zintgraf2019fast, finn2017model}. After training, the iteration with the best policy performance or the final policy at the end of training was used to conduct meta-testing evaluations to produce the final result. During meta-testing, the policy was evaluated using a number of tasks sampled from the task distribution and it was adapted (fine-tuned) for each task using $4$ inner loop gradient updates. All policy networks employed ReLU non-linearity across all experiments. 

The CAVIA experimental configurations across all environments are presented in Table \ref{tab:experimental-configurations-cavia}, with \emph{2D Nav} denoting the 2D navigation benchmark, \emph{Ch Dir and Ch Vel} denoting the Half-Cheetah direction and Half-Cheetah Velocity benchmarks, \emph{ML 1 and ML45} denoting the meta-world ML1 and ML45 benchmarks, \emph{CT d2, d3, d4} denoting the CT-graph depth2, 3 and 4 benchmarks respectively.

\begin{table}[ht!]
    \begin{tabular}{ |m{12em}|>{\centering}m{2em}|>{\centering}m{2em}|>{\centering}m{2em}|>{\centering}m{2em}|>{\centering}m{2em}|>{\centering}m{2em}|>{\centering}m{2em}|>{\centering\arraybackslash}m{2em}| }
        \hline
        \textbf{} & \textbf{2D Nav} & \textbf{Ch Dir}  & \textbf{Ch Vel} & \textbf{ML 1} & \textbf{ML 45} & \textbf{CT d2} & \textbf{CT d3} & \textbf{CT d4} \\
        \hline
        
        ~ \newline Number of iterations & 500 & 500 & 500 & 500 & 500 & 500 & 700 & 1500 \\
        \hline
        
        Number tasks per iteration \newline (meta-batch size) & 20 & 40 & 40 & 40 & 45 & 20 & 25 & 20 \\
        \hline
        
        Number inner loop grad steps \newline (for meta-training) & 1 & 1 & 1 & 1 & 1 & 1 & 1 & 1 \\
        \hline
        
        Number trajectories per task \newline (for meta-training) & 20 & 20 & 20 & 20 & 10 & 20 & 25 & 60 \\
        \hline
        
        Number inner loop grad steps \newline (for meta-testing) & 4 & 4 & 4 & 4 & 4 & 4 & 4 & 4 \\
        \hline
        
        Number trajectories per task \newline (for meta-testing) & 20 & 40 & 40 & 40 & 20 & 20 & 40 & 100 \\
        \hline\hline
        
        ~ \newline \textbf{Policy network specification} &  &  &  &  &  &  &  &  \\
        \hline
        
        ~ \newline Number of context parameters & 5 & 50 & 50 & 50 & 100 & 5 & 10 & 20 \\
        \hline
        
        ~ \newline Number of hidden layers & 2 & 2 & 2 & 2 & 2 & 2 & 2 & 2 \\
        \hline
        
        ~ \newline Hidden layer size & 100 & 200 & 200 & 200 & 200 & 200 & 300 & 600 \\
        \hline
        
        Neuromodulator size (\emph{for neuromodulated policy only}) & 4 & 32 & 32 & 32 & 32 & 8 & 16 & 32 \\
        \hline
    \end{tabular}
    \caption{CAVIA experimental configurations.}
    \label{tab:experimental-configurations-cavia}
\end{table}

Across all experiments in CAVIA, Trust Region Policy Optimization (TRPO) \citep{Schulmanetal_ICML2015} was employed as the outer loop update algorithm. Vanilla policy gradient \citep{williams1992simple} with generalized advantage estimation (GAE) \citep{Schulmanetal_ICLR2016} was employed as the inner loop update algorithm with learning rate of $0.5$ for the 2D navigation and the CT-graph experiments, and $10.0$ for half-cheetah and meta-world experiments. Both the inner and outer loop training employed a linear feature baseline introduced in \cite{duan2016benchmarking}. The hyperparameters for TRPO are presented in Table \ref{tab:trpo-hyperparameters}. Furthermore, finite-differences was employed to compute the Hessian-vector product for TRPO in order to avoid computing third-order derivatives as highlighted in \cite{finn2017model}. During sampling of data for each task in environments, multiprocessing was employed using $4$ workers.

\begin{table}[ht!]
    \centering
    \begin{tabular}{ |l|l| }
     \hline
     \textbf{Name} & \textbf{Value} \\ 
     \hline
     maximum KL-divergence & $1 \times 10^{-2}$ \\ 
     \hline
     number of conjugate gradient iterations & $10$ \\ 
     \hline
     conjugate gradient damping & $1 \times 10^{-5}$ \\ 
     \hline
     maximum number of line search iterations & $15$ \\
     \hline
     backtrack ratio for line search iterations & $0.8$ \\
     \hline
    \end{tabular}
    \caption{TRPO hyperparameters}
    \label{tab:trpo-hyperparameters}
\end{table}

\subsection{PEARL}
\label{appendix:experimental-configurations-pearl}
Similar to CAVIA, the original experimental configurations in PEARL were followed for the half-cheetah benchmarks. Also, most of the configurations of PEARL in the original meta-world experiments were followed. 

Across all PEARL experiments in this work, the learning rate across all neural components (policy, Q, value and context networks) was set to 3e-4, with KL penalty (KL lambda) set to 0.1. Furthermore, for experiments that involved the use of neuromodulation, the neuromodulator was employed only in the policy (actor) neural component. Table \ref{tab:experimental-configurations-pearl} highlights some of the PEARL configurations across the evaluation environments.

\begin{table}[ht!]
    \begin{tabular}{ |m{16em}|>{\centering}m{3em}|>{\centering}m{3em}|>{\centering}m{3em}|>{\centering}m{3em}|>{\centering\arraybackslash}m{3em}| }
        \hline
        ~ \newline \textbf{} & \textbf{2D Nav} & ~ \newline \textbf{Ch Dir}  & ~ \newline \textbf{Ch Vel} & ~ \newline \textbf{ML 1} & ~ \newline \textbf{ML 45} \\
        \hline
        
        ~ \newline Number of iterations & 500 & 500 & 500 & 1000 & 1000 \\
        \hline

        ~ \newline Number of train task & 40 & 2 & 100 & 50 & 225 \\
        \hline

        ~ \newline Number of test task & 40 & 2 & 30 & 50 & 25 \\
        \hline

        ~ \newline Number of initial steps & 1000 & 2000 & 2000 & 4000 & 4000 \\
        \hline

        ~ \newline Number of steps prior & 400 & 1000 & 400 & 750 & 750 \\
        \hline

        ~ \newline Number of steps posterior & 0 & 0 & 0 & 0 & 0 \\
        \hline

        ~ \newline Number of extra posterior steps & 600 & 1000 & 600 & 750 & 750 \\
        \hline

        ~ \newline Reward scale & 5 & 5 & 5 & 10 & 5 \\
        \hline\hline
        
        ~ \newline \textbf{Policy network specification} &  &  &  &  & \\
        \hline
        
        ~ \newline Context vector size & 5 & 5 & 5 & 7 & 7 \\
        \hline

        Network size \newline (\emph{policy, Q and value networks}) & 300 & 300 & 300 & 300 & 300 \\
        \hline
        
        ~ \newline Inference (context) network size & 200 & 200 & 200 & 200 & 200 \\
        \hline
        
        Number of hidden layers (\emph{policy, Q, value and context networks}) & 3 & 3 & 3 & 3 & 3 \\
        \hline
        
        Neuromodulator size \newline (\emph{for neuromodulated policy only}) & 4 & 32 & 32 & 32 & 32 \\
        \hline
    \end{tabular}
    \caption{PEARL experimental configurations.}
    \label{tab:experimental-configurations-pearl}
\end{table}

\section{CT-graph}
\label{appendix:CT-graph}
\begin{figure}[t!]
	\centering
	\includegraphics[width=0.7\textwidth]{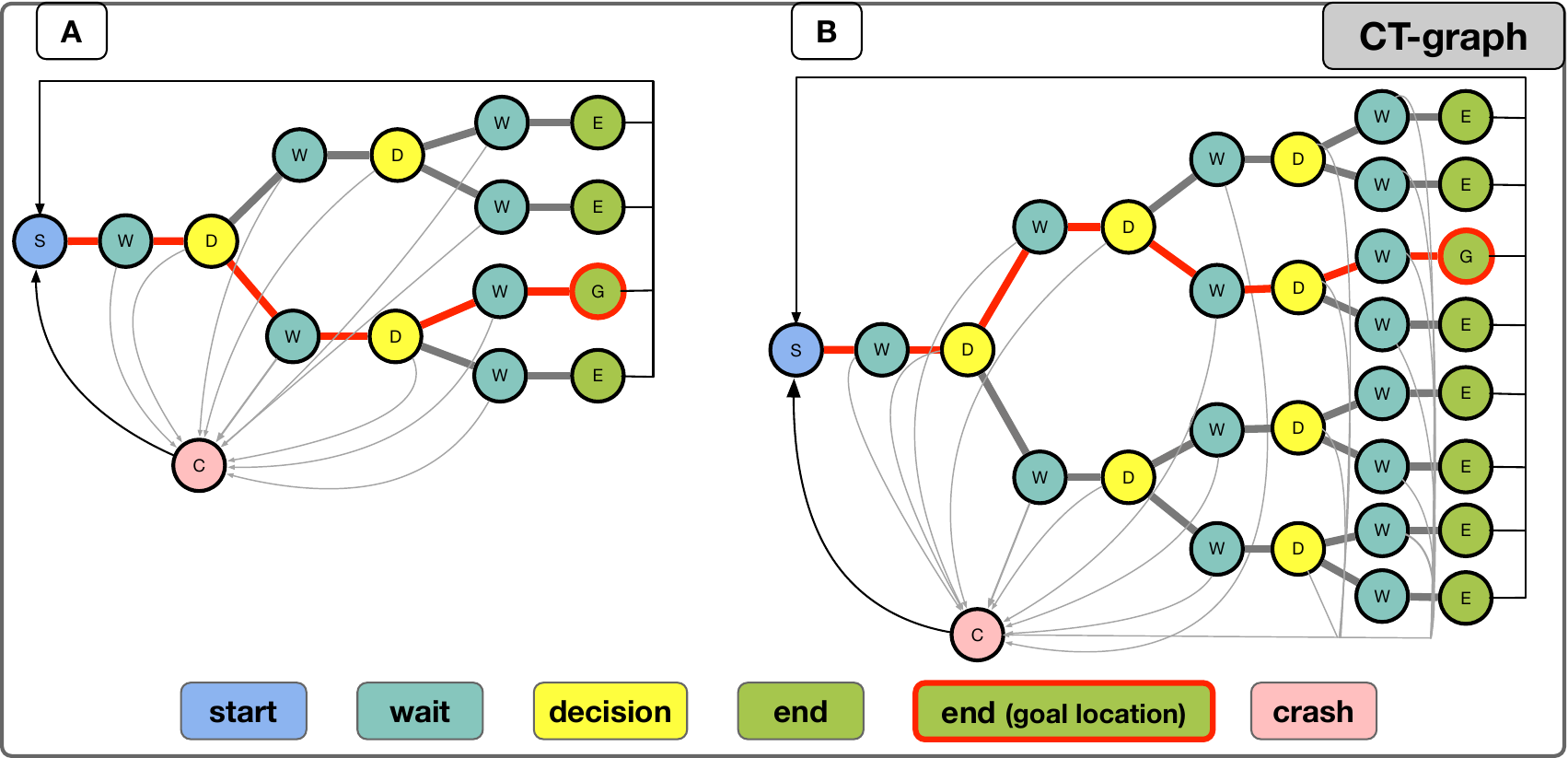}
	\caption{(A) CT-graph depth2 and (B) CT-graph depth3. The coloured legends represent the various state types in the environment.}
	\label{fig:CT-graph-env}
\end{figure}

Each environment instance in the CT-graph is composed of a start state, a crash state, a number of wait states, decision states, end states and a goal state (one of the end states designated as the goal). A wait state is found between decision states (the tree graph splits at decision states). The wait state requires an agent to take a wait (forward) action, a decision state requires an agent to take one of the decision (turn) actions. Any decision action at a wait state, or wait action at a decision state leads to a crash where the agent is punished with a negative reward of -0.01 and returns to the start. When an agent navigates to the correct end state (the goal location), it receives a positive reward of 1.0. Otherwise, the agent receives a reward of 0.0 at every time step. An episode is terminated either at a crash state or when the agent navigates to any end state. The observations are 1D vector (with full observability of each state) whose length depends on the environment instance configuration.

The environment's complexity is defined via a number of configuration parameters that is used to specify the graph size (using branch $b$ and $d$), sequence length, reward function, and level of state observability. The three CT-graph instances used in this work were setup with varying depth parameter. The simplest instance has $d$ set to 2 (CT-graph depth2), and the next has $d$ set to 3 (CT-graph depth3) and the most complex instance has $d$ set to 4 (CT-graph depth4). Figure \ref{fig:CT-graph-env} depicts a graphical view of CT-graph depth2 and 3.

\section{Implementation}
\label{appendix:implementation}
A code snippet demonstrating the extension of the fully connected layer with neuromodulation is presented below using PyTorch code style.
{\small \begin{spverbatim}
class NMLinear(Module):
  def __init__(self, in_features, out_features, nm_features, bias=True, gate=None):
    super(NMLinear, self).__init__()
    self.in_features = in_features
    self.out_features = out_features
    self.nm_features = nm_features
    self.std = nn.Linear(in_features, out_features, bias=bias)
    self.in_nm = nn.Linear(in_features, nm_features, bias=bias)
    self.out_nm =nn.Linear(nm_features, out_features, nm_features)
    self.in_nm_act = F.relu
    self.out_nm_act = torch.tanh
    self.gate = gate

  def forward(self, data, params=None):
    output = self.std(data)
    mod_features = self.in_nm_act(self.in_nm(data))
    projected_mod_features = self.out_nm_act(self.out_nm(mod_features))
    if self.gate == 'strict':
        projected_mod_features = torch.sign(projected_mod_features)
        projected_mod_features[projected_mod_features == 0.] = 1.
    output *= projected_mod_features
    return output
\end{spverbatim}}

The full implementation (including experimental setup and test scripts) is open sourced at \url{https://github.com/dlpbc/nm-metarl}. The codebase is an extension of the original CAVIA and PEARL open source (MIT license) implementations that can be found at \url{https://github.com/lmzintgraf/cavia} and \url{https://github.com/katerakelly/oyster} respectively.

\section{Additional Experiments}
\label{appendix:additional-experiments}

\subsection{CARLA Environment}
\label{appendix:carla-environment}
\begin{figure}[ht!]
	\centering
	\begin{subfigure}{0.45\textwidth}
		\centering
		\includegraphics[width=\textwidth]{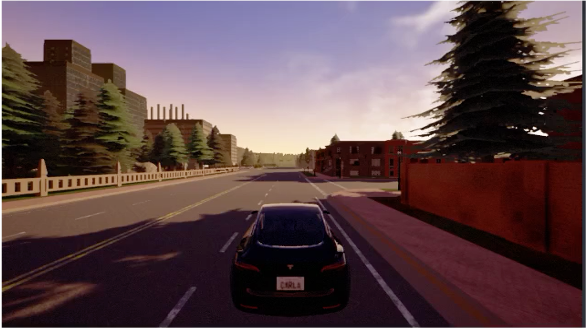}
		\caption{A snapshot of the CARLA environment. \emph{Note, the view of environment in this figure is for the purpose of presentation}. The actual environment instance used in this work is configured as a first person view with a smaller observation (image) size.}
		\label{fig:carla-env}
	\end{subfigure}
    \hfill
	\begin{subfigure}{0.45\textwidth}
		\centering
		\includegraphics[width=\textwidth]{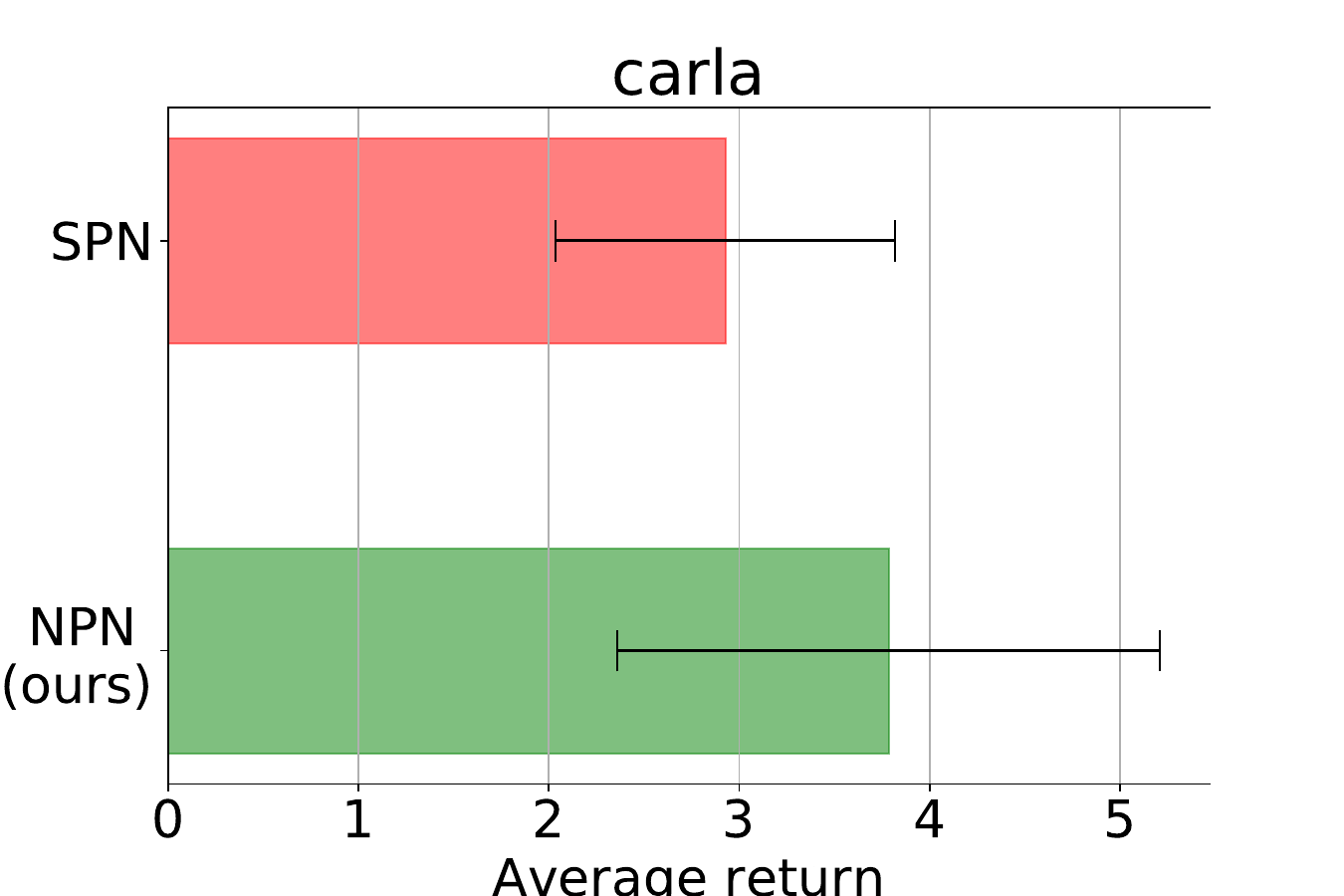}
		\caption{Adaptation performance of standard policy network (SPN) and neuromodulated policy network (NPN) in the CARLA environment.}
		\label{fig:res-carla}
	\end{subfigure}
	\caption{CARLA environment and results}
	\label{fig:carla-env-result}
\end{figure}

Additional experiments were conducted in an autonomous driving environment called CARLA \citep{Dosovitskiy17} to provide preliminary evidence on whether the method scales to complex RGB input distributions such as those in autonomous driving. Given the limited nature of these experiments and the limited analysis, they are not included in the main paper, but provide additional validation on the robustness of the proposed approach. CARLA (see Figure \ref{fig:carla-env}), is an open source experimentation platform for autonomous driving research. It contains a host of configuration parameters that is used to specify an environment instance (for example, weather). MACAD \citep{palanisamy2020multi}, a wrapper on top of CARLA with OpenAI gym interface, was employed to run the experiments. In this work, the environment was configured to use RGB observations (images of size 64x64x3), $9$ discrete actions (coast, turn left, turn right, forward, brake, forward left, forward right, brake left, and brake right), and a clear (sunny) noon weather. 

The agent (vehicle) is presented with a goal of navigating from a start position to an end position. The start and end points are randomly set from a pre-defined list of co-ordinates. We setup two distinct tasks in the environment - drive aggressively and drive passively - defined by reward functions, that can be sampled from a uniform task distribution. Although the tasks are quite similar, they are challenging due to the domain of the problem (learning to drive) and the RGB pixel observations from the environment. Therefore, it is a suitable environment to further scale up meta-RL algorithms. 

Each experiment processes the environment's observations through a variational autoencoder (VAE) \citep{kingma2013auto, rezende2014stochastic} that was pre-trained using samples collected from taking random actions in the environment. Using CAVIA, the latent features from the VAE were concatenated with the context parameters and then passed as input to the policy network. Only the policy network was updated during the meta-training and testing, while the VAE was kept fixed.

Due to the computational load of the environment, both the standard and the neuromodulated policy network were evaluated for $300$ iterations, with $4$ sampled tasks per iteration and context parameter size of $10$. For each task, $2$ episodes are collected before and after one step of inner loop gradient update. The results are presented in Figure \ref{fig:res-carla}, with the neuromodulated policy network showing an advantage over the standard policy network. In general, the results show promise towards scaling meta-RL algorithms to even more challenging problem domains.

\section{Discussions on Meta-World Performance}
\label{appendix:meta-world-perf-disccusions}
The performance difference between the baselines in the original Meta-World paper and the present submission is due to an update of the reward function in the recent version of the Meta-World environment\footnote{https://github.com/rlworkgroup/metaworld/pull/312}\footnote{https://github.com/rlworkgroup/metaworld/commit/ffe2c}. Issues about the reward function of the originally released Meta-World was reported and discussed in \url{https://github.com/rlworkgroup/metaworld/issues/226}. This led the environment developers to rewriting many of the reward functions in the environment that are now part of the current version (informally referred to as v2 environments). The results reported in the original Meta-World paper used the old (and now replaced) reward function, while the results in the present submission are based on a recent version cited above. It is nevertheless possible that the baseline results could be improved with better hyperparameter tuning, although the same is true for the novel approach that we propose. As we aim to observe performance differences between the neuromodulated meta-RL and the standard meta-RL, we did not perform hyperparameter search and tuning.

\section{Analysis Plots}
\label{appendix:analysis}
This section presents additional analysis plots of the representation similarity across tasks for the standard and neuromodulated policy networks in the various evaluation environments employed in this work. The additional plots further highlights the usefulness of neuromodulation to facilitate efficient (distinct) representations across tasks in problems of increasing complexity as earlier showcased in Section \ref{subsection:analysis}.

\subsection{Second hidden layer: 2D Navigation and CT-graph depth2 environments}
\label{appendix:2dnav-ctgraphd2}

\begin{figure}[ht!]
	\centering
	\begin{subfigure}{\textwidth}
		\centering
		\includegraphics[width=0.9\textwidth]{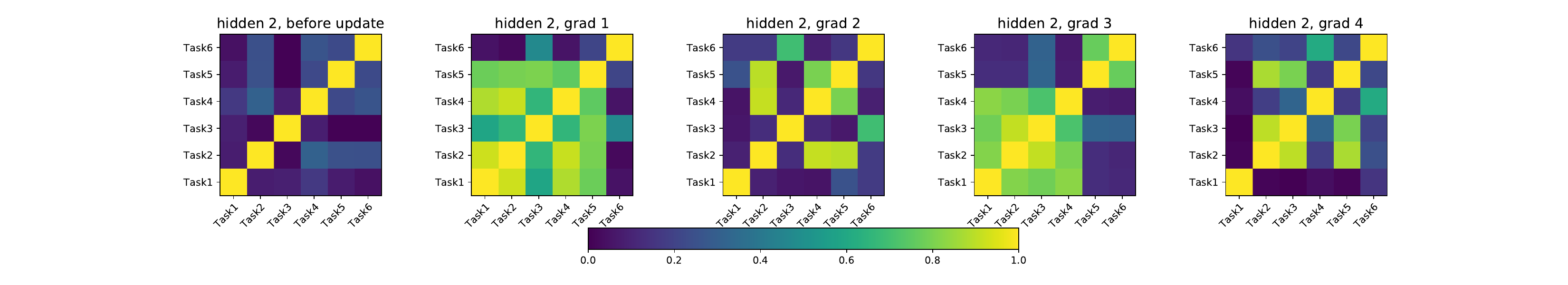}
		\caption{standard policy network, second hidden layer.}
		\label{fig:analysis-2dnav-spn-l2}
	\end{subfigure}
	\begin{subfigure}{\textwidth}
		\centering
		\includegraphics[width=0.9\textwidth]{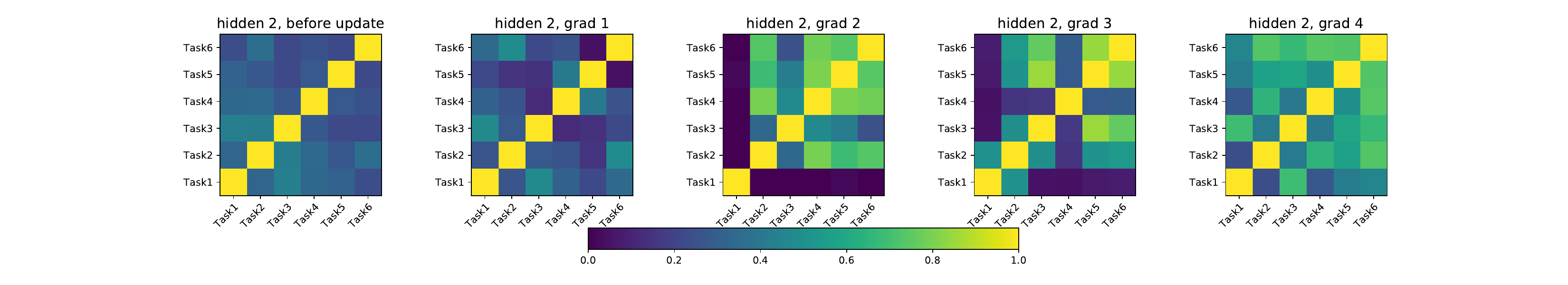}
		\caption{neuromodulated policy network, second hidden layer.}
		\label{fig:analysis-2dnav-npn-l2}
	\end{subfigure}
	\caption{Representation similarities between tasks in the 2D Navigation environment.}
	\label{fig:analysis-2dnav-l2}
\end{figure}
\begin{figure}[ht!]
	\centering
	\begin{subfigure}{\textwidth}
		\centering
		\includegraphics[width=0.9\textwidth]{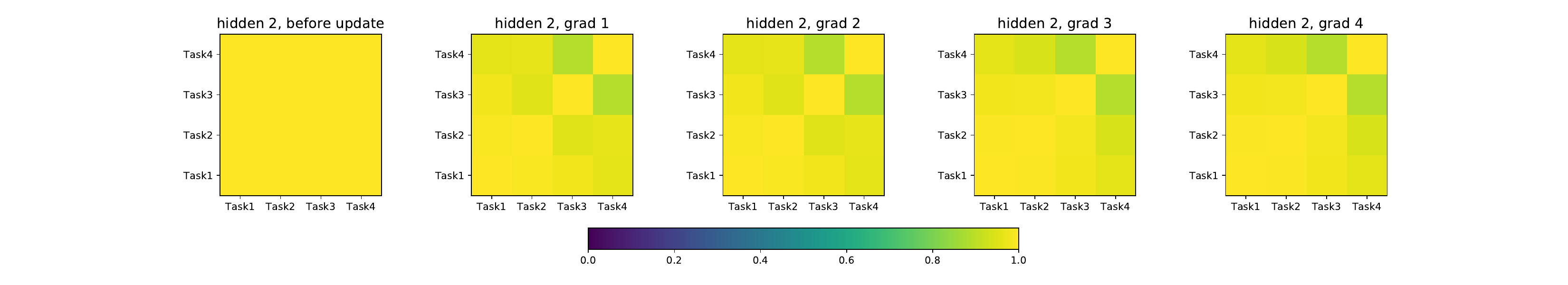}
		\caption{standard policy network, second hidden layer.}
		\label{fig:analysis-ctgraphd2-spn-l2}
	\end{subfigure}
	\begin{subfigure}{\textwidth}
		\centering
		\includegraphics[width=0.9\textwidth]{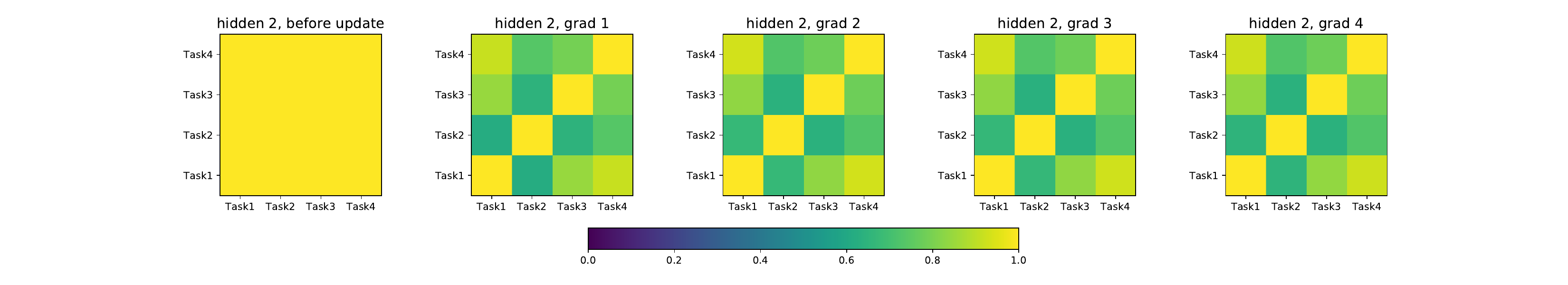}
		\caption{neuromodulated policy network, second hidden layer.}
		\label{fig:analysis-ctgraphd2-npn-l2}
	\end{subfigure}
	\caption{Representation similarities between tasks in the CT-graph-depth2 environment.}
	\label{fig:analysis-ctgraphd2-l2}
\end{figure}

\textbf{2D Navigation: }Figure \ref{fig:analysis-2dnav-l2} present the representation similarity between tasks, across inner loop gradient updates for the second hidden layer of both policy networks in the 2D navigation environment. Again, similar patterns as highlighted in the first hidden layer of the policy networks in Figure \ref{fig:analysis-2dnav} emerge. Both networks are able to learn good (dissimilar) representations between tasks after few steps of inner loop gradient update. Also, both networks, before any gradient update, already have some level of representation disimilarity between tasks. Thus, this further highlights the fact the 2D Navigation environment has a low  complexity and requires very little adaptation of network parameters.

\textbf{CT-graph depth2: }Figure \ref{fig:analysis-ctgraphd2-l2} present the representation similarity between tasks, across inner loop gradient updates for the second hidden layer of both policy networks in the CT-graph depth2 benchmark. With increased problem complexity in comparison to the 2D navigation, only the neuromodulated policy network succeeds in learning distinct representations across tasks. The distinct representations thus allow the neuromodulated policy network to adapt optimally across tasks while the standard policy network struggles, as indicated by the performance plot in Figure \ref{fig:res-ctgraph-depth2}.

\subsection{CT-graph depth3 and depth4 Environments}
\label{analysis:ctgraph-d3-4}

\begin{figure}[ht!]
	\centering
	\begin{subfigure}{\textwidth}
		\centering
		\includegraphics[width=0.75\textwidth]{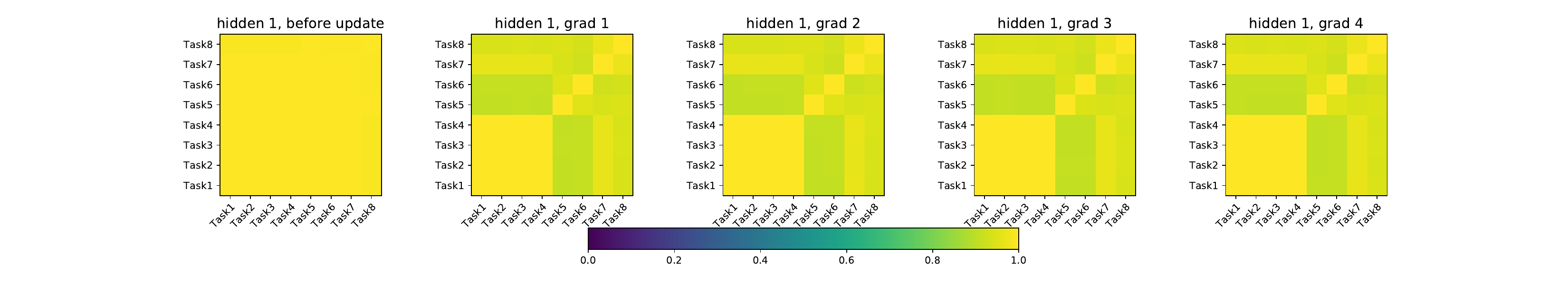}
		\caption{standard policy network, first hidden layer.}
		\label{fig:analysis-ctgraphd3-spn-l1}
	\end{subfigure}
    
	\begin{subfigure}{\textwidth}
		\centering
		\includegraphics[width=0.75\textwidth]{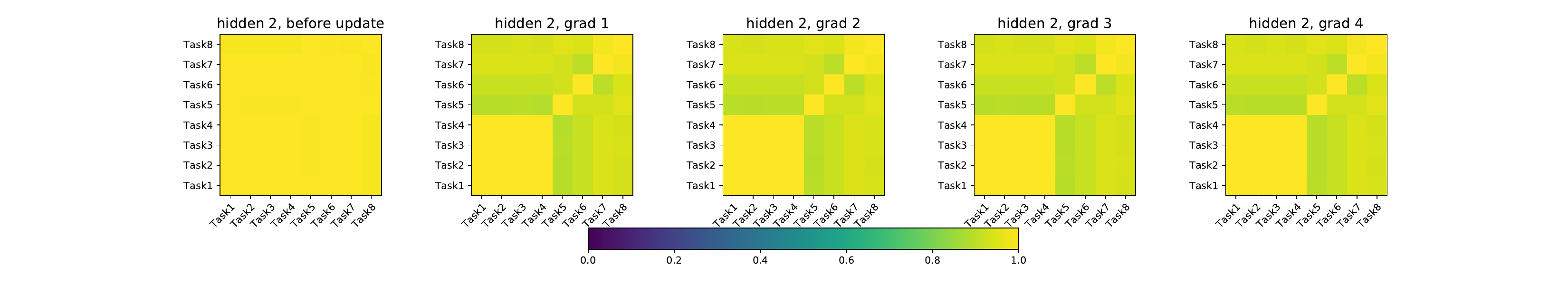}
		\caption{standard policy network, second hidden layer.}
		\label{fig:analysis-ctgraphd3-spn-l2}
	\end{subfigure}
    
	\begin{subfigure}{\textwidth}
		\centering
		\includegraphics[width=0.75\textwidth]{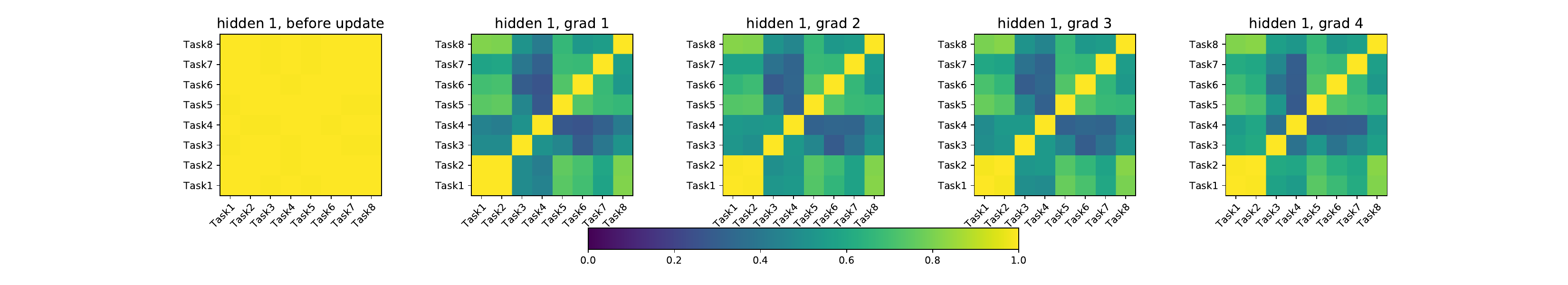}
		\caption{neuromodulated policy network, first hidden layer.}
		\label{fig:analysis-ctgraphd3-npn-l1}
	\end{subfigure}
    
	\begin{subfigure}{\textwidth}
		\centering
		\includegraphics[width=0.75\textwidth]{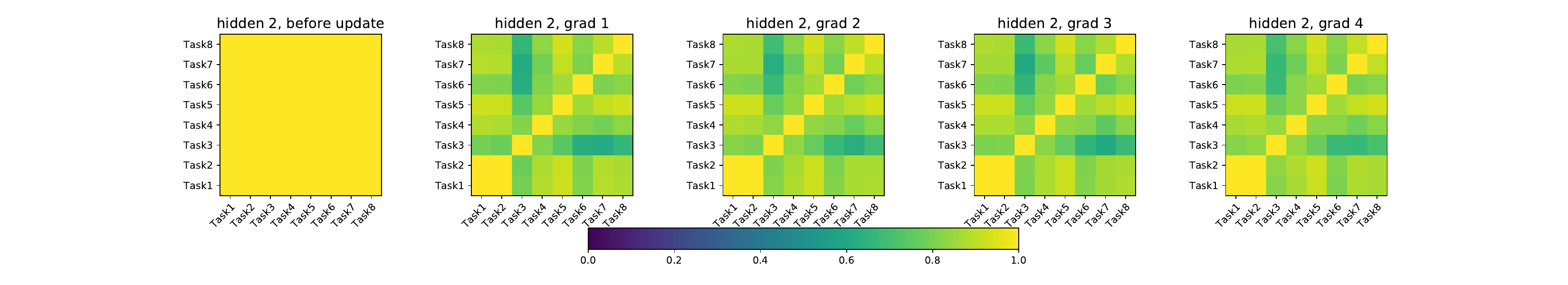}
		\caption{neuromodulated policy network, second hidden layer.}
		\label{fig:analysis-ctgraphd3-npn-l2}
	\end{subfigure}
	\caption{Representation similarities between tasks in the CT-graph depth3 environment.}
	\label{fig:analysis-ctgraphd3}
\end{figure}

\begin{figure}[ht!]
	\centering
	\begin{subfigure}{\textwidth}
		\centering
		\includegraphics[width=0.75\textwidth]{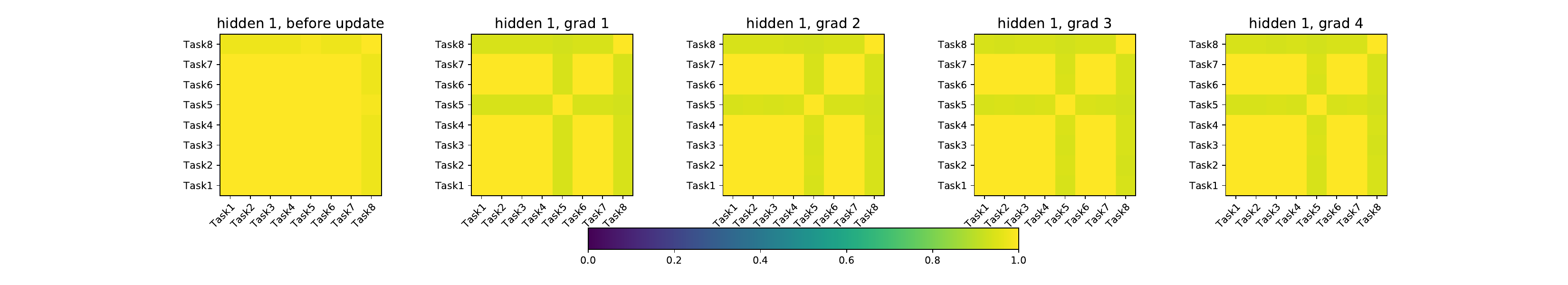}
		\caption{standard policy network, first hidden layer.}
		\label{fig:analysis-ctgraphd4-spn-l1}
	\end{subfigure}
    
	\begin{subfigure}{\textwidth}
		\centering
		\includegraphics[width=0.75\textwidth]{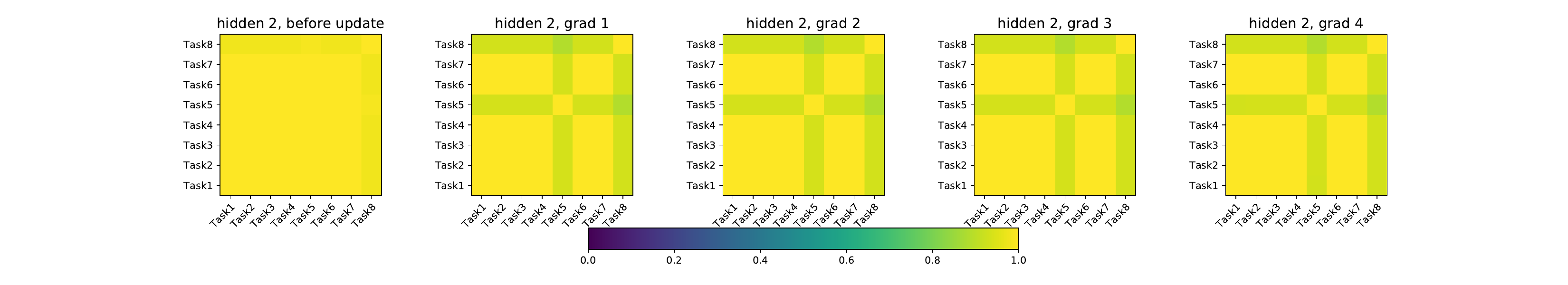}
		\caption{standard policy network, second hidden layer.}
		\label{fig:analysis-ctgraphd4-spn-l2}
	\end{subfigure}
    
	\begin{subfigure}{\textwidth}
		\centering
		\includegraphics[width=0.75\textwidth]{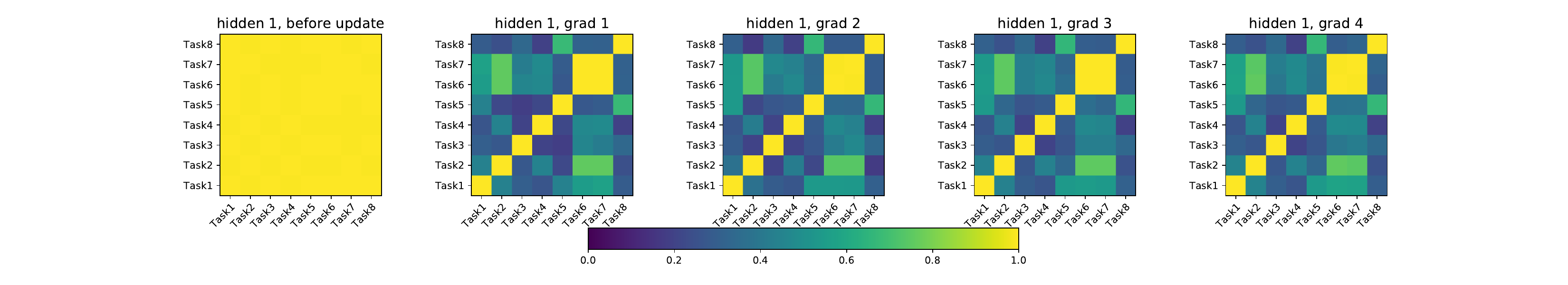}
		\caption{neuromodulated policy network, first hidden layer.}
		\label{fig:analysis-ctgraphd4-npn-l1}
	\end{subfigure}
	
	\begin{subfigure}{\textwidth}
		\centering
		\includegraphics[width=0.75\textwidth]{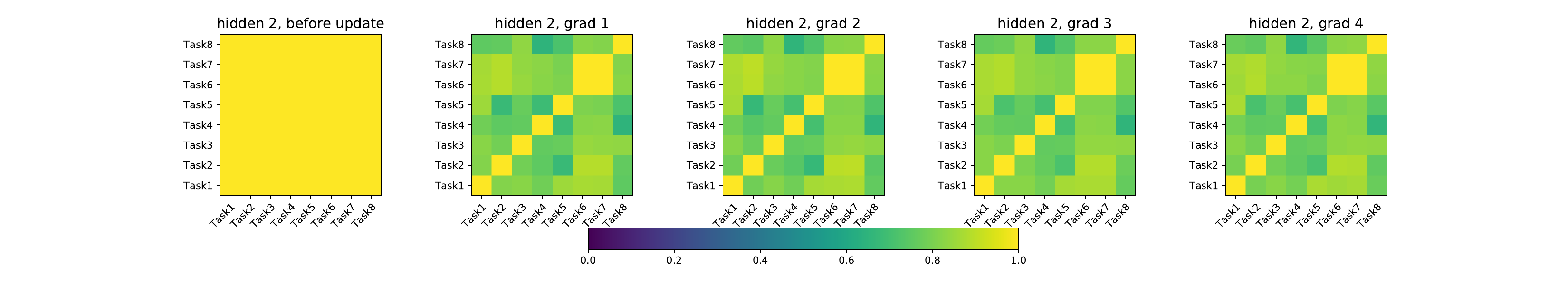}
		\caption{neuromodulated policy network, second hidden layer.}
		\label{fig:analysis-ctgraphd4-npn-l2}
	\end{subfigure}
	
	\caption{Representation similarities between tasks in the CT-graph depth4 environment.}
	\label{fig:analysis-ctgraphd4}
\end{figure}

The representation similarity plots (across inner loop gradient updates) between tasks of the hidden layers of both the standard and the neuromodulated policy networks in the CT-graph depth3 and depth4 environment instances are presented in Figure \ref{fig:analysis-ctgraphd3} and \ref{fig:analysis-ctgraphd4}. Again, as observed in Section \ref{subsection:analysis}, the standard policy network still struggles to learn distinct representations for each task, whereas, the neuromodulated policy network is able to learn the required task-specific representations. Hence, the neuromodulated policy network is able to perform optimally across tasks. This explains the difference in performance (Figures \ref{fig:res-ctgraph-depth3} and \ref{fig:res-ctgraph-depth4}) between the policy networks.

\subsection{Half-Cheetah and Meta-World Environments}
\label{analysis:cheetah-meta-world}
The CAVIA policies analysis plots for the half-cheetah and meta-world benchmarks are presented in this section.

\begin{figure}[ht!]
	\centering
	\begin{subfigure}{\textwidth}
		\centering
		\includegraphics[width=0.75\textwidth]{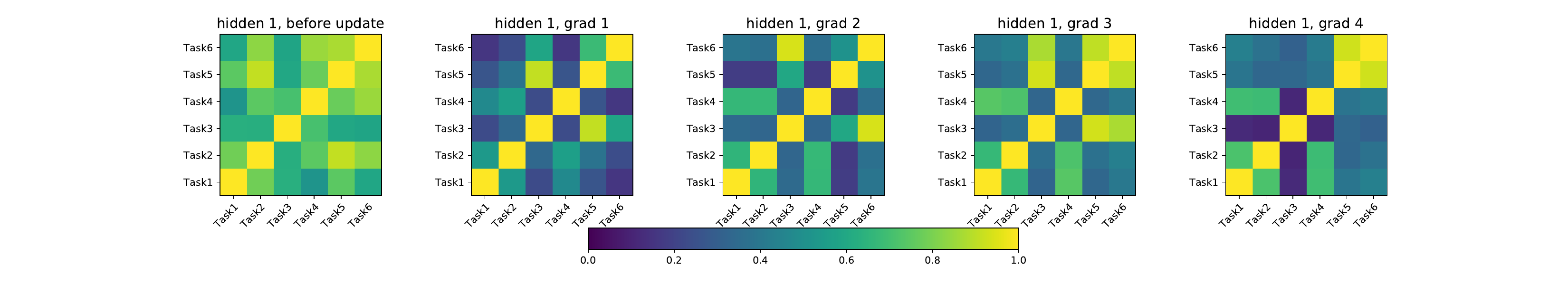}
		\caption{standard policy network, first hidden layer.}
		\label{fig:analysis-halfcheetahdir-spn-l1}
	\end{subfigure}
    \vspace{-0.1cm}
    
	\begin{subfigure}{\textwidth}
		\centering
		\includegraphics[width=0.75\textwidth]{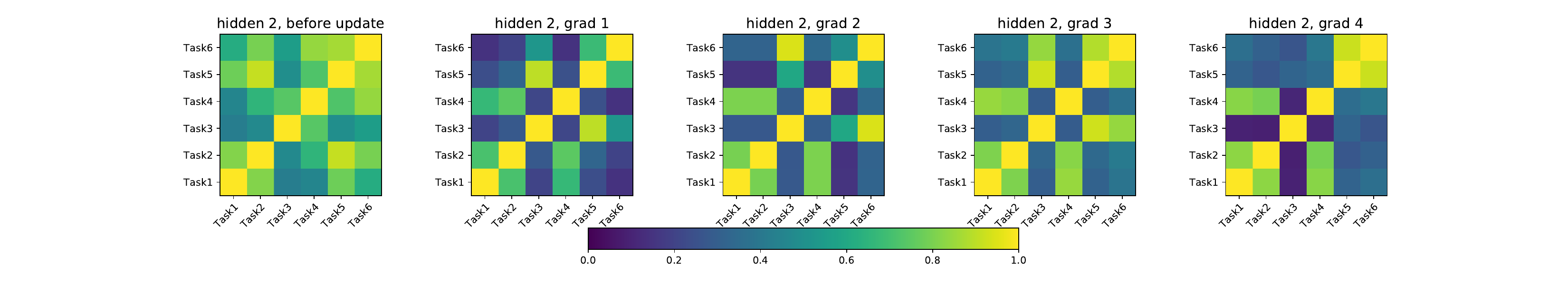}
		\caption{standard policy network, second hidden layer.}
		\label{fig:analysis-halfcheetahdir-spn-l2}
	\end{subfigure}
    \vspace{-0.1cm}
    
	\begin{subfigure}{\textwidth}
		\centering
		\includegraphics[width=0.75\textwidth]{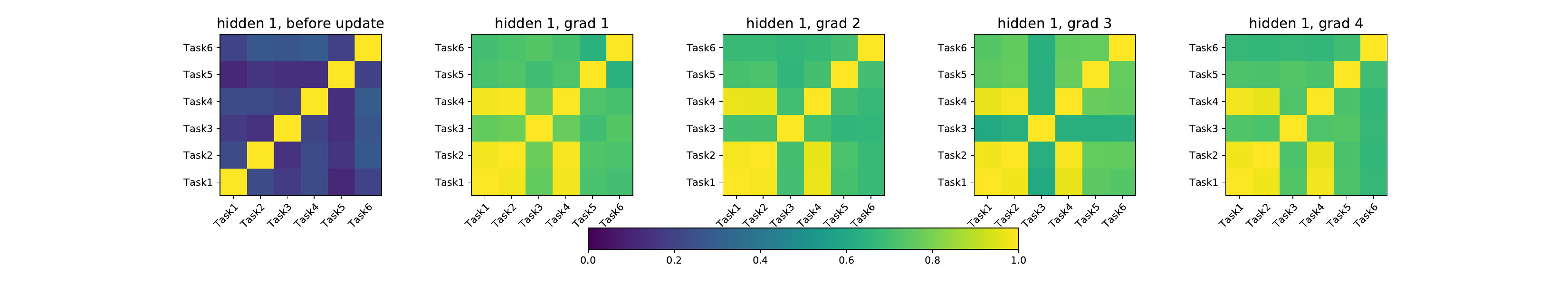}
		\caption{neuromodulated policy network, first hidden layer.}
		\label{fig:analysis-halfcheetahdir-npn-l1}
	\end{subfigure}
    \vspace{-0.1cm}
    
	\begin{subfigure}{\textwidth}
		\centering
		\includegraphics[width=0.75\textwidth]{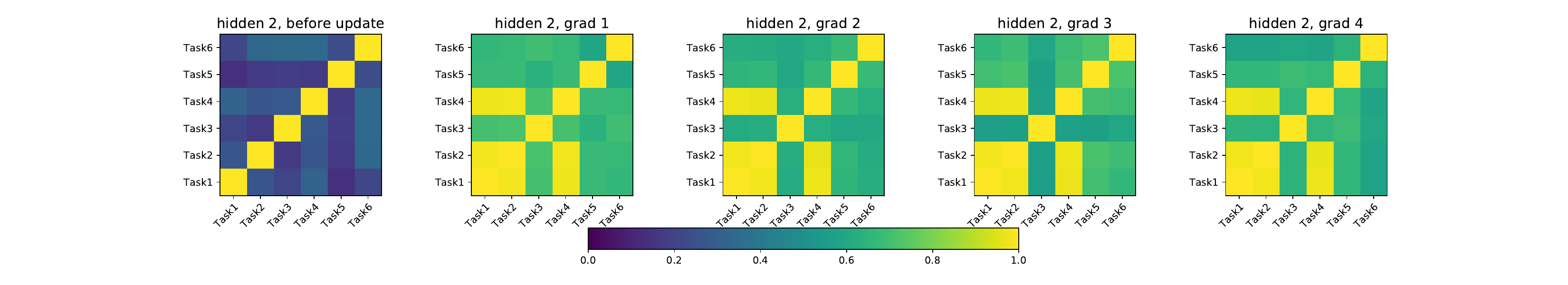}
		\caption{neuromodulated policy network, second hidden layer.}
		\label{fig:analysis-halfcheetahdir-npn-l2}
	\end{subfigure}
	\caption{Representation similarities between tasks in the Half-Cheetah Direction environment.}
	\label{fig:analysis-halfcheetahdir}
\end{figure}

\begin{figure}[ht!]
	\centering
	\begin{subfigure}{\textwidth}
		\centering
		\includegraphics[width=0.75\textwidth]{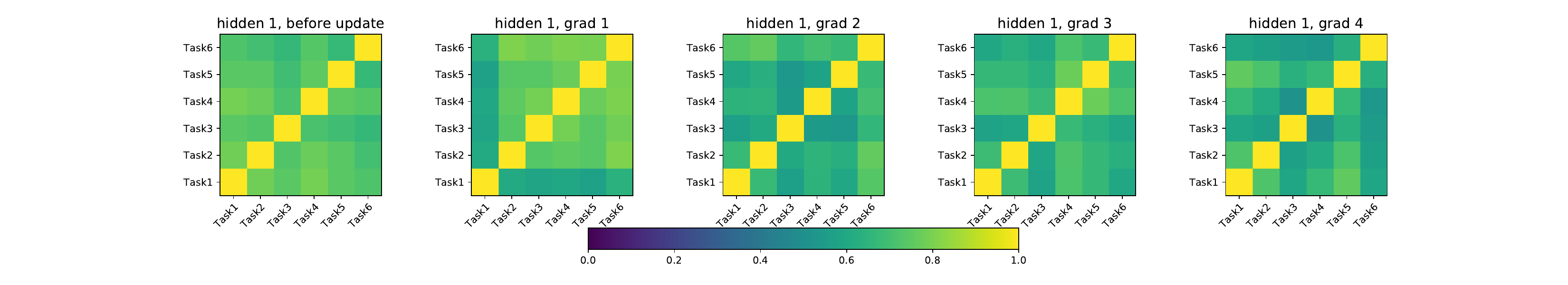}
		\caption{standard policy network, first hidden layer.}
		\label{fig:analysis-halfcheetahvel-spn-l1}
	\end{subfigure}
    \vspace{-0.1cm}
    
	\begin{subfigure}{\textwidth}
		\centering
		\includegraphics[width=0.75\textwidth]{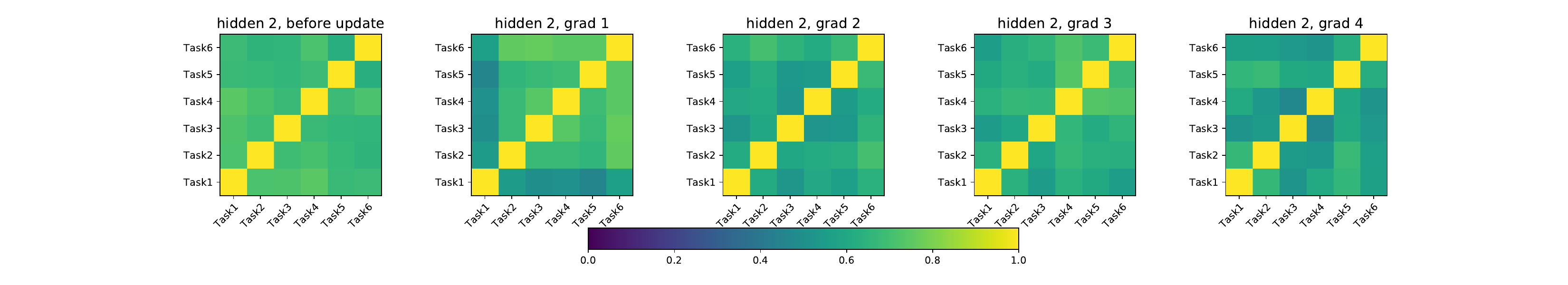}
		\caption{standard policy network, second hidden layer.}
		\label{fig:analysis-halfcheetahvel-spn-l2}
	\end{subfigure}
    \vspace{-0.1cm}
    
	\begin{subfigure}{\textwidth}
		\centering
		\includegraphics[width=0.75\textwidth]{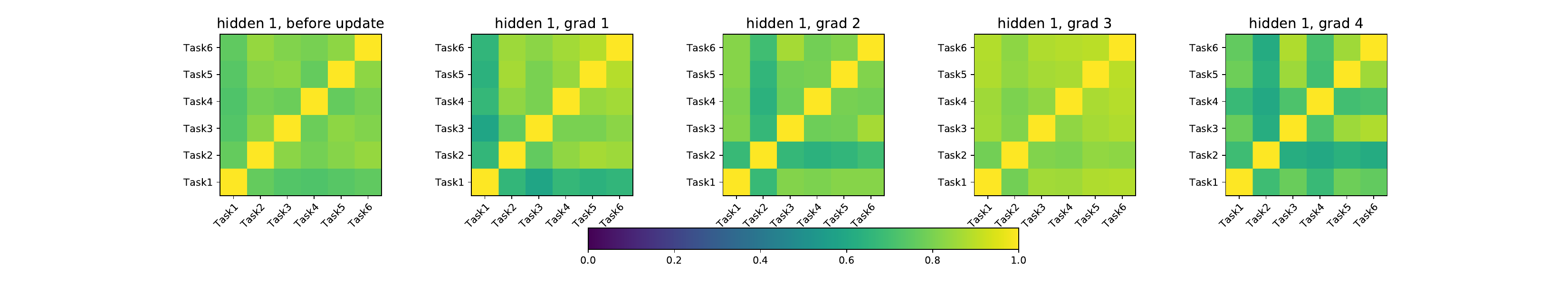}
		\caption{neuromodulated policy network, first hidden layer.}
		\label{fig:analysis-halfcheetahvel-npn-l1}
	\end{subfigure}
    \vspace{-0.1cm}
    
	\begin{subfigure}{\textwidth}
		\centering
		\includegraphics[width=0.75\textwidth]{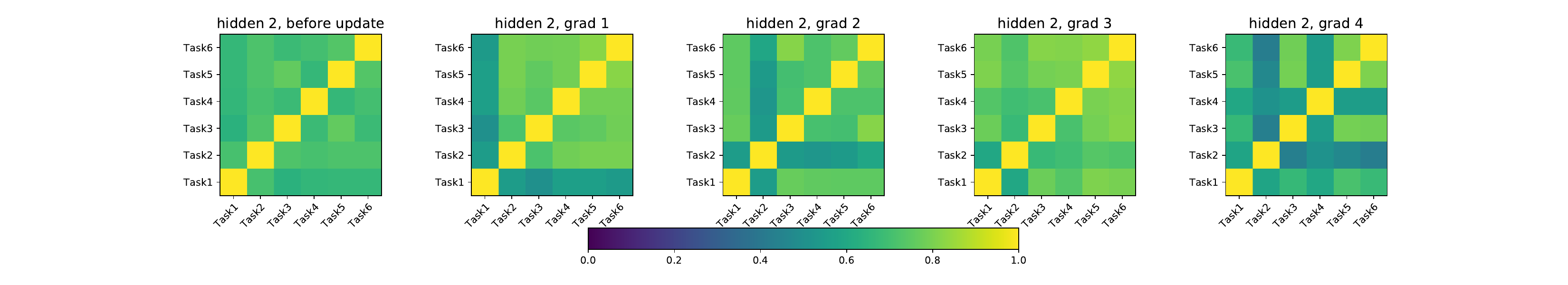}
		\caption{neuromodulated policy network, second hidden layer.}
		\label{fig:analysis-halfcheetahvel-npn-l2}
	\end{subfigure}
	\caption{Representation similarities between tasks in the Half-Cheetah Velocity environment.}
	\label{fig:analysis-halfcheetahvel}
\end{figure}

\begin{figure}[ht!]
	\centering
	\begin{subfigure}{\textwidth}
		\centering
		\includegraphics[width=0.75\textwidth]{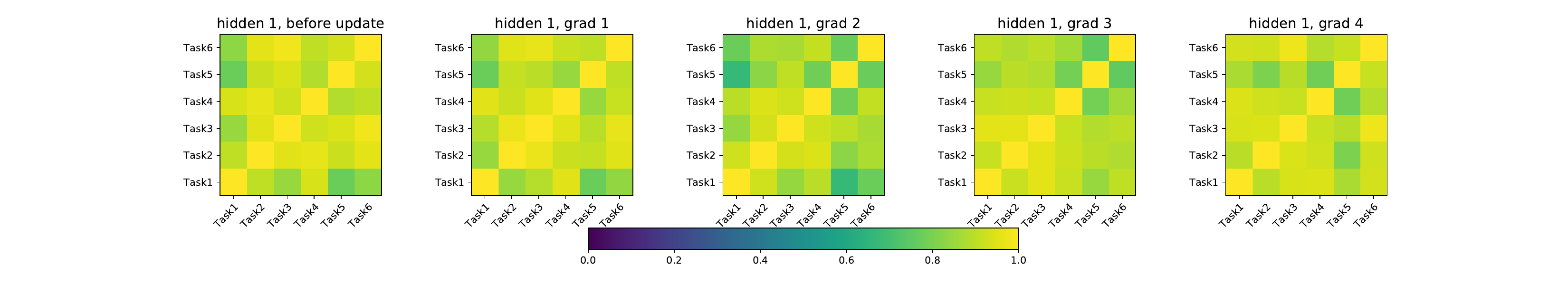}
		\caption{standard policy network, first hidden layer.}
		\label{fig:analysis-ml1-spn-l1}
	\end{subfigure}
    
	\begin{subfigure}{\textwidth}
		\centering
		\includegraphics[width=0.75\textwidth]{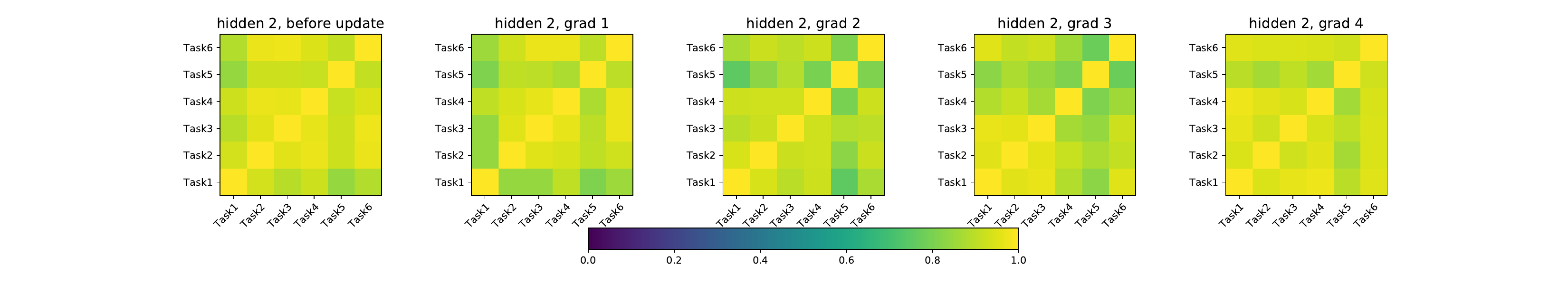}
		\caption{standard policy network, second hidden layer.}
		\label{fig:analysis-ml1-spn-l2}
	\end{subfigure}
    
	\begin{subfigure}{\textwidth}
		\centering
		\includegraphics[width=0.75\textwidth]{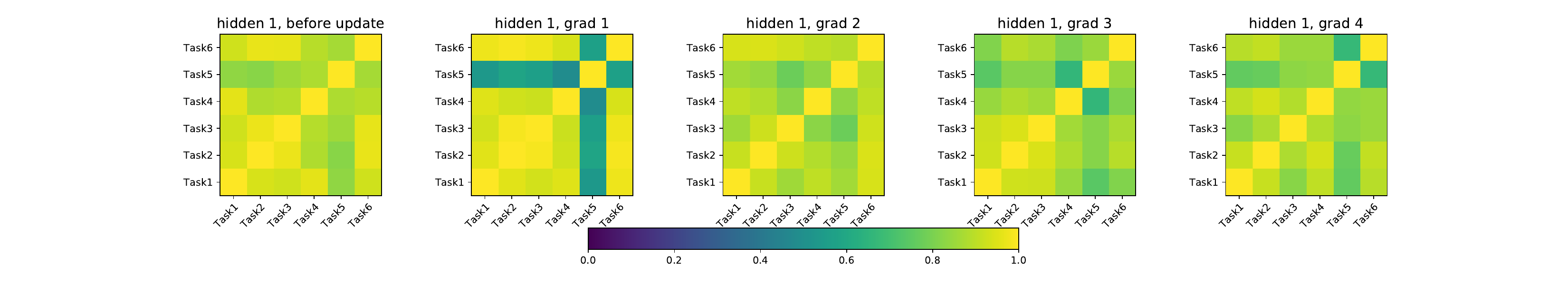}
		\caption{neuromodulated policy network, first hidden layer.}
		\label{fig:analysis-ml1-npn-l1}
	\end{subfigure}
    
	\begin{subfigure}{\textwidth}
		\centering
		\includegraphics[width=0.75\textwidth]{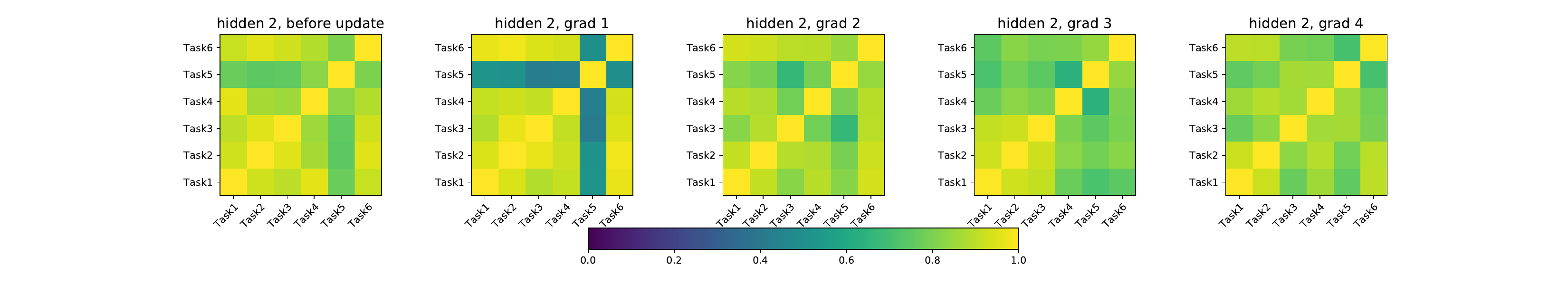}
		\caption{neuromodulated policy network, second hidden layer.}
		\label{fig:analysis-ml1-npn-l2}
	\end{subfigure}
	\caption{Representation similarities between tasks in the ML1 (meta-world) environment.}
	\label{fig:analysis-ml1}
\end{figure}

\begin{figure}[ht!]
	\centering
	\begin{subfigure}{\textwidth}
		\centering
		\includegraphics[width=0.75\textwidth]{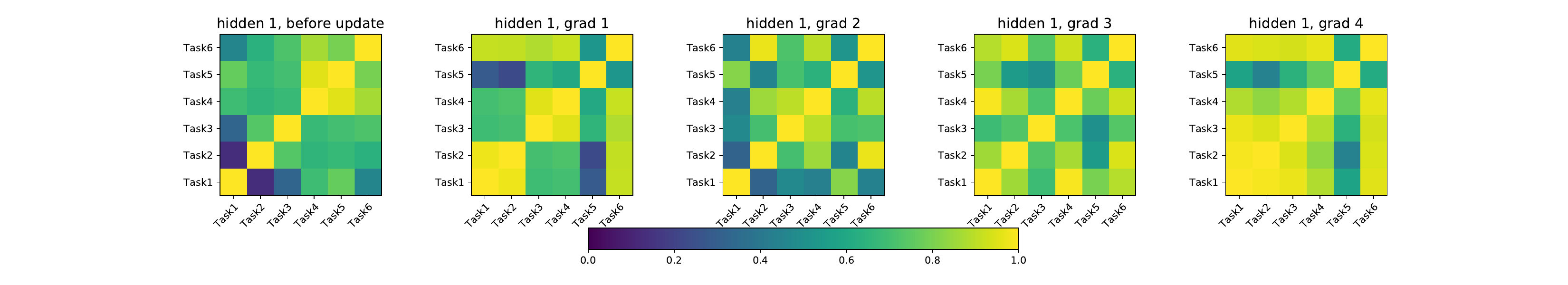}
		\caption{standard policy network, first hidden layer.}
		\label{fig:analysis-ml45-spn-l1}
	\end{subfigure}
    
	\begin{subfigure}{\textwidth}
		\centering
		\includegraphics[width=0.75\textwidth]{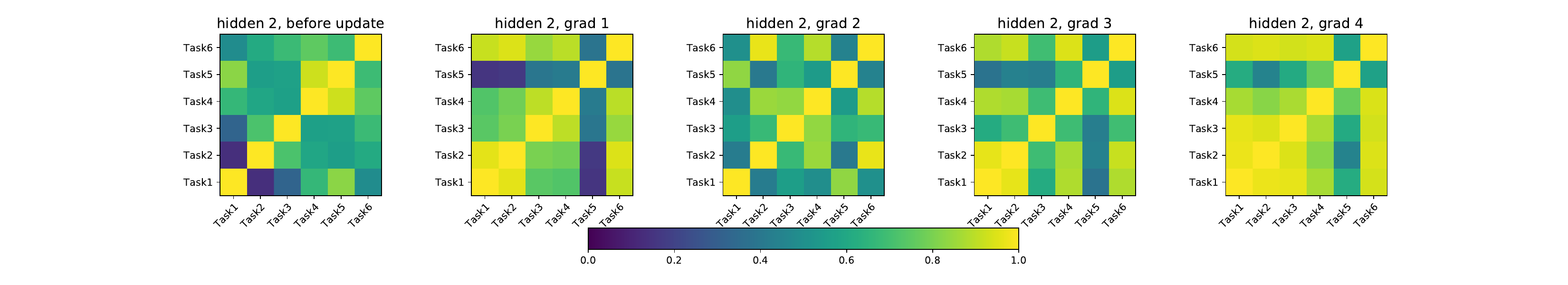}
		\caption{standard policy network, second hidden layer.}
		\label{fig:analysis-ml45-spn-l2}
	\end{subfigure}
    
	\begin{subfigure}{\textwidth}
		\centering
		\includegraphics[width=0.75\textwidth]{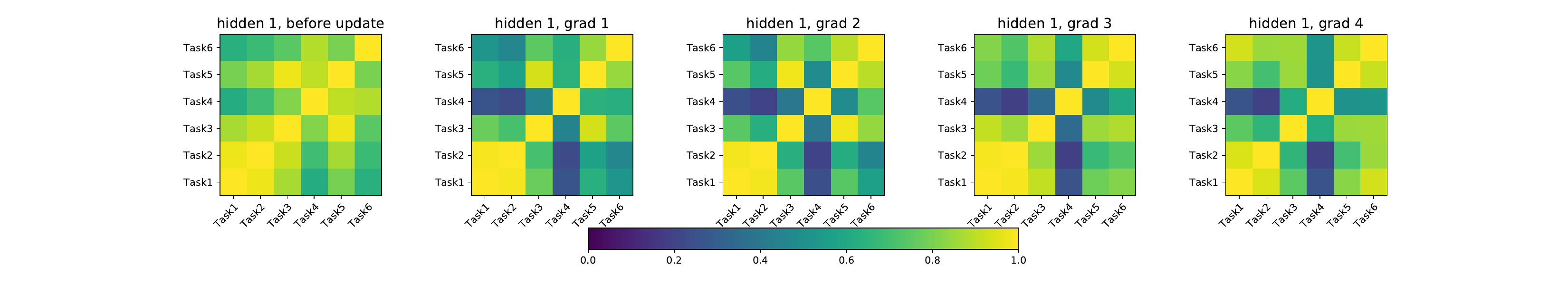}
		\caption{neuromodulated policy network, first hidden layer.}
		\label{fig:analysis-ml45-npn-l1}
	\end{subfigure}
	
	\begin{subfigure}{\textwidth}
		\centering
		\includegraphics[width=0.75\textwidth]{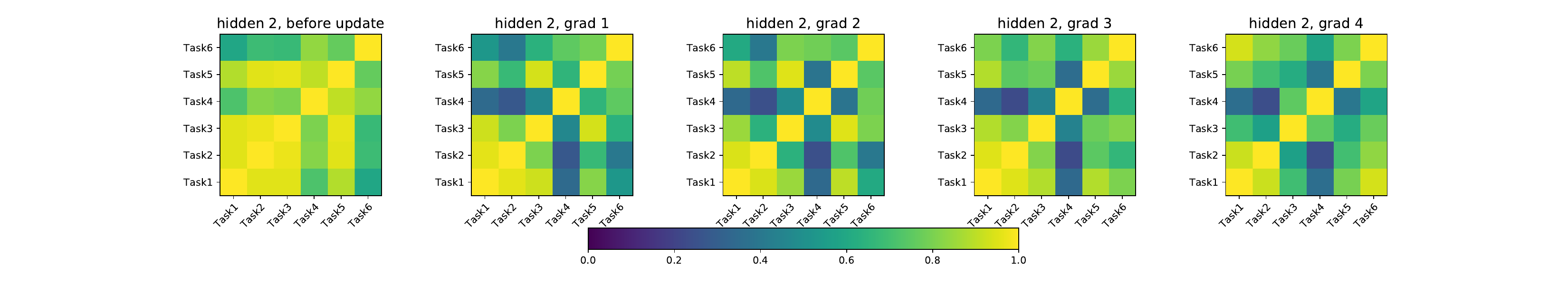}
		\caption{neuromodulated policy network, second hidden layer.}
		\label{fig:analysis-ml45-npn-l2}
	\end{subfigure}
	
	\caption{Representation similarities between tasks in the ML45 (meta-world)  environment.}
	\label{fig:analysis-ml45}
\end{figure}

\textbf{Half-Cheetah: } Figure \ref{fig:analysis-halfcheetahdir} and \ref{fig:analysis-halfcheetahvel} shows the representation similarity plots of the standard and neuromodulated policy networks for the Half-Cheetah direction and velocity environments respectively. In this setting where the problems are of simple to medium complexity, both policy networks are able to learn efficient (dissimilar) representations between tasks, further buttressing the discussions in Section \ref{subsection:analysis}. In fact, we observe that the standard policy network learns better dissimilar representations across tasks for the half-cheetah direction environment due to the simplicity of the tasks in the environment in comparison to the velocity environment.

\textbf{Meta-World: } Also, Figure \ref{fig:analysis-ml1} and \ref{fig:analysis-ml45} depicts the representation similarity plots for the ML1 and ML45 meta-world environment. Similar to the observations in Section \ref{subsection:analysis}, we observe again that as the problem complexity increase (i.e., from ML1 to ML45), the neuromodulated policy network produces better (dissimilar) representations across the sampled tasks in comparison to the standard policy network.

\subsection{Representation similarity of neuromodulatory activities across tasks}
\label{analysis:other-repr-similarity-nm}
\begin{figure}[ht!]
	\centering
	\begin{subfigure}{\textwidth}
		\centering
		\includegraphics[width=\textwidth]{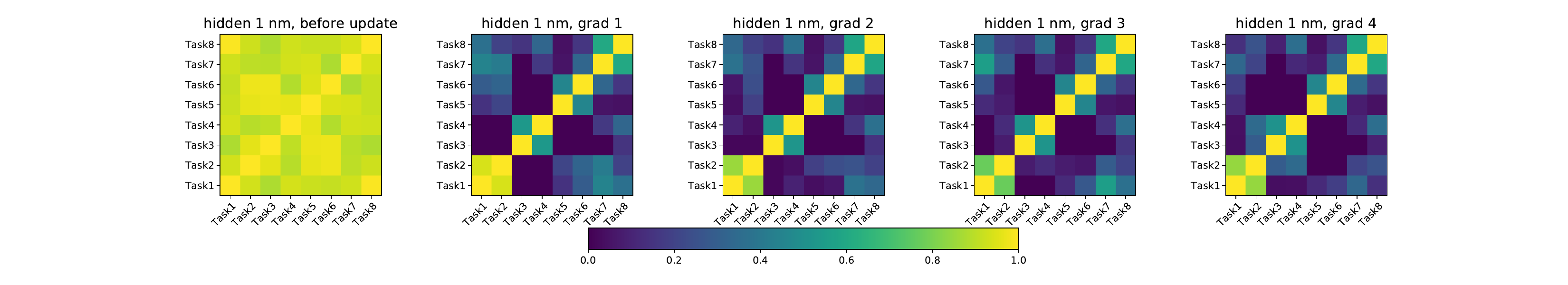}
		\caption{first hidden layer.}
		\label{fig:analysis-ctgraphd3-nm-npn-l1}
	\end{subfigure}
	\begin{subfigure}{\textwidth}
		\centering
		\includegraphics[width=\textwidth]{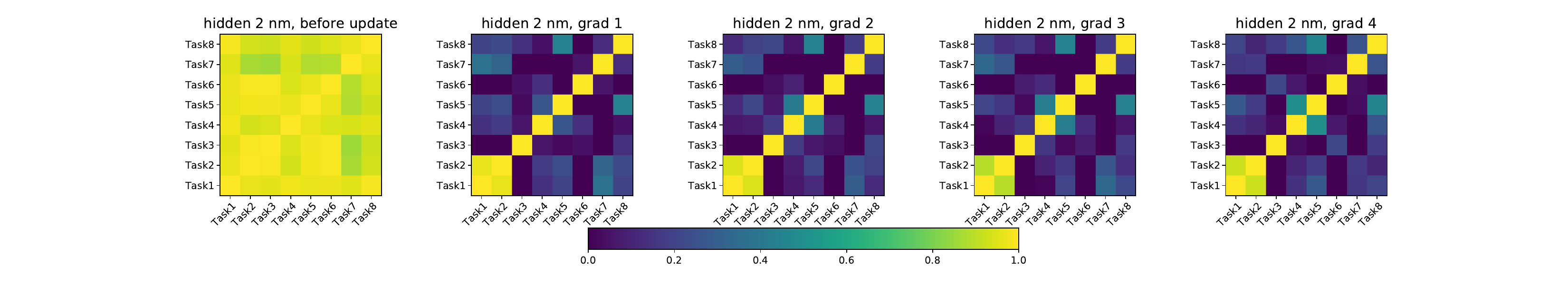}
		\caption{second hidden layer.}
		\label{fig:analysis-ctgraphd3-nm-npn-l2}
	\end{subfigure}
	\caption{Representation similarities of neuromodulatory activities $h^m$ between tasks in the CT-graph depth3 environment.}
	\label{fig:analysis-ctgraphd3-nm}
\end{figure}

\begin{figure}[ht!]
	\centering
	\begin{subfigure}{\textwidth}
		\centering
		\includegraphics[width=\textwidth]{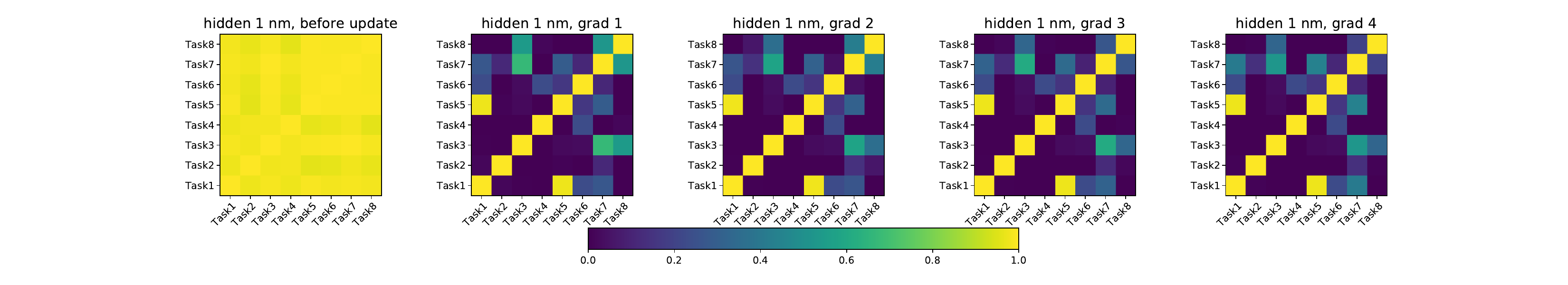}
		\caption{first hidden layer.}
		\label{fig:analysis-ctgraphd4-nm-npn-l1}
	\end{subfigure}
	\begin{subfigure}{\textwidth}
		\centering
		\includegraphics[width=\textwidth]{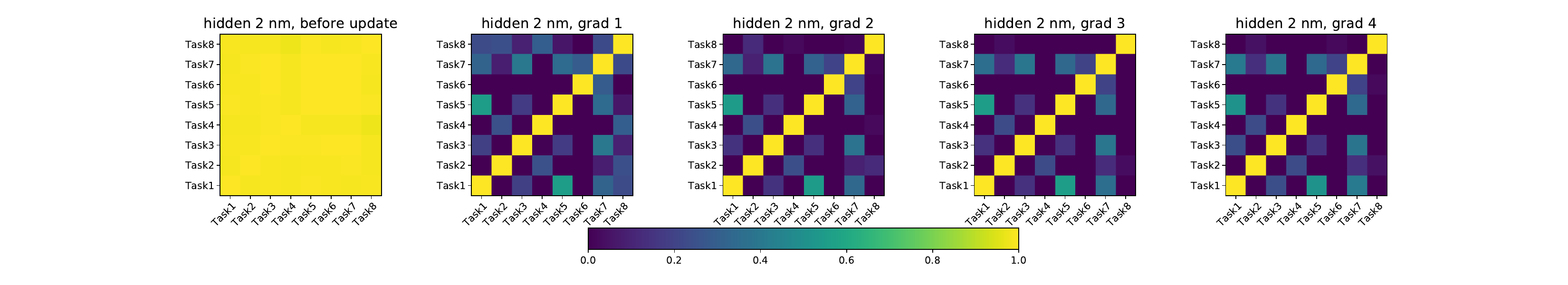}
		\caption{second hidden layer.}
		\label{fig:analysis-ctgraphd4-nm-npn-l2}
	\end{subfigure}
	\caption{Representation similarities of neuromodulatory activities $h^m$ between tasks in the CT-graph depth4 environment.}
	\label{fig:analysis-ctgraphd4-nm}
\end{figure}

\begin{figure}[ht!]
	\centering
	\begin{subfigure}{\textwidth}
		\centering
		\includegraphics[width=\textwidth]{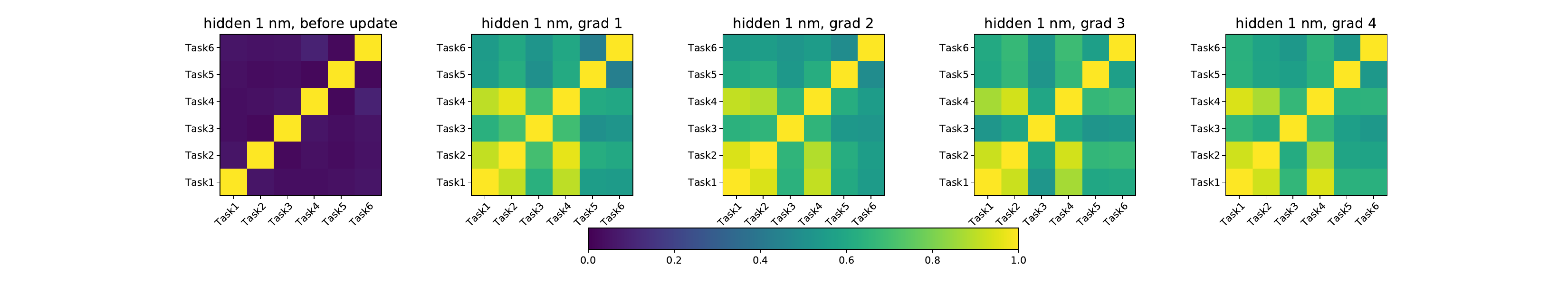}
		\caption{first hidden layer.}
		\label{fig:analysis-halfcheetahdir-nm-npn-l1}
	\end{subfigure}
	\begin{subfigure}{\textwidth}
		\centering
		\includegraphics[width=\textwidth]{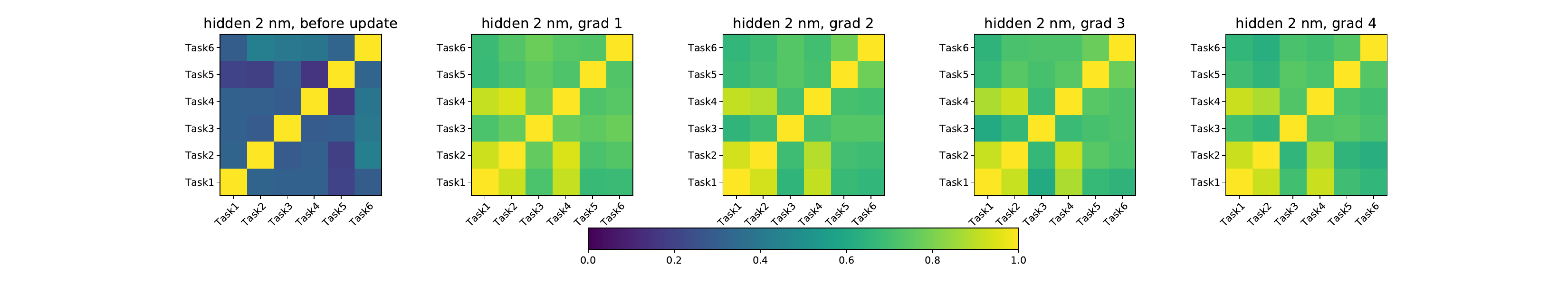}
		\caption{second hidden layer.}
		\label{fig:analysis-halfcheetahdir-nm-npn-l2}
	\end{subfigure}
	\caption{Representation similarities of neuromodulatory activities $h^m$ between tasks in the Half-Cheetah Direction environment.}
	\label{fig:analysis-halfcheetahdir-nm}
\end{figure}

\begin{figure}[ht!]
	\centering
	\begin{subfigure}{\textwidth}
		\centering
		\includegraphics[width=\textwidth]{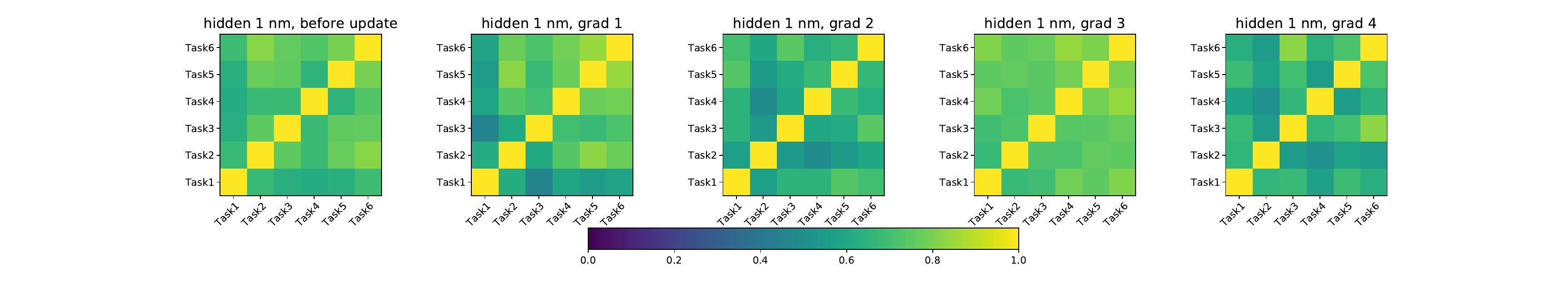}
		\caption{first hidden layer.}
		\label{fig:analysis-halfcheetahvel-nm-npn-l1}
	\end{subfigure}
	\begin{subfigure}{\textwidth}
		\centering
		\includegraphics[width=\textwidth]{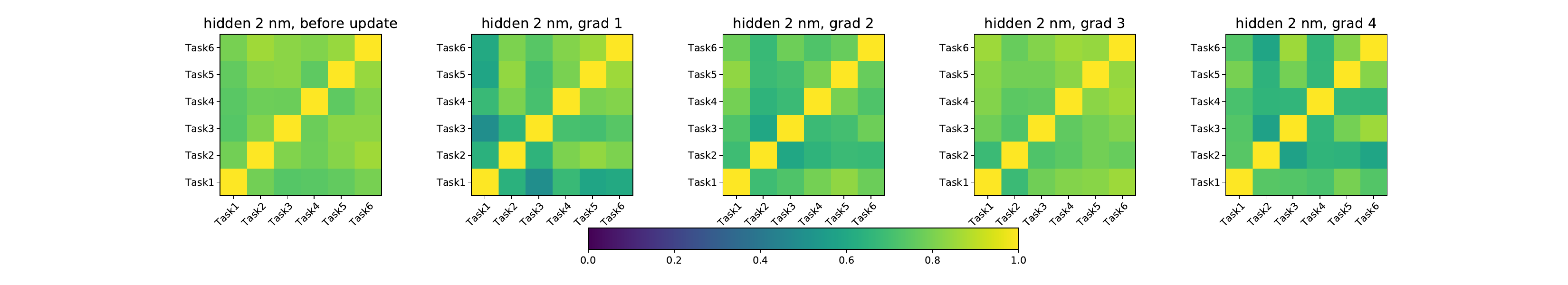}
		\caption{second hidden layer.}
		\label{fig:analysis-halfcheetahvel-nm-npn-l2}
	\end{subfigure}
	\caption{Representation similarities of neuromodulatory activities $h^m$ between tasks in the Half-Cheetah Velocity environment.}
	\label{fig:analysis-halfcheetahvel-nm}
\end{figure}

\begin{figure}[ht!]
	\centering
	\begin{subfigure}{\textwidth}
		\centering
		\includegraphics[width=\textwidth]{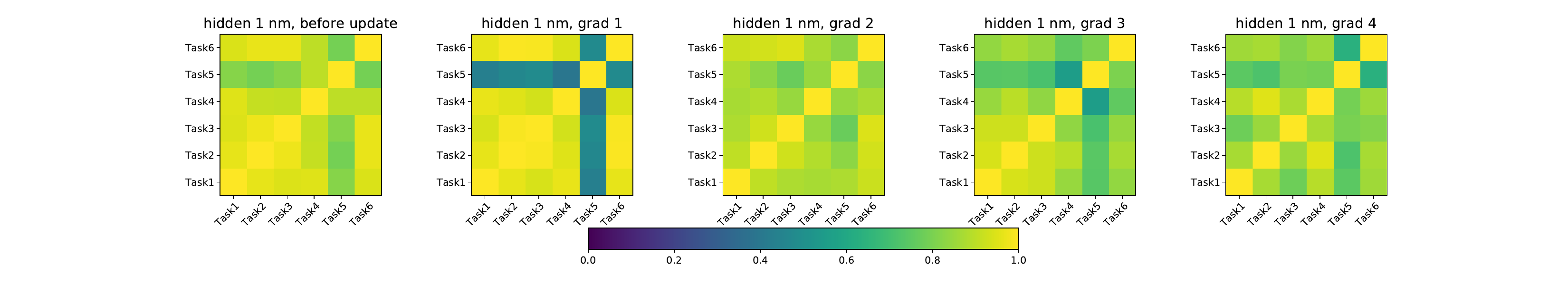}
		\caption{first hidden layer.}
		\label{fig:analysis-ml1-nm-npn-l1}
	\end{subfigure}
	\begin{subfigure}{\textwidth}
		\centering
		\includegraphics[width=\textwidth]{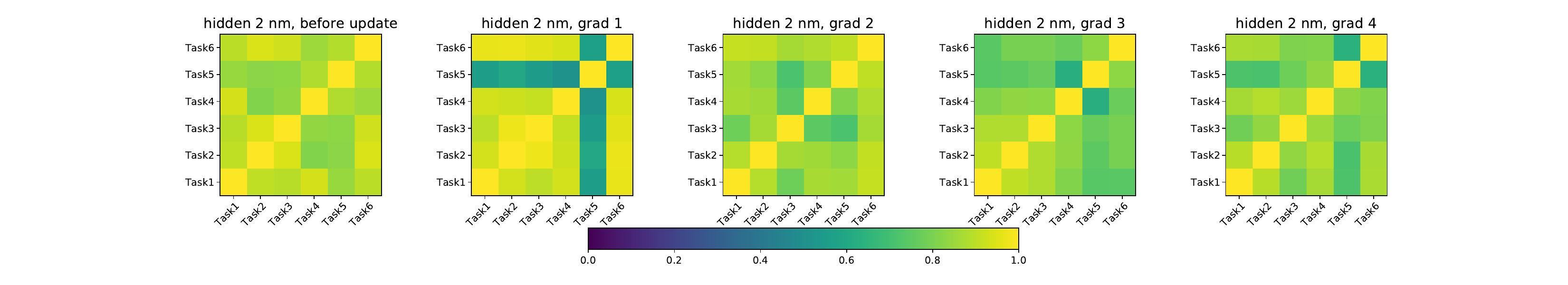}
		\caption{second hidden layer.}
		\label{fig:analysis-ml1-nm-npn-l2}
	\end{subfigure}
	\caption{Representation similarities of neuromodulatory activities $h^m$ between tasks in the Meta-World ML1 environment.}
	\label{fig:analysis-ml1-nm}
\end{figure}

The representation similarity plots of the neuromodulatory activities $h^m$ across tasks are presented for the CT-graph depth 3 and 4, Half-Cheetah Velocity and Direction, and Meta-World ML1 environments. Again, we observe dissimilar representation across tasks for the neuromodulatory activities. Hence, neuromodulation facilitates the dynamic representations in the policy network enabling optimal adaptation behaviour in complex problems with dissimilar tasks.

\clearpage

\bibliographystyle{elsarticle-harv} 
\bibliography{main}

\end{document}